\documentclass{article}

\usepackage[english]{babel}

\usepackage[letterpaper,top=2cm,bottom=2cm,left=3cm,right=3cm,marginparwidth=1.75cm]{geometry}

\usepackage{amsmath, amssymb}
\usepackage{amsthm}
\usepackage{graphicx}
\usepackage[colorlinks=true, allcolors=blue]{hyperref}
\usepackage{colortbl} % for setting cell colors via \cellcolor{red}
\usepackage{makecell}
\usepackage{xspace} % for conditional spacing in commands (i.e. no spacing when there's punctuation and spacing otherwise)
\usepackage{enumitem} % for todo lists
\usepackage{array}
\usepackage{rotating}
\usepackage{booktabs}
\usepackage{xurl}
\usepackage[htt]{hyphenat}
\usepackage{alltt}
\usepackage{natbib}

\newcommand*\justify{%
  \fontdimen2\font=0.4em% interword space
  \fontdimen3\font=0.2em% interword stretch
  \fontdimen4\font=0.1em% interword shrink
  \fontdimen7\font=0.1em% extra space
  \hyphenchar\font=`\-% allowing hyphenation
}

\newlist{todolist}{itemize}{2}
\setlist[todolist]{label=$\square$}

\makeatletter
\let\over\@@over
\makeatother

\usepackage[T1]{fontenc} % so we can use curly brackets inside \texttt{...}

\usepackage{adjustbox}
\usepackage{array}

\newcolumntype{R}[2]{%
    >{\adjustbox{angle=#1,lap=\width-(#2)}\bgroup}%
    l%
    <{\egroup}%
}
\newcommand*\rot{\multicolumn{1}{R{45}{1em}}}% no optional argument here, please!

\newcommand{\gpt}{\emph{GPT--$4$o}\xspace}
\newcommand{\minigpt}{\emph{GPT--$4$o-mini}\xspace}

\newcommand{\oone}{\emph{o$1$--preview}\xspace}
\newcommand{\newclaude}{\emph{Claude--$3.5$ Sonnet}\xspace}
\newcommand{\minillama}{\emph{Llama 3.1 ($8$b)}\xspace}
\newcommand{\llama}{\emph{Llama 3.1 ($405$b)}\xspace}

\newcommand{\partialsuccessbox}{\colorbox[rgb]{1,0.75,0}{\phantom{XX}}}
\newcommand{\failurebox}{\colorbox[rgb]{1,0.6,0.6}{\phantom{XX}}}
\newcommand{\successbox}{\colorbox[rgb]{0.6,1,0.6}{\phantom{XX}}}
\newcommand{\NAbox}{\colorbox[rgb]{1,1,1}{\phantom{XX}}}

\newcommand{\partialsuccessboxsmall}{\colorbox[rgb]{1,0.75,0}{\phantom{X}}}
\newcommand{\failureboxsmall}{\colorbox[rgb]{1,0.6,0.6}{\phantom{X}}}
\newcommand{\successboxsmall}{\colorbox[rgb]{0.6,1,0.6}{\phantom{X}}}
\newcommand{\NAboxsmall}{\colorbox[rgb]{1,1,1}{\phantom{X}}}

\newcommand{\oonelabel}{\emph{o1-preview}}
\newcommand{\newclaudelabel}{\emph{claude-3.5-sonnet}}
\newcommand{\gptlabel}{\emph{gpt-4o}}
\newcommand{\llamalabel}{\emph{llama-3.1-405b}}
\newcommand{\minillamalabel}{\emph{llama-3.1-8b}}
\newcommand{\minigptlabel}{\emph{gpt-4o-mini}}
\newcommand{\olabel}{\emph{o1}}
\newcommand{\react}{\emph{react-}}
\newcommand{\delegate}{\emph{delegate-}}
\newcommand{\plan}{\emph{plan-}}
\newcommand{\plandelegate}{\emph{plan-delegate-}}

\title{Towards a Realistic Long-Term Benchmark for Open-Web Research Agents}
\author{Peter Mühlbacher\\
\texttt{peter@futuresearch.ai} \and Nikos I. Bosse\\
\texttt{nikos@futuresearch.ai} \and Lawrence Phillips\\
\texttt{lawrence@futuresearch.ai}}

\begin{document}
\maketitle

\begin{abstract}
    % \textbf{Background}
    % Agents are very important and have the potential to profoundly transform society. Well-designed benchmarks are important because they help us to a) measure progress and b) provide evidence that might help us estimate future progress. 
    % Past benchmarks have focused on X and failed to account for Y. 

    % \textbf{Methods}
    % In our paper we do X, which is the first step of a bigger project XYZ. We evaluated a set of X LLMs-agents using four different architectures with respect to their ability to solve real-world tasks such as XXX. We created agents that either can or cannot do centralized planning and can or cannot delegate subtasks to subagents and defined a set of XX tasks. For each task we defined a scoring rubric and evaluated LLMs on their performance to solve the tasks, assigning partial credit. We manually inspected the agent traces and reflect on our observations.

    % \textbf{Results}
    % Architecture Y was best. LLM X was best. We found Z variation in performance. Clear progress from model A to Z. Larger models performing better. Claude-Sonnet was better at reasoning, but knew less than GPT-4o. All models showed substantial reasoning errors. Tendency to fail to follow instructions. Getting stuck in loops. 

    % \textbf{Conclusions}. 
    % Coming up with a benchmark is difficult. Challenge is that capabilities can jump quite sudden. We saw relatively smooth prgoress. Some jumps? Predicitve capability limited, but first good indicicator. Will it scale? Future steps. 

    We present initial results of a forthcoming benchmark for evaluating LLM agents on white-collar tasks of economic value. We evaluate agents on real-world ``messy'' open-web research tasks of the type that are routine in finance and consulting. In doing so, we lay the groundwork for an LLM agent evaluation suite where good performance directly corresponds to a large economic and societal impact.

    We built and tested several agent architectures with \oone, \gpt, \newclaude, \llama, and \minigpt. On average, LLM agents powered by \newclaude and \oone substantially outperformed agents using \gpt, with agents based on \llama and \minigpt lagging noticeably behind. Across LLMs, a ReAct architecture with the ability to delegate subtasks to subagents performed best. In addition to quantitative evaluations, we qualitatively assessed the performance of the LLM agents by inspecting their traces and reflecting on their observations.

    Our evaluation represents the first in-depth assessment of agents' abilities to conduct challenging, economically valuable analyst-style research on the real open web.
\end{abstract}
% \tableofcontents
\newpage
\section{Introduction}

The short-to-medium term societal and economic impact of large language model (LLM) agents remains a subject of intense debate. Predictions range from minimal influence—where current enthusiasm is viewed as premature or overhyped—to profound transformations, including widespread automation of white-collar jobs and unprecedented revenue generation for AI companies.

A good understanding of the capabilities of LLM agents, and of how these capabilities are evolving over time, will shed light on this important question. Indeed, assuming that capabilities continue to improve with some degree of continuity, a suitably designed benchmark could act as an early warning system for the onset of an unprecedented socioeconomic transformation. For a benchmark to act as a useful indicator of potential for future impact, the following properties are important:
\begin{itemize}
\item \emph{Performance improves incrementally upon frontier model releases, for a large number of releases.} \\
The main way in which benchmarks can inform us about future impact is via extrapolation, and in order for extrapolation to be possible, performance must evolve somewhat smoothly as new frontier models are released. Ideally, we would observe smooth gains across a large number of releases, increasing our belief that extrapolations are genuinely informative.
\item\emph{Performance is clearly linked to impact.} \\
Good performance on the benchmark should correspond as clearly as possible to the potential for significant impact. The benchmark should evaluate performance on tasks that are widely accepted as important, and it should aim to evaluate performance on real-world versions of these tasks.
\end{itemize}

% stretch: current benchmarks don't have these properties.
% MISSING: ANY MENTION OF PAST LITERATURE
% - what others have done: Past researchers have done XYZ, this has failed for ZYX. (this part is currently missing, we don't really have a lot of references there)
% - This leaves ABC as a research gap. We're filling this in the following way: EFG. (basically what we have in the current "motivation" section)
% EFG is part of a bigger project EFGHIJK, this paper does the first part. 

% Much previous work has focussed on creating benchmarks measuring reasoning abilities of LLMs and LLM agents. Notable examples include GLUE \citep{wangGLUEMultiTaskBenchmark2018}, MMLU \citep{hendrycksMeasuringMassiveMultitask2020}, BIG-bench \citep{srivastavaImitationGameQuantifying2022}, WebBench \citep{zhou2023webarena}, WebShop \citep{yao2022webshop}, and Mind2Web \citep{deng2024mind2web}. While measuring important aspects of general reasoning capabilities, they often do not quite capture the complexity and relevance of real-world tasks performed by businesses on a daily basis. Tasks like ``order a pizza to the following address'' \citep{deng2024mind2web}, ``find a product that matches this description'' \citep{yao2022webshop}, or ``find all the commits of person x'' \citep{zhou2023webarena} are complex, but do not necessarily translate directly into real-world impact and general problem-solving capabilities. %
Since the advent of LLM agents, the ability to perform web-based tasks has been recognized as a particularly high-impact capability, and has been the focus of several benchmarks \citep{zhou2023webarena, yao2022webshop, deng2024mind2web}. While these benchmarks succeed in assessing agents' abilities to carry out basic tasks in highly realistic (albeit highly restricted) web environments, however, they do not exhibit the key properties outlined above. Firstly, the tasks making up existing web-based benchmarks --- ``order a pizza to the following address'' \cite{deng2024mind2web}, ``find a product that matches this description'' \cite{yao2022webshop}, ``find all commits authored by person x'' \cite{zhou2023webarena} --- do not comprise a meaningful portion of any real-world job. Therefore, good performance on these benchmarks is only very weak evidence for the potential for transformative economic impact. Secondly, these benchmarks do not provide continuity: they are composed of tasks that are broadly similar in difficulty, and their partial credit measures are either difficult to interpret or non-existent.

Our aim is to develop a benchmark based on real-world open-web research tasks that, unlike existing benchmarks, fulfils the requirements outlined above. We focus on ``messy'' open-web research tasks such as compiling a list of AI labs that have trained large models, estimating the number of paying users for a product, and tracking down the original source of a claim. These tasks were directly inspired by requests we received from clients and from our own experience as a startup providing AI-augmented research and forecasting services. They were chosen to be representative of the kinds of real-world tasks that are routinely performed across a range of economically important roles (investment analysts, consultants, and forecasters, for example).

Creating a benchmark that is both realistic and based on open-web research tasks poses an interesting challenge, however. To ensure an appropriate distribution of difficulty across tasks, we have to assess how difficult each task is. This often requires effectively solving the task or at least making substantial progress against it. Similarly, assessing an agent's performance usually requires a detailed understanding of the task and possible solutions. This makes the tasks time-consuming to design, evaluate. At the same time, tasks are subject to constant change. The ideal approach, the correct answer, and the task's difficulty all fluctuate rapidly as the world changes and new information appears on the web.

How are we to maintain a benchmark with a carefully controlled difficulty distribution, that can remain in use for long enough so that we can be confident in extrapolation, without expending an unreasonable amount of time and effort constantly designing and evaluating new tasks?
Our proposed solution is as follows:
\begin{enumerate}
    \item Manually work through a number of real-world open-web research tasks at snapshots in time. This includes carefully recording what we observe on the web; the strategies, mechanics, and adaptations required to solve the task well; and the pitfalls encountered along the way.
    \item Then use these observations to construct simulated tasks that mimic the real thing as closely as possible.
\end{enumerate}

In this paper we focus on the first phase of this project. We describe a set of real-world tasks, outline our criteria for good performance and note the challenges and idiosyncrasies that we intend to replicate in future simulated tasks. We quantitatively and qualitatively evaluate a suite of LLMs and agent architectures against the real-world tasks, 
%allowing us to gauge difficulty and to test the extent to which performance is preserved in the simulations, and 
providing (to our knowledge) the first in-depth assessment of agents' abilities to conduct challenging analyst-style research on the real open web.

Section \ref{sec:methods} gives an overview of our methods, providing details on the tasks, the setup of LLMs and agent architectures we use, and the way we evaluate agent performance. Section \ref{sec:results} presents both numerical results as well as a detailed qualitative assessment of agent traces. In particular, Section \ref{sec:overall-performance} gives an overview of the overall performance of the different LLMs and agent architectures. Section \ref{sec:individual-tasks} provides details on the performance of agents on individual tasks. Section \ref{sec:taxonomy-of-failures} gives a more detailed overview of failure modes for the different LLMs and agent architectures. 

\section{Methods} \label{sec:methods}

\subsection{Tasks} \label{sec:task-selection}
The tasks we chose are derived from our experience as a startup providing AI-augmented research and forecasting services to clients in finance and consulting. The tasks cover the full range of domains in which we have carried out research. The construct we are measuring here --- the ability to carry out messy analytical tasks using the open web --- underpins a wide range of economically valuable jobs, and we expect that the ability to perform well on this benchmark should correspond to the ability to replace a significant amount of real-world white-collar labor, across a number of domains.
%--- we believe that the construct we are measuring here is rather general, such that a benchmark focused on a single narrow domain (e.g. geopolitical forecasting) runs the risk of understating the potential impact of agents who perform well on it.
As of now\footnote{ We intend this to be a living document, and plan to update it with new tasks as the scope of our research grows.}, our tasks encompass the following domains:
\begin{itemize}
\item Geopolitical forecasting
\item Financial forecasting
\item Epidemiological forecasting
\item Competitor analysis
\item Market sizing
\end{itemize}
To increase the chances that performance improves continuously, we design tasks spanning a wide range of difficulties, starting with tasks that are partially solved by weaker open-source models, and progressing to tasks on which even state-of-the-art models struggle to make meaningful progress.
Since the total number of tasks is small, we enhance continuity further by providing meaningful partial credit measures for each task, derived from our own solutions (see Section \ref{sec:evaluation} for more).

%To minimize the risk of client confidentiality issues, many of our tasks are not directly derived from research questions that we have addressed. We have designed them to be representative and plausible, however, and we explain their relevance briefly at the beginning of each task.

\subsection{Agents and LLMs} \label{sec:agent-selection}
An agent in the context of this work is understood as an agent architecture powered by a single underlying LLM. We analyze five different LLMs, classified as ``strong'' (\oone, \newclaude, \gpt, \llama) and ``cheap'' (\minigpt --- we dropped \minillama after initial experiments suggesting that it is too weak to support an agent flow). We combine these five LLMs with four different architectures, varying across the following dimensions:
\begin{itemize}
    \item The presence or absence of explicit planning
    \item The presence or absence of delegation to sub-agents
\end{itemize}
The canonical example of a non-planning, non-delegating agent architecture is ReAct \citep{yao2022react} (note that ReAct is free to plan in its \emph{reason} step, but since explicit planning is not a core aspect of the architecture, we define it as non-planning). 
The GPT-4-delegate architecture from \cite{METR2023} is an example of a non-planning, delegating architecture. AdaPlanner \citep{sun2024adaplanner} is an example of a planning, non-delegating architecture. AutoGPT is (arguably) an example of a planning, delegating architecture. For details on the architectures we use, see Section \ref{sec:architectures}.

Our agents are then defined by the combination of an LLM with an architecture (\ref{sec:architectures}). Note that we do not pair the cheap LLMs (\minigpt) with the more complicated architectures (i.e. those that plan or delegate) as they (empirically) cannot support those more complicated architectures. We label agents via their behavior followed by the LLM: For example, we label the \newclaude-powered, planning, non-delegating agent as \plan\newclaudelabel. Non-planning, non-delegating agents are referred to as \emph{ReAct} agents, e.g. \react\gptlabel.

% Suggest putting that into the discussion.
% Therefore, we evaluate a range of architectures. In an attempt to choose a representative subset of extant architectures, we identify two dimensions along which architectures vary, and use these to define our space of agents to evaluate.

\subsection{Architectures} \label{sec:architectures}
Here we give a brief overview of our implementations of our various architectures.

\subsubsection{Non-planning, non-delegating}
We implement our non-planning, non-delegating architecture as straightforward \emph{ReAct}, as described in \cite{yao2022react}, with a single example agent trace (constant across and unrelated to all tasks).
\subsubsection{Non-planning, delegating}
An agent based on this architecture consists of an orchestrator \emph{ReAct} agent architecture that delegates tasks to \emph{ReAct} sub-agents. The orchestrator has access to only a single tool --- \emph{delegate} --- which takes a task description as input and assigns this task to a sub-agent.
Sub-agents have access to the full set of tools. The delegate tool returns with the sub-agent's completion of the task, or a report of failure, together with a summary of what it did.
\subsubsection{Planning, non-delegating}
Agents based on this architecture maintain explicit, stateful plans (with collected information). Tool calls are chosen based on the current plan. The plan is revised at the end of each iteration, either to mark the attempted step as complete, or to adapt to a failure.

\subsubsection{Planning, delegating}
This architecture is similar to the planning, non-delegating architecture, but each step in the plan is delegated to a sub-agent. Rather than seeing the entire history of tool calls, the orchestrator agent sees only the current plan, the history of plan revisions, and the history of reports from sub-agents.

\subsection{Tools}

Agents are equipped with the set of tools outlined below. Notably, we do not currently provide ways to visually interact with a website (e.g. viewing a graph or clicking a button to show more), nor do we provide a way to access dynamically rendered content (e.g. access content ``hidden'' in infinite scrolling UIs, getting past captchas, etc.).
\subsubsection{Google}
We use \emph{Serper}\footnote{\url{https://serper.dev/}}, which returns the contents of the first page of Google for a given query, and parses the returned JSON object into something markdown-like.
\paragraph{Example input:} \texttt{current CEO of OpenAI}
\paragraph{Example output (truncated):} 
\begin{alltt}
\justify
Google results for the query `current CEO of OpenAI`.:
# answerBox:
    title: OpenAI / CEO
    answer: Sam Altman
# organic:
    title: Sam Altman is officially reinstated as OpenAI's CEO just two weeks ...
    link: https://fortune.com/2023/11/29/sam-altman-officially-reinstated-openai-ceo-board-bret-taylor/
    snippet: OpenAI said that Sam Altman was officially reinstated as chief executive officer of the company and it has a new initial three-member board ...
    date: Nov 29, 2023
    position: 1
    title: Sam Altman - Wikipedia
    link: https://en.wikipedia.org/wiki/Sam_Altman
    snippet: Samuel Harris Altman (born April 22, 1985) is an American entrepreneur and investor best known as the CEO of OpenAI since 2019 He is also CEO of AltC ...
    position: 2
    ...
    title: Meet Sam Altman: OpenAI Cofounder and CEO, Doomsday Prepper
    link: https://www.businessinsider.com/sam-altman-chatgpt-openai-ceo-career-net-worth-ycombinator-prepper-2023-1
    snippet: OpenAI CEO Sam Altman is now a billionaire thanks to his investments, Forbes recently reported.
    date: Apr 8, 2024
    position: 8
    title: Who is Emmett Shear, the new CEO of OpenAI? - Washington Post
    link: https://www.washingtonpost.com/technology/2023/11/20/emmett-shear-openai-ceo/
    snippet: Shear was named interim CEO of OpenAI to replace Altman and Mira Murati, who held the interim CEO title for just two days. Shear co-founded the ...
    date: Nov 20, 2023
    position: 9
\end{alltt}

\subsubsection{PythonREPL}
Provides access to a PythonREPL (Read-Eval-Print-Loop; an interactive programming environment that allows you to execute Python code one line or expression at a time). State is maintained between tool calls, so that the agent can interactively explore data and debug code.
\subsubsection{Query link text}
This tool expects a URL to a website or a \texttt{.pdf} file and will return a specified number of excerpts (HTML turned to markdown, unless the content is part of a table, because many websites contain a lot of HTML just for styling purposes) most likely to contain the answer to a specified question.
\paragraph{Example input:}
\begin{alltt}
\justify
\{
  "url": "\url{https://en.wikipedia.org/wiki/OpenAI}",
  "query": "Who are the most influential people at OpenAI today?",
  "num_excerpts": 6
\}
\end{alltt}
\paragraph{Example output (truncated):}
\begin{alltt}
\justify
Extracted excerpts for query 'most influential people at OpenAI today' page \url{https://en.wikipedia.org/wiki/OpenAI}.:
### Board of Directors of the OpenAI
nonprofit[95][114][[edit](/w/index.php?title=OpenAI&action=edit&section=8
"Edit section: Board of Directors of the OpenAI nonprofit[95][114]")]

    * [Microsoft](/wiki/Microsoft "Microsoft") (observer)
    * [Bret Taylor](/wiki/Bret_Taylor "Bret Taylor") (chairman)
    * [Sam Altman](/wiki/Sam_Altman "Sam Altman")
    * [Lawrence Summers](/wiki/Lawrence_Summers "Lawrence Summers")
    * [Adam D'Angelo](/wiki/Adam_D%27Angelo "Adam D'Angelo")
    * [Sue Desmond-Hellmann](/wiki/Sue_Desmond-Hellmann "Sue Desmond-Hellmann")
    * [Nicole Seligman](/wiki/Nicole_Seligman "Nicole Seligman")
    * [Fidji Simo](/wiki/Fidji_Simo "Fidji Simo")

### Principal individual
investors[112][[edit](/w/index.php?title=OpenAI&action=edit&section=9 "Edit
section: Principal individual investors[112]")]...

...

...In December 2015, Sam Altman, Greg Brockman, [Reid Hoffman](/wiki/Reid_Hoffman
"Reid Hoffman"), [Jessica Livingston](/wiki/Jessica_Livingston "Jessica
Livingston"), [Peter Thiel](/wiki/Peter_Thiel "Peter Thiel"), [Elon
Musk](/wiki/Elon_Musk "Elon Musk"), [Amazon Web
Services](/wiki/Amazon_Web_Services "Amazon Web Services") (AWS),
[Infosys](/wiki/Infosys "Infosys"), and [YC Research](/wiki/YC_research "YC
research") announced[21] the formation of OpenAI and pledged over $1 billion
to the venture. The actual collected total amount of contributions was only
$130 million until 2019.[12] According to an investigation led by
[TechCrunch](/wiki/TechCrunch "TechCrunch"), Musk was its largest donor while
YC Research did not contribute anything at all.[22] The organization stated it
would "freely collaborate" with other institutions and researchers by making
its patents and research open to the public.[23][24] OpenAI is headquartered...
\end{alltt}

\subsection{Evaluation} \label{sec:evaluation}

For each task and agent, we evaluate a the output of a single agent run.
In our evaluations we not only score whether an agents has solved a task, but also assign partial credit to incomplete solutions. This is possible because the tasks we selected only allow for a relatively small number of plausible solution strategies. For each task, we enumerate viable strategies. For each strategy, we define criteria for an ideal agent trace, such that the ideal trace would satisfy \emph{all} of these criteria. Solutions in which more of the criteria are satisfied are preferred and we assign partial credit accordingly.

In order to compare different LLMs we score them as follows:
For a given strategy $S$ with $n$ criteria for partial credit $\{x_i\}_{i=1}^n$, each $x_i$ takes a value in (``Success'', ``Partial Success'', ``Failure'', or ``NA''). The first three labels are self-explanatory. The label ``NA'' is used when an agent did not have a chance to succeed on that criterion, as it failed a to complete the prerequisites.
We then compute the score for a strategy $S$ as
\begin{equation}
    \text{Score}_S = \frac{|\{x_i:x_i=\text{Success}\}| + {1\over3}|\{x_i:x_i=\text{Partial Success}\}|}{n}.
\end{equation}
Defining the end-to-end performance score as
\begin{equation}
    \text{Score}_\text{end-to-end}=\begin{cases}
        1 &\text{if overall completion was successful,}\\
        {1\over3}&\text{if overall completion was partially successful,}\\
        0&\text{otherwise.}
    \end{cases},
\end{equation}
we finally compute the total score as the mean of the top-performing strategy's score and its overall end-to-end performance score, i.e. 
\begin{equation}
    \text{Score}_\text{final} = {1\over2}\max_S \text{Score}_S + {1\over2}\text{Score}_\text{end-to-end}.
\end{equation}
This way we end up assigning a score of 0 if an agent does not make any progress at all. The agent receives a score of 0.5 if it successfully completes all subtasks for at least one strategy, but failed to produce a final answer by combining intermediate results. An agent receives a score of 1 if it succeeds with at least one strategy and successfully puts intermediate results together to produce a good final response. 
% See Table~\ref{tab:scoring_example} for an example.

\

Alongside performance against our partial-credit criteria, we make qualitative observations on agent performance. With these, we aim to give a richer sense of how agents are performing, highlight issues whose resolution could unlock good performance, and flag cases where deficiencies in our implementation are to blame for failures (we try to minimize cases like this, but we encountered a long tail of tweaks that could fix specific failures, and, as of yet, have not had time to implement them all).

\

For every task and agent, we evaluated a single agent run. We set every LLM's temperature (except \oone, which does not support this setting) to 0, but our experience of working with LLM agents is that there may be a decent amount of variation from run to run. Some of the expected variance is due to LLMs not being fully deterministic even at temperature 0, some is due to Google searches not being static over time, and most is due to small differences in inputs resulting in vastly different agent behavior. Thus one should not necessarily read too much into an agent's performance on any given single task, but look at the aggregate score over several tasks.

\

\section{Results} \label{sec:results}
This section presents the numerical scores as well as a qualitative assessment of the performance of each agent. We first present overall results aggregated across tasks, then show performance on individual tasks, and lastly provide a detailed taxonomy of different failure modes we observed across LLMs and agent architectures. 
For each task, we present a table detailing performance of each agent against the task's partial-credit criteria. Successes are marked green, partial successes are marked yellow, failures are marked red, and NAs appear as blank cells.

\subsection{Overall performance of LLMs and architectures} \label{sec:overall-performance}
\begin{figure}[!htbp]
    \includegraphics[width=\textwidth]{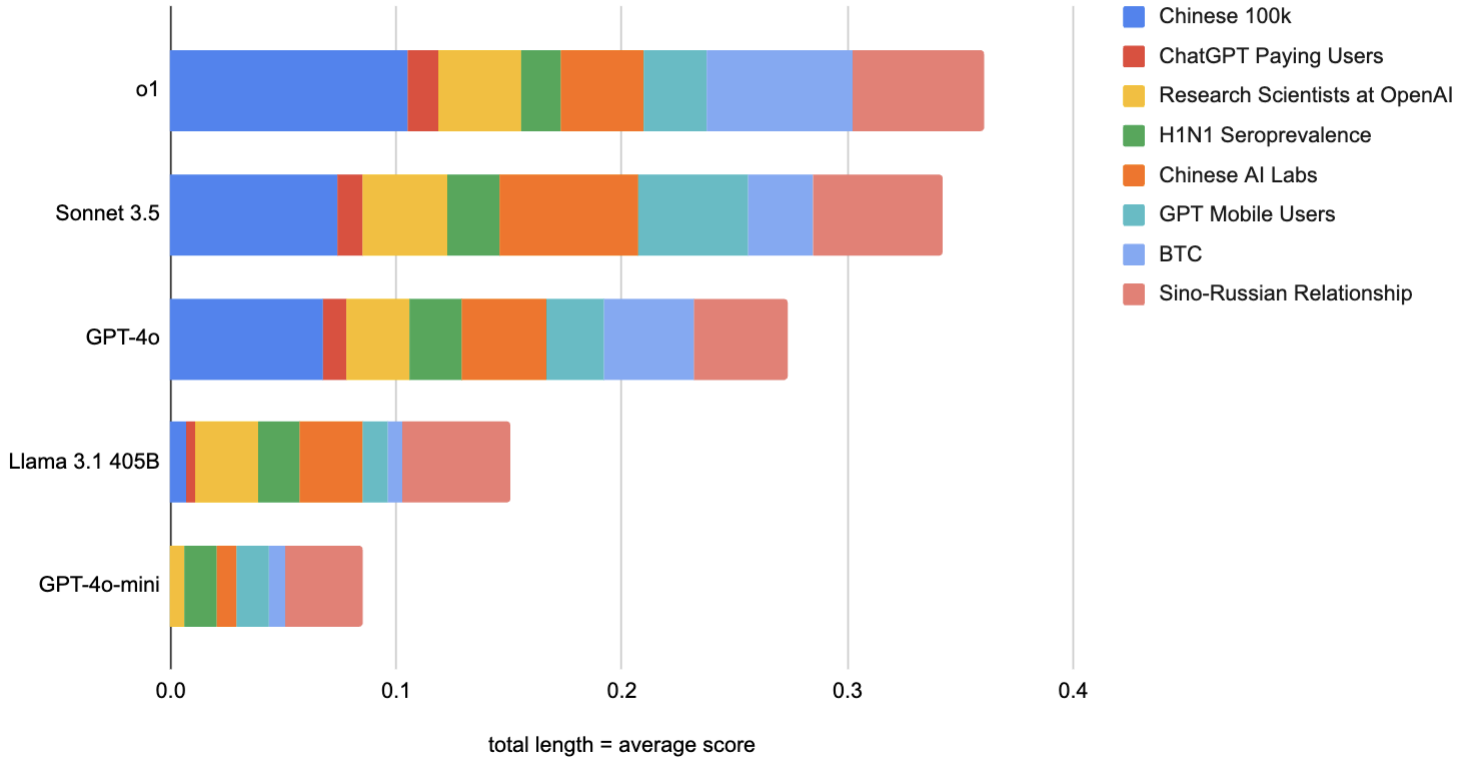}
    \caption{\label{fig:comparison_of_LLMs}Performance broken down by LLM and task. For each LLM and each task, we (post-hoc) chose the architecture that performed the best and recorded its score. See Table \ref{tab:comparison_of_LLMs} for the exact values.}
\end{figure}

\begin{figure}[!htbp]
    \includegraphics[width=\textwidth]{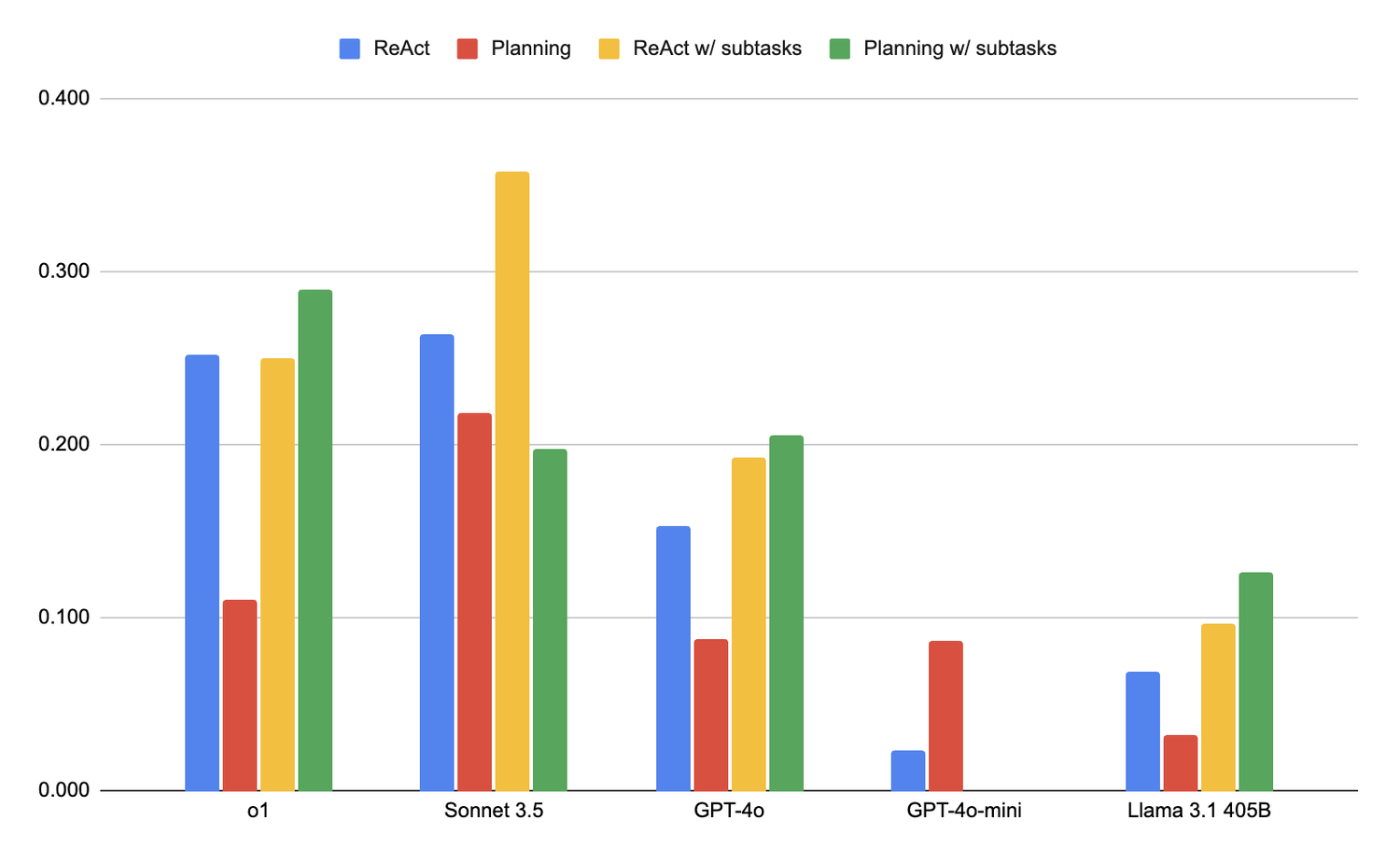}
    \caption{\label{fig:comparison_of_agents}Scores broken down by agents (i.e. architecture-LLM combinations). See Table \ref{tab:comparison_of_agents} for the exact values.}
\end{figure}

There are many metrics according to which we may declare one LLM to be better than another. A natural one might be derived from the following approach: For a given LLM, run all architectures and (assuming a little bit of oversight) pick the one that performed the best. See Figure \ref{fig:comparison_of_LLMs} for such a comparison. \oone eked out a victory over \newclaude. This is mostly due to \oone doing exceptionally well at some tasks while being distinctly average otherwise. \newclaude, in contrast, performs fairly well much more consistently, which can also be seen from \newclaude outperforming \oone on average (average taken over all tasks and architectures, see Figure \ref{fig:comparison_of_agents}). Both easily beat \gpt, which is not much worse according to these metrics, but feels noticeably worse when looking through its agent traces. \llama usually fails to follow instructions and ends up being significantly worse. \minigpt, the only model in the ``cheap'' category of LLMs here, unsurprisingly takes the last place. For most tasks it does little more than googling the first thing that comes to mind.

We turn now to a fine-grained comparison between agent-architecture pairs --- See Figure \ref{fig:comparison_of_agents}. Perhaps surprisingly, \newclaude with a non-planning, but delegating architecture is the clear winner. Again, this is due to \newclaude being much more consistent than the highly variable \oone, where each architecture (apart from the planning, but non-delegating one) seems to do very well occasionally, but not consistently.

We also note that the planning and delegating architecture working better for \oone than for \newclaude is unlikely to be a fluke: \oone is a lot more verbose (in a good way) and this seems to prevent it from failing to specify subtasks that make sense--a problem that all delegating agents using different LLMs suffer (more) from.

It also (weakly) suggests that delegating architectures tend to outperform non-delegating architectures, but we are not certain this would replicate. The simple ReAct or planning architectures seemed to outperform their delegating counterparts on some tasks that did not lend themselves to parallelisation as much and occasionally a delegating architecture merely solved a task by repeatedly tasking a ReAct agent with its task.

% In this section we present our scoring rubric for each task, and show the performance of our agents against these rubrics.
% We also explain the real-world relevance of the tasks, which in some cases is non-obvious, since many are sub-tasks that arise in the context of larger, more-obviously-valuable tasks.
% Finally, we give high-level qualitative observations on aspects of agent performance and behavior that are not captured by the rubrics.

% For more detail, see Appendix \ref{sec:rubric-derivation}, where we give detailed breakdowns of our own solutions (from which the rubrics are derived), and Appendix \ref{sec:detailed-performance-remarks}, where we give finer-grained qualitative observations on agent performance and behavior.

\subsection{Performance on individual tasks} \label{sec:individual-tasks}

\subsubsection{Compile a list of Chinese AI labs that have carried out training runs using over 1e24 FLOPs}
\paragraph{Exact prompt:}
\texttt{\justify Which Chinese AI labs have carried out training runs using over 1e24 FLOPs? Reply with an exhaustive list of names of AI labs and nothing else.}
\paragraph{Relevance}

\

\noindent US-China competition over AI is an important geopolitical issue, and an understanding of how this is likely to play out will be of value to many firms.
An assessment of China's current capabilities is a key input to forecasts in this space.

\paragraph{Scoring rubric}

\

\

\noindent Strategy 1 (Epoch AI data):
          \begin{itemize}
              \item \textbf{page} Found the page of Epoch\ AI\footnote{https://epochai.org/}, a research organization which hosts a dataset with the necessary information
              \item \textbf{download} Downloaded the dataset
              \item \textbf{answer} Computed the correct answer
          \end{itemize}
OR
\

\

\noindent Strategy 2: (Manually compile list):
          \begin{itemize}
              \item \textbf{models} Compiled a list of "big" models
              \item \textbf{flops} Found FLOPs estimates for each model
          \end{itemize}
\

\paragraph{Performance}
\

\noindent See Table \ref{tab:combined_task_chinese_labs_evals} for results.

\

\begin{table}[!htbp]
    \centering
    \setlength{\tabcolsep}{4pt}
    \begin{tabular}{lp{0.4cm}*{3}{>{\centering\arraybackslash}p{0.8cm}}p{0.4cm}*{2}{>{\centering\arraybackslash}p{0.8cm}}p{0.4cm}p{0.8cm}}
        \toprule
        & & \multicolumn{3}{c}{\makecell[c]{Strategy 1:\\ Epoch AI Data}} & & \multicolumn{2}{c}{\makecell[c]{Strategy 2:\\ Compile list}} & & \\
        \cmidrule(lr){3-5} \cmidrule(lr){7-8}
        & & \rotatebox{45}{page} & \rotatebox{45}{download} & \rotatebox{45}{answer} & & \rotatebox{45}{models} & \rotatebox{45}{flops} & & \rotatebox{45}{\textbf{final}} \\
        \midrule
        \react\oonelabel        & & \NAbox & \NAbox & \NAbox & & \NAbox & \NAbox & & \failurebox \\
        \plan\oonelabel         & & \NAbox & \NAbox & \NAbox & & \NAbox & \NAbox & & \failurebox \\
        \delegate\oonelabel     & & \NAbox & \NAbox & \NAbox & & \partialsuccessbox & \partialsuccessbox & & \partialsuccessbox \\
        \plandelegate\oonelabel & & \NAbox & \NAbox & \NAbox & & \NAbox & \NAbox & & \failurebox \\
        \addlinespace
        \midrule
        \react\newclaudelabel        & & \successbox        & \failurebox & \NAbox      & & \NAbox & \NAbox & & \failurebox \\
        \plan\newclaudelabel         & & \successbox & \failurebox & \NAbox & & \NAbox & \NAbox & & \failurebox \\
        \delegate\newclaudelabel     & & \successbox & \successbox & \partialsuccessbox & & \NAbox & \NAbox & & \partialsuccessbox \\
        \plandelegate\newclaudelabel & & \failurebox & \NAbox & \NAbox & & \partialsuccessbox & \partialsuccessbox & & \failurebox \\
        \addlinespace
        \midrule
        \react\gptlabel              & & \failurebox        & \successbox & \failurebox & & \failurebox & \NAbox & & \failurebox \\
        \plan\gptlabel               & & \failurebox & \NAbox & \NAbox & & \partialsuccessbox & \failurebox & & \failurebox \\
        \delegate\gptlabel           & & \partialsuccessbox & \partialsuccessbox & \NAbox & & \partialsuccessbox & \partialsuccessbox & & \failurebox \\
        \plandelegate\gptlabel       & & \partialsuccessbox & \partialsuccessbox & \NAbox & & \partialsuccessbox & \successbox & & \failurebox \\
        \addlinespace
        \midrule
        \react\llamalabel            & & \partialsuccessbox & \NAbox      & \NAbox      & & \NAbox & \NAbox & & \failurebox \\
        \plan\llamalabel             & & \failurebox & \NAbox & \NAbox & & \NAbox & \NAbox & & \failurebox \\
        \delegate\llamalabel         & & \failurebox & \NAbox & \NAbox & & \partialsuccessbox & \NAbox & & \failurebox \\
        \plandelegate\llamalabel     & & \failurebox & \NAbox & \NAbox & & \successbox & \NAbox & & \failurebox \\
        \addlinespace
        \midrule
        \react\minigptlabel          & & \failurebox        & \NAbox      & \NAbox      & & \failurebox & \NAbox & & \failurebox \\
        \plan\minigptlabel           & & \failurebox & \NAbox & \NAbox & & \partialsuccessbox & \NAbox & & \failurebox \\
        \bottomrule
    \end{tabular}
    \caption{\label{tab:combined_task_chinese_labs_evals}Overview of how agents performed on the task of finding all Chinese AI labs that ran training runs using over 1e24 FLOPs. Strategy 1: Agents that tried a strategy based on Epoch AI data. Strategy 2: Agents that tried a strategy based on manually compiling a list of models.}
\end{table}

\subsubsection{Find original sources shedding light on the Sino-Russian relationship after their announcement of ``friendship without limits''.}

\paragraph{Exact prompt:} 
\texttt{\justify We say a source is original if its relevant information cannot be traced back to another URL that it links to or got its information from.
So, for example, a blog post referencing a recent Ukrainian advance would not be "original", but the Reuters article announcing it (and whatever website the blog author likely got it from–directly or indirectly) would be. However, if Reuters discussed the US announcing something, the original source would likely point to an announcement on a US government website, not Reuters.
Given this definition, your task is to find original sources for actions, treaties, deals, ... (/lack thereof; not just inconsequential verbal assurances) showing how good or bad the Sino-Russian relationship is after their announcement of "friendship without limits". Your final response should be a list of links to original sources with a single sentence of explanation of its relevance/ramification each.}

\paragraph{Relevance}

\

\noindent An important component of forecasting is understanding how seriously to take the signals we receive. Rather than taking stories that appear in the media at face value, it is often necessary to understand the context behind them.
Especially for ideologically charged stories, a full understanding of context, and of the basis upon which claims appearing in the media are being made, requires digging into original sources.
\paragraph{Scoring rubric}
\begin{itemize}
    \item [Preparation:]
    \begin{itemize}
        \item \textbf{date} looked up/remembered when Russia and China first announced their ``friendship without limits''
        \item \textbf{plan} came up with a structured plan or a list of areas to look into (as opposed to aimless googling)
    \end{itemize}
    \item [Data collection:]
          \begin{itemize}
              \item \textbf{search} attempted searches in Russian and Chinese
              \item \textbf{wiki} found \url{https://en.wikipedia.org/wiki/China%E2%80%93Russia_relations}
              \item \textbf{kremlin} found Kremlin pages for press releases
          \end{itemize}
    \item [Performance:]
          \begin{itemize}
              \item \textbf{UN} found China's abstention at the UN on Russo-Ukrainian questions
              \item \textbf{trade} found trade data, noting significant growth in Sino-Russian trade
              \item \textbf{drills} found joint naval drills
              \item \textbf{one-china} found Russia's support of the One China principle
              \item \textbf{military} found evidence of exports of Chinese hardware for military purposes
              \item \textbf{no-75-deals} found lack of significant agreements for 75th anniversary of diplomatic relations
              \item \textbf{other} found other things not accounted for in this list
              \item \textbf{original} tracked down original sources
          \end{itemize}
\end{itemize}

\paragraph{Performance}
\

\noindent See Table \ref{tab:friendshipwithoutlimits}.

\begin{table}[!htbp]
    \centering
    \setlength{\tabcolsep}{2pt} 
    \begin{tabular}{l>{\centering\arraybackslash}p{0.2cm}*{13}{>{\centering\arraybackslash}p{0.6cm}}>{\centering\arraybackslash}p{0.2cm}>{\centering\arraybackslash}p{0.6cm}}
        \toprule
        & & \rot{date} & \rot{plan} & \rot{search} & \rot{wikipedia} & \rot{kremlin} & \rot{UN} & \rot{trade} & \rot{drills} & \rot{one-china} & \rot{military} & \rot{\makecell{no-75-deals}} & \rot{other} & \rot{original} & & \rot{\textbf{final}} \\
        \midrule
        \react\oonelabel & & \failureboxsmall & \failureboxsmall & \failureboxsmall & \partialsuccessboxsmall & \failureboxsmall & \failureboxsmall & \partialsuccessboxsmall & \successboxsmall & \successboxsmall & \failureboxsmall & \failureboxsmall & \failureboxsmall & \successboxsmall & & \failureboxsmall \\
        \plan\oonelabel & & \successboxsmall & \failureboxsmall & \NAboxsmall & \NAboxsmall & \NAboxsmall & \failureboxsmall & \partialsuccessboxsmall & \partialsuccessboxsmall & \failureboxsmall & \failureboxsmall & \failureboxsmall & \successboxsmall & \failureboxsmall & & \partialsuccessboxsmall \\
        \delegate\oonelabel & & \NAboxsmall & \successboxsmall & \failureboxsmall & \failureboxsmall & \partialsuccessboxsmall & \failureboxsmall & \successboxsmall & \successboxsmall & \successboxsmall & \failureboxsmall & \failureboxsmall & \successboxsmall & \partialsuccessboxsmall & & \partialsuccessboxsmall \\
        \plandelegate\oonelabel & & \successboxsmall & \successboxsmall & \partialsuccessboxsmall & \failureboxsmall & \successboxsmall & \successboxsmall & \successboxsmall & \successboxsmall & \successboxsmall & \failureboxsmall & \failureboxsmall & \successboxsmall & \successboxsmall & & \partialsuccessboxsmall \\
        \addlinespace
        \midrule
        \react\newclaudelabel & & \successboxsmall & \failureboxsmall & \failureboxsmall & \failureboxsmall & \failureboxsmall & \failureboxsmall & \successboxsmall & \successboxsmall & \successboxsmall & \failureboxsmall & \successboxsmall & \successboxsmall & \failureboxsmall & & \partialsuccessboxsmall \\
        \plan\newclaudelabel & & \successboxsmall & \successboxsmall & \failureboxsmall & \failureboxsmall & \failureboxsmall & \failureboxsmall & \partialsuccessboxsmall & \partialsuccessboxsmall & \successboxsmall & \failureboxsmall & \failureboxsmall & \successboxsmall & \partialsuccessboxsmall & & \partialsuccessboxsmall \\
        \delegate\newclaudelabel & & \successboxsmall & \successboxsmall & \failureboxsmall & \failureboxsmall & \partialsuccessboxsmall & \partialsuccessboxsmall & \successboxsmall & \successboxsmall & \successboxsmall & \successboxsmall & \successboxsmall & \successboxsmall & \successboxsmall & & \partialsuccessboxsmall \\ 
        \plandelegate\newclaudelabel & & \failureboxsmall & \successboxsmall & \successboxsmall & \failureboxsmall & \failureboxsmall & \partialsuccessboxsmall & \successboxsmall & \successboxsmall & \successboxsmall & \successboxsmall & \partialsuccessboxsmall & \successboxsmall & \failureboxsmall & & \partialsuccessboxsmall \\ 
        \addlinespace
        \midrule
        \react\gptlabel & & \failureboxsmall & \failureboxsmall & \failureboxsmall & \failureboxsmall & \failureboxsmall & \failureboxsmall & \failureboxsmall & \failureboxsmall & \failureboxsmall & \failureboxsmall & \failureboxsmall & \failureboxsmall & \failureboxsmall & & \failureboxsmall \\
        \plan\gptlabel & & \failureboxsmall & \partialsuccessboxsmall & \failureboxsmall & \failureboxsmall & \failureboxsmall & \failureboxsmall & \partialsuccessboxsmall & \successboxsmall & \failureboxsmall & \failureboxsmall & \failureboxsmall & \successboxsmall & \partialsuccessboxsmall & & \partialsuccessboxsmall \\
        \delegate\gptlabel & & \failureboxsmall & \partialsuccessboxsmall & \failureboxsmall & \failureboxsmall & \failureboxsmall & \failureboxsmall & \failureboxsmall & \failureboxsmall & \successboxsmall & \failureboxsmall & \failureboxsmall & \successboxsmall & \successboxsmall & & \partialsuccessboxsmall \\
        \plandelegate\gptlabel & & \failureboxsmall & \successboxsmall & \failureboxsmall & \failureboxsmall & \failureboxsmall & \partialsuccessboxsmall & \successboxsmall & \successboxsmall & \successboxsmall & \successboxsmall & \failureboxsmall & \failureboxsmall & \failureboxsmall & & \partialsuccessboxsmall \\
        \addlinespace
        \midrule
        \react\llamalabel & & \failureboxsmall & \failureboxsmall & \failureboxsmall & \failureboxsmall & \failureboxsmall & \failureboxsmall & \failureboxsmall & \failureboxsmall & \failureboxsmall & \failureboxsmall & \failureboxsmall & \failureboxsmall & \failureboxsmall & & \failureboxsmall \\
        \plan\llamalabel & & \failureboxsmall & \failureboxsmall & \failureboxsmall & \failureboxsmall & \failureboxsmall & \failureboxsmall & \failureboxsmall & \failureboxsmall & \failureboxsmall & \failureboxsmall & \failureboxsmall & \failureboxsmall & \failureboxsmall & & \failureboxsmall \\
        \delegate\llamalabel & & \successboxsmall & \successboxsmall & \failureboxsmall & \failureboxsmall & \failureboxsmall & \failureboxsmall & \successboxsmall & \successboxsmall & \successboxsmall & \failureboxsmall & \failureboxsmall & \successboxsmall & \successboxsmall & & \partialsuccessboxsmall \\
        \plandelegate\llamalabel & & \failureboxsmall & \successboxsmall & \failureboxsmall & \failureboxsmall & \failureboxsmall & \failureboxsmall & \successboxsmall & \successboxsmall & \partialsuccessboxsmall & \failureboxsmall & \failureboxsmall & \failureboxsmall & \partialsuccessboxsmall & & \partialsuccessboxsmall \\
        \addlinespace
        \midrule
        \react\minigptlabel & & \failureboxsmall & \failureboxsmall & \failureboxsmall & \failureboxsmall & \failureboxsmall & \failureboxsmall & \failureboxsmall & \failureboxsmall & \failureboxsmall & \failureboxsmall & \failureboxsmall & \failureboxsmall & \failureboxsmall & & \failureboxsmall \\
        \plan\minigptlabel & & \successboxsmall & \failureboxsmall & \failureboxsmall & \failureboxsmall & \partialsuccessboxsmall & \failureboxsmall & \partialsuccessboxsmall & \successboxsmall & \failureboxsmall & \failureboxsmall & \failureboxsmall & \successboxsmall & \failureboxsmall & & \partialsuccessboxsmall \\
        \addlinespace
        \midrule
        \react\minillamalabel & & \failureboxsmall & \failureboxsmall & \failureboxsmall & \failureboxsmall & \failureboxsmall & \failureboxsmall & \failureboxsmall & \failureboxsmall & \failureboxsmall & \failureboxsmall & \failureboxsmall & \failureboxsmall & \failureboxsmall & & \failureboxsmall \\
        \plan\minillamalabel & & \failureboxsmall & \failureboxsmall & \failureboxsmall & \failureboxsmall & \failureboxsmall & \failureboxsmall & \failureboxsmall & \failureboxsmall & \failureboxsmall & \failureboxsmall & \failureboxsmall & \failureboxsmall & \failureboxsmall & & \failureboxsmall \\
        \bottomrule
    \end{tabular}
    \caption{\label{tab:friendshipwithoutlimits}Overview of how agents performed on the task of finding primary sources shedding light on the Sino-Russian relationship after their announcement of their ``friendship without limits''.}
\end{table}

\subsubsection{Find forecasts helpful for predicting whether BTC will trade at over USD90k before 2025}
\paragraph{Exact prompt:} \texttt{\justify Find reliable related probabilistic forecasts for the forecasting question "Will BTC trade at >90k before 2025?".}

\paragraph{Relevance}

\

\noindent If forecasts related to the question at hand are available from sources with good track records, forecasters will often incorporate them into their predictions.

\paragraph{Scoring rubric}
\begin{itemize}
    \item [Data collection:]
          \begin{itemize}
            \item \textbf{option:} attempted to search option markets
              \item \textbf{prediction:} attempted to search prediction markets (e.g. Manifold)
              \item \textbf{finance-sites:} attempted to search finance websites
              \item \textbf{aggregator:} attempted to search prediction aggregators (e.g. Metaculus)
          \end{itemize}
    \item [Performance:]
          \begin{itemize}
              \item \textbf{exclude-low-qual:} excluded/flagged low-quality and biased forecasts
              \item \textbf{exclude-outdated:} excluded/flagged outdated forecasts
              \item \textbf{before-2025:} did not confuse forecasts for before 2025 with forecasts for other dates
              \item \textbf{found-option:} found relevant option prices
              \item \textbf{found-prediction:} found relevant questions on prediction markets and/or prediction aggregators
              \item \textbf{found-sources:} found good sources
              \item \textbf{calculations:} performed correct calculations
          \end{itemize}
\end{itemize}
\paragraph{Performance}
\

\noindent See Table \ref{tab:btcforecasts}.

\begin{table}[!htbp]
    \centering
    \setlength{\tabcolsep}{2pt}  % Adjust this value to control space between cells
    \begin{tabular}{l>{\centering\arraybackslash}p{0.2cm}*{11}{>{\centering\arraybackslash}p{0.6cm}}>{\centering\arraybackslash}p{0.2cm}>{\centering\arraybackslash}p{0.6cm}}
        \toprule
        & & \rot{option} & \rot{prediction} & \rot{finance-sites} & \rot{aggregator} & \rot{exclude-low-qual} & \rot{exclude-outdated} & \rot{before-2025} & \rot{found-option} & \rot{found-prediction} & \rot{found-sources} & \rot{calculations} & & \rot{\textbf{final}} \\
        \midrule
        \react\oonelabel & & \successboxsmall & \successboxsmall & \successboxsmall & \successboxsmall & \successboxsmall & \successboxsmall & \successboxsmall & \successboxsmall & \failureboxsmall & \failureboxsmall & \successboxsmall & & \partialsuccessboxsmall \\
        \plan\oonelabel & & \failureboxsmall & \failureboxsmall & \successboxsmall & \successboxsmall & \NAboxsmall & \NAboxsmall & \NAboxsmall & \NAboxsmall & \NAboxsmall & \NAboxsmall & \NAboxsmall & & \failureboxsmall \\
        \delegate\oonelabel & & \failureboxsmall & \successboxsmall & \successboxsmall & \successboxsmall & \partialsuccessboxsmall & \successboxsmall & \successboxsmall & \failureboxsmall & \successboxsmall & \failureboxsmall & \NAboxsmall & & \partialsuccessboxsmall \\
        \plandelegate\oonelabel & & \failureboxsmall & \successboxsmall & \successboxsmall & \successboxsmall & \failureboxsmall & \failureboxsmall & \successboxsmall & \failureboxsmall & \successboxsmall & \failureboxsmall & \partialsuccessboxsmall & & \failureboxsmall \\
        \addlinespace
        \midrule
        \react\newclaudelabel & & \failureboxsmall & \successboxsmall & \successboxsmall & \successboxsmall & \partialsuccessboxsmall & \successboxsmall & \successboxsmall & \failureboxsmall & \successboxsmall & \failureboxsmall & \failureboxsmall & & \failureboxsmall \\
        \plan\newclaudelabel & & \failureboxsmall & \failureboxsmall & \partialsuccessboxsmall & \failureboxsmall & \failureboxsmall & \failureboxsmall & \failureboxsmall & \failureboxsmall & \failureboxsmall & \failureboxsmall & \failureboxsmall & & \failureboxsmall \\
        \delegate\newclaudelabel & & \failureboxsmall & \successboxsmall & \successboxsmall & \successboxsmall & \NAboxsmall & \NAboxsmall & \successboxsmall & \failureboxsmall & \successboxsmall & \failureboxsmall & \successboxsmall & & \failureboxsmall \\
        \plandelegate\newclaudelabel & & \failureboxsmall & \failureboxsmall & \failureboxsmall & \failureboxsmall & \failureboxsmall & \failureboxsmall & \failureboxsmall & \failureboxsmall & \failureboxsmall & \failureboxsmall & \partialsuccessboxsmall & & \failureboxsmall \\ 
        \addlinespace
        \midrule
        \react\gptlabel & & \failureboxsmall & \failureboxsmall & \successboxsmall & \partialsuccessboxsmall & \successboxsmall & \successboxsmall & \partialsuccessboxsmall & \failureboxsmall & \successboxsmall & \failureboxsmall & \failureboxsmall & & \partialsuccessboxsmall \\
        \plan\gptlabel & & \failureboxsmall & \failureboxsmall & \failureboxsmall & \failureboxsmall & \failureboxsmall & \failureboxsmall & \successboxsmall & \failureboxsmall & \failureboxsmall & \failureboxsmall & \failureboxsmall & & \failureboxsmall \\
        \delegate\gptlabel & & \failureboxsmall & \successboxsmall & \successboxsmall & \successboxsmall & \failureboxsmall & \failureboxsmall & \failureboxsmall & \failureboxsmall & \successboxsmall & \failureboxsmall & \failureboxsmall & & \failureboxsmall \\
        \plandelegate\gptlabel & & \failureboxsmall & \failureboxsmall & \successboxsmall & \failureboxsmall & \failureboxsmall & \failureboxsmall & \failureboxsmall & \failureboxsmall & \failureboxsmall & \failureboxsmall & \failureboxsmall & & \failureboxsmall \\
        \addlinespace
        \midrule
        \react\llamalabel & & \failureboxsmall & \failureboxsmall & \failureboxsmall & \failureboxsmall & \failureboxsmall & \failureboxsmall & \failureboxsmall & \failureboxsmall & \failureboxsmall & \failureboxsmall & \failureboxsmall & & \failureboxsmall \\
        \plan\llamalabel & & \failureboxsmall & \failureboxsmall & \successboxsmall & \failureboxsmall & \failureboxsmall & \failureboxsmall & \failureboxsmall & \failureboxsmall & \failureboxsmall & \failureboxsmall & \failureboxsmall & & \failureboxsmall \\
        \delegate\llamalabel & & \failureboxsmall & \failureboxsmall & \successboxsmall & \failureboxsmall & \failureboxsmall & \failureboxsmall & \failureboxsmall & \failureboxsmall & \failureboxsmall & \failureboxsmall & \partialsuccessboxsmall & & \failureboxsmall \\
        \plandelegate\llamalabel & & \failureboxsmall & \failureboxsmall & \failureboxsmall & \failureboxsmall & \failureboxsmall & \failureboxsmall & \NAboxsmall & \failureboxsmall & \NAboxsmall & \failureboxsmall & \failureboxsmall & & \failureboxsmall \\
        \addlinespace
        \midrule
        \react\minigptlabel & & \failureboxsmall & \failureboxsmall & \partialsuccessboxsmall & \failureboxsmall & \failureboxsmall & \failureboxsmall & \failureboxsmall & \failureboxsmall & \failureboxsmall & \failureboxsmall & \failureboxsmall & & \failureboxsmall \\
        \plan\minigptlabel & & \failureboxsmall & \failureboxsmall & \partialsuccessboxsmall & \partialsuccessboxsmall & \failureboxsmall & \failureboxsmall & \failureboxsmall & \failureboxsmall & \successboxsmall & \failureboxsmall & \failureboxsmall & & \failureboxsmall \\
        \bottomrule
    \end{tabular}
    \caption{\label{tab:btcforecasts}Overview of how agents performed on the task of finding forecasts related to BTC prices.}
\end{table}

\subsubsection{Use official government data to estimate how many Chinese have an annual disposable income exceeding 100,000 Yuan.}

\paragraph{Exact prompt:} 

\texttt{\justify Using official government figures, work out (to the best of your ability) how many Chinese have a disposable income of more than 100k yuan per year. You may fit parametric distributions or reason by analogy, if you cannot infer the answer directly from government sources. Today's date is \{today\}.}

\paragraph{Relevance}

\

\noindent To size a market, it is instructive to understand how many people could afford certain products.

\paragraph{Scoring rubric}
In common for all strategies:
\begin{itemize}
    \item \textbf{outdated:} did not get thrown off by outdated data
    \item \textbf{unofficial:} did not get thrown off by unofficial (low-quality) data
    \item \textbf{disposable:} did not conflate disposable income with income or household income
\end{itemize}
\noindent Strategy 1 (Fit a log-normal to given mean and percentile):
\begin{itemize}
    \item \textbf{found-data:} found a press release like \url{https://www.stats.gov.cn/english/PressRelease/202404/t20240424_1948702.html} from which mean and median annual per capita disposable income can be inferred
    \item \textbf{infer:} successfully inferred mean the above article
    \item \textbf{no-income:} thought to decompose the disposable income distribution into a delta distribution at 0 (children, etc. do not have any disposable income) and a log-normal distribution for the remaining population, weighted by the fraction of the population which does not have a disposable income; needs to adjust mean and then fit a log-normal to the adjusted mean and a different percentile 
    \item \textbf{fit:} fit a log-normal distribution to given mean and percentile and inferred an estimate of the fraction of the population earning a disposable income exceeding 100,000 Yuan
\end{itemize}
OR
\

\

\noindent Strategy 2 (Fit a reasonable distribution to quintile data):
\begin{itemize}
    \item \textbf{found-data:} find quintile data from official sources like \url{https://www.stats.gov.cn/sj/ndsj/2023/indexeh.htm} or \url{http://english.scio.gov.cn/pressroom/2024-01/26/content_116967913.htm}
    \item \textbf{fit:} fit a reasonable distribution to quintile data and inferred an estimate of the fraction of the population earning a disposable income exceeding 100,000 Yuan
\end{itemize}
OR
\

\

\noindent Strategy 3 (Fit a reasonable distribution to inferred decile data):
\begin{itemize}
    \item \textbf{found-stats:} find ``income share held by highest X\%''-type claims plausibly based on official data like \url{https://data.worldbank.org/indicator/SI.DST.10TH.10?locations=CN}
    \item \textbf{infer:} successfully inferred decentile averages\footnote{If we know that the top decentile earns $29.4\%$ of the national income and we know the national income (after all we know the mean and the number of people), we can get the average over the top decentile. Of course this is just a biased approximation since we are talking about income, not disposable income here.}
    \item \textbf{fit:} fit a reasonable distribution to decentile data and inferred an estimate of the fraction of the population earning a disposable income exceeding 100,000 Yuan
\end{itemize}
OR
\

\

\noindent Strategy 4 (Fit a reasonable distribution to mean/median and Gini coefficient):
\begin{itemize}
    \item \textbf{found-Gini:} found China's Gini coefficient based on plausibly official data like \url{https://data.worldbank.org/indicator/SI.POV.GINI?locations=CN}
    \item \textbf{fit:} fit a reasonable distribution to the mean/median and Gini coefficient and inferred an estimate of the fraction of the population earning a disposable income exceeding 100,000 Yuan
\end{itemize}
\paragraph{Performance}
\

\

\noindent See Table \ref{tab:chineseincome_combined}.

\

\begin{table}[!htbp]
    \centering
    \resizebox{\textwidth}{!}{
    \setlength{\tabcolsep}{2pt}  % Adjust this value to control space between cells
    \begin{tabular}{l>{\centering\arraybackslash}p{0.1cm}*{3}{>{\centering\arraybackslash}p{0.4cm}}p{0.15cm}*{4}{>{\centering\arraybackslash}p{0.4cm}}p{0.15cm}*{2}{>{\centering\arraybackslash}p{0.4cm}}p{0.15cm}*{3}{>{\centering\arraybackslash}p{0.4cm}}p{0.15cm}*{2}{>{\centering\arraybackslash}p{0.4cm}}p{0.15cm}>{\centering\arraybackslash}p{0.4cm}}
        \toprule
        & & \multicolumn{3}{c}{\makecell[c]{General\\ considerations}} & & \multicolumn{4}{c}{\makecell[c]{Strategy 1:\\ Fit to \\mean/median}} & & \multicolumn{2}{c}{\makecell[c]{Strategy 2:\\ Fit to \\ quintiles}} & & \multicolumn{3}{c}{\makecell[c]{Strategy 3:\\ Fit to \\ deciles}} & & \multicolumn{2}{c}{\makecell[c]{Strategy 4:\\ Fit to \\ mean/Gini}} & & \\
        \cmidrule(lr){3-5} \cmidrule(lr){7-10} \cmidrule(lr){12-13} \cmidrule(lr){15-17} \cmidrule(lr){19-20}

        & & \rot{outdated} & \rot{unofficial} & \rot{disposable} & & \rot{\makecell{found-data}} & \rot{infer} & \rot{no-income} & \rot{fit} & & \rot{\makecell{found-data}} & \rot{fit} & & \rot{\makecell{found-data}} & \rot{infer} & \rot{fit} & & \rot{\makecell{found-Gini}} & \rot{fit} & & \rot{\textbf{final}} \\
        \midrule
        \react\oonelabel & &        \successboxsmall        & \successboxsmall & \successboxsmall & & \successboxsmall & \successboxsmall & \failureboxsmall & \successboxsmall & & \NAboxsmall             & \NAboxsmall       & & \failureboxsmall    & \NAboxsmall & \NAboxsmall & & \NAboxsmall & \NAboxsmall & & \partialsuccessboxsmall \\
        \plan\oonelabel & &         \partialsuccessboxsmall & \failureboxsmall & \failureboxsmall & & \successboxsmall & \successboxsmall & \failureboxsmall & \successboxsmall & & \NAboxsmall             & \NAboxsmall       & & \NAboxsmall         & \NAboxsmall & \NAboxsmall & & \NAboxsmall & \NAboxsmall & & \failureboxsmall \\
        \delegate\oonelabel & &     \successboxsmall        & \successboxsmall & \failureboxsmall & & \successboxsmall & \successboxsmall & \failureboxsmall & \successboxsmall & & \partialsuccessboxsmall & \NAboxsmall       & & \NAboxsmall         & \NAboxsmall & \NAboxsmall & & \NAboxsmall & \NAboxsmall & & \partialsuccessboxsmall \\
        \plandelegate\oonelabel & & \partialsuccessboxsmall & \successboxsmall & \successboxsmall & & \successboxsmall & \successboxsmall & \failureboxsmall & \successboxsmall & & \successboxsmall        & \successboxsmall  & & \NAboxsmall         & \NAboxsmall & \NAboxsmall & & \NAboxsmall & \NAboxsmall & & \successboxsmall \\
        \addlinespace
        \midrule
        \react\olabel & & \successboxsmall & \successboxsmall & \successboxsmall & & \successboxsmall & \successboxsmall & \failureboxsmall & \successboxsmall & & \NAboxsmall & \NAboxsmall & & \failureboxsmall & \NAboxsmall & \NAboxsmall & & \NAboxsmall & \NAboxsmall & & \partialsuccessboxsmall \\
        \react\newclaudelabel & & \successboxsmall & \failureboxsmall & \successboxsmall & & \successboxsmall & \successboxsmall & \failureboxsmall & \failureboxsmall & & \NAboxsmall & \NAboxsmall & & \partialsuccessboxsmall & \successboxsmall & \partialsuccessboxsmall & & \successboxsmall & \partialsuccessboxsmall & & \failureboxsmall \\
        \plan\newclaudelabel & & \successboxsmall & \successboxsmall & \failureboxsmall & & \NAboxsmall & \NAboxsmall & \NAboxsmall & \NAboxsmall & & \partialsuccessboxsmall & \partialsuccessboxsmall & & \NAboxsmall & \NAboxsmall & \NAboxsmall & & \NAboxsmall & \NAboxsmall & & \failureboxsmall \\
        \delegate\newclaudelabel & & \successboxsmall & \successboxsmall & \successboxsmall & & \successboxsmall & \failureboxsmall & \NAboxsmall & \NAboxsmall & & \NAboxsmall & \NAboxsmall & & \NAboxsmall & \NAboxsmall & \NAboxsmall & & \successboxsmall & \successboxsmall & & \partialsuccessboxsmall \\
        \plandelegate\newclaudelabel & & \successboxsmall & \successboxsmall & \successboxsmall & & \successboxsmall & \successboxsmall & \failureboxsmall & \failureboxsmall & & \NAboxsmall & \NAboxsmall & & \NAboxsmall & \NAboxsmall & \NAboxsmall & & \successboxsmall & \failureboxsmall & & \partialsuccessboxsmall \\
        \addlinespace
        \midrule
        \react\gptlabel & & \successboxsmall & \successboxsmall & \successboxsmall & & \successboxsmall & \failureboxsmall & \NAboxsmall & \NAboxsmall & & \NAboxsmall & \NAboxsmall & & \NAboxsmall & \NAboxsmall & \NAboxsmall & & \NAboxsmall & \NAboxsmall & & \failureboxsmall \\
        \plan\gptlabel & & \NAboxsmall & \NAboxsmall & \NAboxsmall & & \NAboxsmall & \NAboxsmall & \NAboxsmall & \NAboxsmall & & \NAboxsmall & \NAboxsmall & & \NAboxsmall & \NAboxsmall & \NAboxsmall & & \NAboxsmall & \NAboxsmall & & \failureboxsmall \\
        \delegate\gptlabel & & \successboxsmall & \successboxsmall & \successboxsmall & & \successboxsmall & \successboxsmall & \failureboxsmall & \successboxsmall & & \NAboxsmall & \NAboxsmall & & \NAboxsmall & \NAboxsmall & \NAboxsmall & & \NAboxsmall & \NAboxsmall & & \partialsuccessboxsmall \\
        \plandelegate\gptlabel & & \NAboxsmall & \failureboxsmall & \successboxsmall & & \successboxsmall & \successboxsmall & \NAboxsmall & \NAboxsmall & & \partialsuccessboxsmall & \NAboxsmall & & \partialsuccessboxsmall & \NAboxsmall & \NAboxsmall & & \successboxsmall & \NAboxsmall & & \failureboxsmall \\
        \addlinespace
        \midrule
        \react\llamalabel & & \NAboxsmall & \NAboxsmall & \NAboxsmall & & \successboxsmall & \NAboxsmall & \NAboxsmall & \NAboxsmall & & \NAboxsmall & \NAboxsmall & & \NAboxsmall & \NAboxsmall & \NAboxsmall & & \NAboxsmall & \NAboxsmall & & \failureboxsmall \\
        \plan\llamalabel & & \NAboxsmall & \NAboxsmall & \NAboxsmall & & \NAboxsmall & \NAboxsmall & \NAboxsmall & \NAboxsmall & & \NAboxsmall & \NAboxsmall & & \NAboxsmall & \NAboxsmall & \NAboxsmall & & \NAboxsmall & \NAboxsmall & & \failureboxsmall \\
        \plandelegate\llamalabel & & \NAboxsmall & \NAboxsmall & \NAboxsmall & & \NAboxsmall & \NAboxsmall & \NAboxsmall & \NAboxsmall & & \NAboxsmall & \NAboxsmall & & \NAboxsmall & \NAboxsmall & \NAboxsmall & & \NAboxsmall & \NAboxsmall & & \failureboxsmall \\
        \delegate\llamalabel & & \NAboxsmall & \NAboxsmall & \NAboxsmall & & \NAboxsmall & \NAboxsmall & \NAboxsmall & \NAboxsmall & & \NAboxsmall & \NAboxsmall & & \NAboxsmall & \NAboxsmall & \NAboxsmall & & \NAboxsmall & \NAboxsmall & & \failureboxsmall \\
        \addlinespace
        \midrule
        \react\minigptlabel & & \NAboxsmall & \NAboxsmall & \NAboxsmall & & \NAboxsmall & \NAboxsmall & \NAboxsmall & \NAboxsmall & & \NAboxsmall & \NAboxsmall & & \NAboxsmall & \NAboxsmall & \NAboxsmall & & \NAboxsmall & \NAboxsmall & & \failureboxsmall \\
        \plan\minigptlabel & & \NAboxsmall & \NAboxsmall & \NAboxsmall & & \NAboxsmall & \NAboxsmall & \NAboxsmall & \NAboxsmall & & \NAboxsmall & \NAboxsmall & & \NAboxsmall & \NAboxsmall & \NAboxsmall & & \NAboxsmall & \NAboxsmall & & \failureboxsmall \\
        \bottomrule
    \end{tabular}
    }
    \caption{\label{tab:chineseincome_combined}Overview of how agents performed on various aspects of the task of finding how many Chinese earn an annual disposable income exceeding 100,000 Yuan. General considerations are tasks common to all strategies. Strategy 1: fitting a distribution to the mean and median data from official sources. Strategy 2: fitting a distribution to quintile data. Strategy 3: fitting a distribution to decile data. Strategy 4: fitting a distribution to the mean/median and Gini coefficient.}
\end{table}

\subsubsection{How much time passed from the beginning of 2009 H1N1 until the first seroprevalence study was published?}
\paragraph{Exact prompt:} \texttt{\justify How much time passed from the beginning of 2009 H1N1 until the first seroprevalence study was published?}
\paragraph{Relevance}

\

\noindent This kind of question is useful for coming up with base rates for how long it usually takes until we get seroprevalence studies when a new pandemic emerges.

\paragraph{Scoring rubric}
In common for all strategies:
\begin{itemize}
    \item \textbf{start:} worked out when 2009 H1N1 began
    \item \textbf{seroprevalence:} checked whether candidates for the earliest published seroprevalence paper were actually seroprevalence papers
    \item \textbf{date:} checked the exact publication date for candidates for the earliest published seroprevalence paper (in particular did not mix up the date when the study was conducted with its publication date)
\end{itemize}
\noindent Strategy 1 (Review articles):
\begin{itemize}
    \item \textbf{article:} found at least one review article on 2009 H1N1 seroprevalence studies
    \item \textbf{earliest:} found earliest cited article from the review article(s)
\end{itemize}
OR
\

\

\noindent Strategy 2 (Targeted comprehensive search):
\begin{itemize}
    \item \textbf{search:} performed a targeted search for seroprevalence studies in 2009 (and maybe early 2010)
    \item \textbf{found:} identified a reasonable candidate for the earliest seroprevalence study
\end{itemize}

\paragraph{Performance}
\

\

\noindent See Table \ref{tab:seroprevalence_combined}.

\

\begin{table}[!htbp]
    \centering
    \setlength{\tabcolsep}{2pt}  % Adjust this value to control space between cells
    \begin{tabular}{l>{\centering\arraybackslash}p{0.4cm}*{3}{>{\centering\arraybackslash}p{0.8cm}}p{0.4cm}*{2}{>{\centering\arraybackslash}p{0.8cm}}p{0.4cm}*{2}{>{\centering\arraybackslash}p{0.8cm}}p{0.4cm}>{\centering\arraybackslash}p{0.8cm}}
        \toprule
        & & \multicolumn{3}{c}{\makecell[c]{General\\ considerations}} & & \multicolumn{2}{c}{\makecell[c]{Strategy 1:\\ Review}} & & \multicolumn{2}{c}{\makecell[c]{Strategy 2:\\ Search}} & & \\
        \cmidrule(lr){3-5} \cmidrule(lr){7-8} \cmidrule(lr){10-11}
        & & \rot{start} & \rot{seroprevalence} & \rot{\makecell{date}} & & \rot{article} & \rot{earliest} & & \rot{search} & \rot{found} & & \rot{\textbf{final}} \\
        \midrule
        \react\oonelabel        & & \successbox & \failurebox   & \partialsuccessbox & & \NAbox & \NAbox & & \partialsuccessbox & \failurebox & & \failurebox \\
        \plan\oonelabel         & & \failurebox & \NAbox        & \NAbox             & & \NAbox & \NAbox & & \NAbox             & \NAbox      & & \failurebox \\
        \delegate\oonelabel     & & \successbox & \failurebox   & \failurebox        & & \NAbox & \NAbox & & \failurebox        & \failurebox & & \failurebox \\
        \plandelegate\oonelabel & & \successbox & \NAbox        & \NAbox             & & \NAbox & \NAbox & & \NAbox             & \NAbox      & & \failurebox \\
        \addlinespace
        \midrule
        \react\newclaudelabel & & \successbox & \failurebox & \successbox & & \partialsuccessbox & \failurebox & & \partialsuccessbox & \failurebox & & \failurebox \\
        \plan\newclaudelabel & & \successbox & \failurebox & \failurebox & & \partialsuccessbox & \partialsuccessbox & & \NAbox & \NAbox & & \failurebox \\
        \delegate\newclaudelabel & & \successbox & \successbox & \successbox & & \partialsuccessbox & \failurebox & & \NAbox & \NAbox & & \failurebox \\
        \plandelegate\newclaudelabel & & \successbox & \NAbox & \NAbox & & \partialsuccessbox & \failurebox & & \NAbox & \NAbox & & \failurebox \\
        \addlinespace
        \midrule
        \react\gptlabel & & \successbox & \failurebox & \partialsuccessbox & & \NAbox & \NAbox & & \partialsuccessbox & \failurebox & & \failurebox \\
        \plan\gptlabel & & \successbox & \failurebox & \failurebox & & \failurebox & \failurebox & & \NAbox & \NAbox & & \failurebox \\
        \delegate\gptlabel & & \successbox & \NAbox & \successbox & & \failurebox & \failurebox & & \NAbox & \NAbox & & \failurebox \\
        \plandelegate\gptlabel & & \successbox & \NAbox & \successbox & & \partialsuccessbox & \failurebox & & \NAbox & \NAbox & & \failurebox \\
        \addlinespace
        \midrule
        \react\llamalabel & & \successbox & \NAbox & \NAbox & & \NAbox & \NAbox & & \NAbox & \NAbox & & \failurebox \\
        \plan\llamalabel & & \successbox & \failurebox & \failurebox & & \NAbox & \NAbox & & \NAbox & \NAbox & & \failurebox \\
        \delegate\llamalabel & & \successbox & \failurebox & \failurebox & & \failurebox & \NAbox & & \NAbox & \NAbox & & \failurebox \\
        \plandelegate\llamalabel & & \successbox & \NAbox & \successbox & & \NAbox & \NAbox & & \NAbox & \NAbox & & \failurebox \\
        \addlinespace
        \midrule
        \react\minigptlabel & & \successbox & \NAbox & \NAbox & & \partialsuccessbox & \failurebox & & \NAbox & \NAbox & & \failurebox \\
        \plan\minigptlabel & & \successbox & \failurebox & \failurebox & & \failurebox & \failurebox & & \NAbox & \NAbox & & \failurebox \\
        \bottomrule
    \end{tabular}
    \caption{\label{tab:seroprevalence_combined}Overview of how agents performed on various aspects of the task of finding how long it took for the first seroprevalence study for 2009 H1N1 to be published. General considerations are tasks common to all strategies. Strategy 1: checking review papers on seroprevalence studies and finding the earliest citation. Strategy 2: performing a targeted comprehensive search.}
\end{table}

\subsubsection{Estimate the number of paying ChatGPT users across all tiers excluding Team as of June 2024}
\paragraph{Exact prompt:}
\texttt{\justify Estimate the number of paying ChatGPT users across all tiers excluding Team as of June 2024.}
\paragraph{Relevance}

\

\noindent OpenAI is a private company at the time of writing, and many firms take an interest in its (undisclosed) financials.
In order to estimate the revenue OpenAI makes from ChatGPT, one needs an estimate for the number of paid subscribers.
We exclude the Team tier here; because of the lack of available data, there are many ways of estimating the number of Team subscribers, violating the ``small number of strategies'' criterion from Section \ref{sec:evaluation}.
\paragraph{Scoring rubric:}
\begin{itemize}
    \item \textbf{tiers:} Finds that ChatGPT has 3 paid tiers: Plus, Team, and Enterprise.
    \item \textbf{US-subscribers:} Finds the claim that ChatGPT Plus had 3.9 million US subscribers as of March 2024.
    \item\textbf{subscribers-source:}  Traces the above claim to its original source
    \item \textbf{plus-growth:} Attempts to account for growth in Plus subscribers
    \item \textbf{old-US-subscribers:} Finds the claim that ChatGPT Plus had 2 million US subscribers as of July 2023.
    \item \textbf{old-subscribers-source:} Traces the above claim to its original source
    \item \textbf{reasonable-growth:} Estimates Plus subscriber growth in a reasonable way
    \item \textbf{global:} Attempts to account for the difference between US and global subscriber numbers
    \item \textbf{mobile-ratio:} Finds US-global subscriber ratios for Plus Mobile
    \item \textbf{estimate-ratio:} Estimates US-global ratio in a reasonable way
    \item \textbf{enterprise-seats:} Finds 600k Enterprise seats as of April 4, 2024
    \item \textbf{enterprise-growth:} Attempts to account for growth in Enterprise subscribers
    \item \textbf{old-enterprise-seats:} Finds 150k Enterprise seats as of January 11, 2024
    \item \textbf{enterprise-growth:} Estimates Enterprise subscriber growth in a reasonable way
\end{itemize}

\paragraph{Performance}
See Table \ref{tab:num_chatgpt_users}.

\begin{table}[!htbp]
    \centering
    \setlength{\tabcolsep}{2pt} 
    \begin{tabular}{l>{\centering\arraybackslash}p{0.1cm}*{3}{>{\centering\arraybackslash}p{0.5cm}}p{0.2cm}*{3}{>{\centering\arraybackslash}p{0.5cm}}p{0.2cm}*{3}{>{\centering\arraybackslash}p{0.5cm}}p{0.2cm}*{4}{>{\centering\arraybackslash}p{0.5cm}}p{0.2cm}>{\centering\arraybackslash}p{0.5cm}}
        \toprule
        & & \rot{tiers} & \rot{US-subscribers} & \rot{subscribers-source} &  \rot{plus-growth} & \rot{old-US-subscribers} & \rot{old-subscribers-source}  & \rot{reasonable-growth} &  \rot{global} & \rot{mobile-ratio} & \rot{estimate-ratio} &  \rot{enterprise-seats} & \rot{enterprise-growth} & \rot{old-enterprise-seats} & \rot{enterprise-growth} & & \rot{\textbf{final}} \\ 
        \midrule
        \react\oonelabel        & & \failureboxsmall & \successboxsmall & \failureboxsmall & \partialsuccessboxsmall & \successboxsmall & \failureboxsmall  & \partialsuccessboxsmall & \successboxsmall  &  \NAboxsmall  & \NAboxsmall & \NAboxsmall & \NAboxsmall & \NAboxsmall & \NAboxsmall & & \failureboxsmall \\ 
        \plan\oonelabel         & & \failureboxsmall & \failureboxsmall & \NAboxsmall & \partialsuccessboxsmall & \successboxsmall & \failureboxsmall         & \failureboxsmall & \NAboxsmall  &  \NAboxsmall  & \NAboxsmall & \NAboxsmall & \NAboxsmall & \NAboxsmall& \NAboxsmall & & \failureboxsmall \\ 
        \delegate\oonelabel     & & \failureboxsmall & \failureboxsmall & \NAboxsmall & \partialsuccessboxsmall & \failureboxsmall & \NAboxsmall         & \failureboxsmall & \NAboxsmall  &  \NAboxsmall  & \NAboxsmall & \NAboxsmall & \NAboxsmall & \NAboxsmall& \NAboxsmall & & \failureboxsmall \\ 
        \plandelegate\oonelabel & & \failureboxsmall & \failureboxsmall & \NAboxsmall & \failureboxsmall        & \NAboxsmall & \NAboxsmall         & \failureboxsmall & \NAboxsmall  &  \NAboxsmall  & \NAboxsmall & \NAboxsmall & \NAboxsmall & \NAboxsmall& \NAboxsmall & & \failureboxsmall \\ 
        \addlinespace
        \midrule
        \react\newclaudelabel & & \failureboxsmall & \successboxsmall & \failureboxsmall  & \partialsuccessboxsmall & \failureboxsmall & \NAboxsmall  & \failureboxsmall & \partialsuccessboxsmall & \failureboxsmall & \failureboxsmall & \NAboxsmall & \NAboxsmall & \NAboxsmall & \NAboxsmall & & \failureboxsmall \\ 
        \plan\newclaudelabel & & \failureboxsmall & \successboxsmall & \failureboxsmall  & \successboxsmall & \failureboxsmall & \NAboxsmall  & \failureboxsmall & \successboxsmall & \failureboxsmall & \failureboxsmall & \NAboxsmall & \NAboxsmall & \NAboxsmall & \NAboxsmall  & & \failureboxsmall \\ 
        \delegate\newclaudelabel & & \failureboxsmall & \successboxsmall & \failureboxsmall  & \successboxsmall & \failureboxsmall & \NAboxsmall  & \failureboxsmall & \successboxsmall & \failureboxsmall & \failureboxsmall & \NAboxsmall & \NAboxsmall & \NAboxsmall & \NAboxsmall & & \failureboxsmall \\ 
        \plandelegate\newclaudelabel & & \failureboxsmall & \successboxsmall & \failureboxsmall  & \failureboxsmall & \NAboxsmall & \NAboxsmall  & \NAboxsmall & \failureboxsmall & \NAboxsmall & \NAboxsmall & \successboxsmall & \NAboxsmall & \NAboxsmall & \NAboxsmall & & \failureboxsmall \\ 
        \addlinespace
        \midrule
        \react\gptlabel & & \failureboxsmall & \failureboxsmall & \NAboxsmall  & \failureboxsmall & \NAboxsmall & \NAboxsmall & \NAboxsmall & \failureboxsmall & \NAboxsmall  & \NAboxsmall & \NAboxsmall & \NAboxsmall & \NAboxsmall & \NAboxsmall & & \failureboxsmall \\ 
        \plan\gptlabel & & \failureboxsmall & \successboxsmall & \failureboxsmall  & \failureboxsmall & \NAboxsmall & \NAboxsmall & \NAboxsmall & \successboxsmall & \failureboxsmall  & \failureboxsmall & \NAboxsmall & \NAboxsmall & \NAboxsmall  & \NAboxsmall & & \failureboxsmall \\ 
        \delegate\gptlabel & & \successboxsmall & \successboxsmall & \failureboxsmall  & \failureboxsmall & \NAboxsmall & \NAboxsmall & \NAboxsmall & \successboxsmall & \failureboxsmall  & \failureboxsmall & \failureboxsmall & \failureboxsmall & \NAboxsmall & \NAboxsmall  & & \failureboxsmall \\ 
        \plandelegate\gptlabel & & \failureboxsmall & \successboxsmall & \failureboxsmall  & \partialsuccessboxsmall & \failureboxsmall & \NAboxsmall & \failureboxsmall & \failureboxsmall & \NAboxsmall  & \NAboxsmall & \NAboxsmall & \NAboxsmall & \NAboxsmall & \NAboxsmall  & & \failureboxsmall \\ 
        \addlinespace
        \midrule
        \react\llamalabel & & \failureboxsmall & \failureboxsmall & \NAboxsmall  & \failureboxsmall & \failureboxsmall & \NAboxsmall  & \NAboxsmall & \NAboxsmall & \failureboxsmall  & \NAboxsmall & \NAboxsmall & \NAboxsmall & \NAboxsmall & \NAboxsmall  & & \failureboxsmall \\ 
        \plan\llamalabel & &  \failureboxsmall & \failureboxsmall & \NAboxsmall  & \failureboxsmall & \failureboxsmall & \NAboxsmall  & \NAboxsmall & \NAboxsmall & \failureboxsmall  & \NAboxsmall & \NAboxsmall & \NAboxsmall & \NAboxsmall & \NAboxsmall  & & \failureboxsmall \\ 
        \delegate\llamalabel & & \successboxsmall & \failureboxsmall & \NAboxsmall  & \failureboxsmall & \failureboxsmall & \NAboxsmall  & \NAboxsmall & \NAboxsmall & \failureboxsmall  & \NAboxsmall & \failureboxsmall & \failureboxsmall & \NAboxsmall & \NAboxsmall  & & \failureboxsmall \\ 
        \plandelegate\llamalabel & & \failureboxsmall & \failureboxsmall & \NAboxsmall  & \failureboxsmall & \failureboxsmall & \NAboxsmall  & \NAboxsmall & \NAboxsmall & \failureboxsmall  & \NAboxsmall & \NAboxsmall & \NAboxsmall & \NAboxsmall & \NAboxsmall  & & \failureboxsmall \\ 
        \addlinespace
        \midrule
        \react\minigptlabel & & \failureboxsmall & \failureboxsmall & \NAboxsmall  & \failureboxsmall & \failureboxsmall & \NAboxsmall  & \NAboxsmall & \NAboxsmall & \failureboxsmall  & \NAboxsmall & \NAboxsmall & \NAboxsmall & \NAboxsmall & \NAboxsmall  & & \failureboxsmall \\ 
        \plan\minigptlabel & & \failureboxsmall & \failureboxsmall & \NAboxsmall  & \failureboxsmall & \failureboxsmall & \NAboxsmall  & \NAboxsmall & \NAboxsmall & \failureboxsmall  & \NAboxsmall & \NAboxsmall & \NAboxsmall & \NAboxsmall & \NAboxsmall  & & \failureboxsmall \\ 
        \bottomrule
    \end{tabular}
    \caption{\label{tab:num_chatgpt_users}Overview of how agents performed on the task of finding how many paying ChatGPT users there are.}
\end{table}

\subsubsection{How many research scientists are there at OpenAI?}
\paragraph{Exact prompt:} \texttt{\justify Work out how many research scientists there currently are at OpenAI to the best of your ability.}
\paragraph{Relevance}

\

\noindent Estimating how many researchers big AI labs employ is a natural subquestion when working on how much they spend on employees as well as for researching AI progress and safety-related questions.

\paragraph{Scoring rubric}
\noindent Strategy 1 (Infer from academic publications):
\begin{itemize}
    \item \textbf{article:} found a publication like the GPT-4 technical report
    \item \textbf{composition:} realized that not all 182 authors are research scientists
    \item \textbf{estimate:} estimate what fraction of authors are research scientists (e.g. by subsampling and checking some of them manually)
\end{itemize}
OR
\

\

\noindent Strategy 2 (Estimate fraction directly):
\begin{itemize}
    \item \textbf{total:} estimated how many employees there are at OpenAI
    \item \textbf{estimate:} reasonably estimated what fraction of employees are research scientists
\end{itemize}

\paragraph{Performance}
\

\

\noindent See
Table \ref{tab:combined_openai}.

\

\begin{table}[!htbp]
    \centering
    \setlength{\tabcolsep}{2pt}  % Adjust this value to control space between cells
    \begin{tabular}{l>{\centering\arraybackslash}p{0.4cm}*{3}{>{\centering\arraybackslash}p{0.8cm}}p{0.4cm}*{2}{>{\centering\arraybackslash}p{0.8cm}}p{0.4cm}>{\centering\arraybackslash}p{0.8cm}}
        \toprule
        & & \multicolumn{3}{c}{\makecell[c]{Strategy 1:\\ Publications}} & & \multicolumn{2}{c}{\makecell[c]{Strategy 2:\\ Estimate}}  \\
        \cmidrule(lr){3-5} \cmidrule(lr){7-8}
        & & \rot{article} & \rot{composition} & \rot{estimate} & & \rot{total} & \rot{estimate} & & \rot{\textbf{final}} \\
        \midrule
        \react\oonelabel        & & \NAbox              & \NAbox & \NAbox & & \successbox        & \partialsuccessbox & & \failurebox \\ 
        \plan\oonelabel         & & \NAbox              & \NAbox & \NAbox & & \NAbox             & \NAbox             & & \failurebox \\ 
        \delegate\oonelabel     & & \NAbox              & \NAbox & \NAbox & & \partialsuccessbox & \partialsuccessbox & & \failurebox \\ 
        \plandelegate\oonelabel & & \partialsuccessbox  & \NAbox & \NAbox & & \partialsuccessbox & \NAbox             & & \failurebox \\ 
        \addlinespace
        \midrule
        \react\newclaudelabel & & \successbox & \successbox & \failurebox & & \successbox & \partialsuccessbox & & \failurebox \\ 
        \plan\newclaudelabel & & \NAbox & \NAbox & \NAbox & & \successbox & \failurebox & & \failurebox \\ 
        \delegate\newclaudelabel & & \NAbox & \NAbox & \NAbox & & \successbox & \partialsuccessbox & & \failurebox \\ 
        \plandelegate\newclaudelabel & & \failurebox & \NAbox & \NAbox & & \failurebox & \failurebox & & \failurebox \\ 
        \addlinespace
        \midrule
        \react\gptlabel & & \NAbox & \NAbox & \NAbox & & \failurebox & \failurebox & & \failurebox \\ 
        \plan\gptlabel & & \NAbox & \NAbox & \NAbox & & \failurebox & \failurebox & & \failurebox \\ 
        \delegate\gptlabel & & \failurebox & \NAbox & \NAbox & & \failurebox & \failurebox & & \failurebox \\ 
        \plandelegate\gptlabel & & \failurebox & \NAbox & \NAbox & & \successbox & \failurebox & & \failurebox \\ 
        \addlinespace
        \midrule
        \react\llamalabel & & \NAbox & \NAbox & \NAbox & & \successbox & \failurebox & & \failurebox \\ 
        \plan\llamalabel & & \NAbox & \NAbox & \NAbox & & \NAbox & \NAbox & & \failurebox \\ 
        \delegate\llamalabel & & \NAbox & \NAbox & \NAbox & & \partialsuccessbox & \failurebox & & \failurebox \\ 
        \plandelegate\llamalabel & & \successbox & \NAbox & \NAbox & & \NAbox & \NAbox & & \failurebox \\ 
        \addlinespace
        \midrule
        \react\minigptlabel & & \NAbox & \NAbox & \NAbox & & \failurebox & \failurebox & & \failurebox \\ 
        \plan\minigptlabel & & \partialsuccessbox & \NAbox & \NAbox & & \failurebox & \failurebox & & \failurebox \\ 
        \bottomrule
    \end{tabular}
    \caption{\label{tab:combined_openai}Overview of how agents performed on the task of finding how many research scientists there are at OpenAI, based on different strategies. Strategy 1: going through publications. Strategy 2: estimating the fraction of research scientists.}
\end{table}

\subsubsection{Track down original source for a blog post claim about number of paying ChatGPT mobile users.}
\paragraph{Exact prompt:} \texttt{\justify Track down where the claim that ChatGPT mobile had about 230,000 to 250,000 paying users as of October 2023 (originally found on https://nerdynav.com/chatgpt-statistics/) came from.}
\paragraph{Scoring rubric:}
\begin{itemize}
    \item [Preparation:]
          \begin{itemize}
              \item \textbf{nerdynav:} read the nerdynav blog post \& identified the right section
              \item \textbf{search:} realised that it ought to search for sources on monthly app revenue
          \end{itemize}
    \item [Data collection:]
          \begin{itemize}
            \item \textbf{techcrunch:} found the right TechCrunch article
          \end{itemize}
    \item [Verification:]
          \begin{itemize}
            \item \textbf{dates:} checked dates to see if sources can be original sources for given claim
            \item \textbf{derive:} checked if data can be used to derive given claim
          \end{itemize}
    \item [Performance:]
          \begin{itemize}
            \item \textbf{growth:} used month-on-month growth figure
            \item \textbf{gross/net:} got the difference between gross and net revenue
            \item \textbf{appfigures:} found appfigures.com
          \end{itemize}
\end{itemize}
\paragraph{Performance}
\

\

\noindent See
Table \ref{tab:nerdynav_evals}.

\begin{table}[!htbp]
    \centering
    \setlength{\tabcolsep}{2pt}  % Adjust this value to control space between cells
    \begin{tabular}{l>{\centering\arraybackslash}p{0.2cm}*{8}{>{\centering\arraybackslash}p{0.5cm}}p{0.2cm}>{\centering\arraybackslash}p{0.5cm}}
        \toprule
        & & \rot{nerdynav} & \rot{search} & \rot{techcrunch} & \rot{dates} & \rot{derive} & \rot{growth} & \rot{gross/net} & \rot{appfigures} & & \rot{\textbf{final}} \\
        \midrule
        \react\oonelabel        & & \successboxsmall & \failureboxsmall & \successboxsmall & \NAboxsmall        & \successboxsmall & \failureboxsmall & \failureboxsmall & \failureboxsmall & & \failureboxsmall \\
        \plan\oonelabel         & & \successboxsmall & \failureboxsmall & \successboxsmall & \NAboxsmall        & \successboxsmall & \failureboxsmall & \failureboxsmall & \failureboxsmall & & \failureboxsmall \\
        \delegate\oonelabel     & & \successboxsmall & \successboxsmall & \successboxsmall & \NAboxsmall        & \successboxsmall & \failureboxsmall & \failureboxsmall & \failureboxsmall & & \failureboxsmall \\
        \plandelegate\oonelabel & & \successboxsmall & \failureboxsmall & \successboxsmall & \successboxsmall   & \successboxsmall & \failureboxsmall & \failureboxsmall & \failureboxsmall & & \failureboxsmall \\
        \addlinespace
        \midrule
        \react\newclaudelabel & & \failureboxsmall & \successboxsmall & \partialsuccessboxsmall & \partialsuccessboxsmall & \partialsuccessboxsmall & \failureboxsmall & \failureboxsmall & \successboxsmall & & \partialsuccessboxsmall \\
        \plan\newclaudelabel & & \successboxsmall & \successboxsmall & \successboxsmall & \failureboxsmall & \partialsuccessboxsmall & \failureboxsmall & \NAboxsmall & \successboxsmall & & \partialsuccessboxsmall \\
        \delegate\newclaudelabel & & \successboxsmall & \partialsuccessboxsmall & \successboxsmall & \partialsuccessboxsmall & \partialsuccessboxsmall & \failureboxsmall & \failureboxsmall & \partialsuccessboxsmall & & \failureboxsmall \\
        \plandelegate\newclaudelabel & & \successboxsmall & \successboxsmall & \successboxsmall & \failureboxsmall & \partialsuccessboxsmall & \failureboxsmall & \NAboxsmall & \failureboxsmall & & \failureboxsmall \\
        \addlinespace
        \midrule
        \react\gptlabel & & \successboxsmall & \successboxsmall & \successboxsmall & \partialsuccessboxsmall & \partialsuccessboxsmall & \failureboxsmall & \failureboxsmall & \failureboxsmall & & \failureboxsmall \\
        \plan\gptlabel & & \successboxsmall & \failureboxsmall & \successboxsmall & \failureboxsmall & \partialsuccessboxsmall & \failureboxsmall & \NAboxsmall & \failureboxsmall & & \failureboxsmall \\
        \delegate\gptlabel & & \successboxsmall & \failureboxsmall & \successboxsmall & \failureboxsmall & \partialsuccessboxsmall & \failureboxsmall & \failureboxsmall & \failureboxsmall & & \failureboxsmall \\
        \plandelegate\gptlabel & & \successboxsmall & \partialsuccessboxsmall & \failureboxsmall & \NAboxsmall & \failureboxsmall & \NAboxsmall & \NAboxsmall & \failureboxsmall & & \failureboxsmall \\
        \addlinespace
        \midrule
        \react\llamalabel & & \successboxsmall & \failureboxsmall & \partialsuccessboxsmall & \failureboxsmall & \failureboxsmall & \NAboxsmall & \NAboxsmall & \failureboxsmall & & \failureboxsmall \\
        \plan\llamalabel & & \successboxsmall & \partialsuccessboxsmall & \partialsuccessboxsmall & \NAboxsmall & \NAboxsmall & \NAboxsmall & \NAboxsmall & \failureboxsmall & & \failureboxsmall \\
        \delegate\llamalabel & & \failureboxsmall & \NAboxsmall & \NAboxsmall & \NAboxsmall & \NAboxsmall & \NAboxsmall & \NAboxsmall & \failureboxsmall & & \failureboxsmall \\
        \plandelegate\llamalabel & & \successboxsmall & \partialsuccessboxsmall & \NAboxsmall & \NAboxsmall & \NAboxsmall & \NAboxsmall & \NAboxsmall & \failureboxsmall & & \failureboxsmall \\
        \addlinespace
        \midrule
        \react\minigptlabel & & \failureboxsmall & \failureboxsmall & \successboxsmall & \failureboxsmall & \failureboxsmall & \failureboxsmall & \NAboxsmall & \failureboxsmall & & \failureboxsmall \\
        \plan\minigptlabel & & \successboxsmall & \failureboxsmall & \successboxsmall & \failureboxsmall & \failureboxsmall & \failureboxsmall & \NAboxsmall & \failureboxsmall & & \failureboxsmall \\
        \bottomrule
    \end{tabular}
    \caption{\label{tab:nerdynav_evals}Overview of how agents performed on the task of tracking down the original source for a claim about the number of paying ChatGPT mobile users.}
\end{table}

\subsection{Taxonomy of failures} \label{sec:taxonomy-of-failures}
\subsubsection{By architecture}
\begin{itemize}
    \item [ReAct] (Prefix: (\react))
        \begin{itemize}
            \item Agents based on ReAct architectures tend to overlook relevant information like datasets that almost certainly would allow them to infer the answer to a question because they do not appear to be as specific as narrower, but basically useless search results. 
            \item More generally, they struggle with how to ``spread their attention'': When trying to look up some piece of information, should they keep at it or change course when they find information suggesting that another approach seems more promising?
            \item They also struggle with tasks that are intrinsically parallelisable, like inferring a list of OpenAI employees and checking for each (or even just a reasonably sized sumsample) whether they are research scientists.
            \item Weaker models tend to get stuck in loops a lot.
        \end{itemize}
    \item [Planning] (Prefix: \plan)
        \begin{itemize}
            \item Occasionally they fail to update their plans, putting themselves in situations where they get stuck in loops repeating the same action over and over again or force themselves to terminate early and hence hallucinate a result.
            \item Sometimes they fail to record information in their ``memory'' or even erase it seemingly at random after several turns.
        \end{itemize}
    \item [ReAct w/ subtasks] (Prefix: (\delegate))
        \begin{itemize}
            \item Many LLMs fundamentally cannot understand that ``Research X'' and ``Once X is researched, look into [something depending on X]'' cannot be run in parallel, despite explicitly being told so.
        \end{itemize}
    \item [Planning w/ subtasks] (Prefix: (\plandelegate))
        \begin{itemize}
            \item Excels at parallelisable tasks, but also turns some non-parallelisable tasks into parallel ones, e.g. ``Look for occurrences of X in the 20th century'' becomes ``Look for occurrences of X in the first/second/third/... decade of the 20th century''. 
            \item Often gets lost in researching fairly inconsequential questions, resulting in lengthy and expensive runs going nowhere.
            \item Often fails to specify task in a self-contained, unambiguous way. The more verbose \oone is better than other models at this since it specifies tasks like \texttt{Using the available per capita disposable income quintile data for China, fit a log-normal distribution to model the income distribution. Then, use this fitted distribution to estimate the proportion of individuals earning more than 100,000 yuan per year. Document the entire analysis process thoroughly, including all calculations, assumptions made, and justifications for the methods used.} whereas other models would simply specify a task like \texttt{Fit a log-normal distribution to quintile data to model the income distribution.} which, despite explicit instructions to make tasks self-contained and intelligible to another person without any context whatsoever, is clearly not self-contained at all. (Which quintile data? For what country? To what end?)
        \end{itemize}
\end{itemize}
Arguably some of these issues could be caused by and solved with prompt engineering, but it is remarkable how much less likely models that are generally agreed upon to be better are to make such mistakes and hence require additional prompt engineering. 
\subsubsection{By model}
\begin{itemize}
    \item [\oone]
          \begin{itemize}
              \item Occasionally aces a task, but when it does not, it is surprisingly mediocre.
              \item Very verbose (in a good way).
              \item Comes up with great plans. For example, when looking for Chinese government data on disposable income, it just knows that China tends to publish quintile data, as well as means and medians, and devises an appropriate plan around this.
              \item In other words, it seems to be able to draw on its latent knowledge more than other LLMs.
          \end{itemize}
    \item [\newclaude]
    \begin{itemize}
              \item Consistently good results.
              \item Comes up with very good plans.
              \item Often prefaces output that is supposed to be JSON and JSON only with a sentence or two justifying its output, so a built-in retry to fix is quite important.
              \item Does not flat out refuse to answer out of fear of violating someone's privacy or spreading misinformation as often as its predecessor or Claude 3 Opus, i.e. basically almost never.
              \item Does not seem to ``know'' as much as \gpt, but its common sense reasoning is much, much better; this becomes quite apparent in more complicated agent runs that end up failing because of silly reasons significantly less often.
          \end{itemize}
          \item [\gpt]
          \begin{itemize}
            \item Gets stuck repeating the same steps noticeably more often than Claude models.
            \item More creative in its approaches than \minigpt, but much less so than \newclaude.
            \item Much less adaptive/more stubborn than \newclaude. For example, if it encountered a piece of information that rendered the approach it is currently pursuing useless, it would not just give up, but would continue until the bitter end.
            \item Potentially just a fluke, but when researching BTC-related questions, all other models seemed to become noticeably dumber, while \gpt seemed to become only a little dumber. So \gpt \emph{might} be more consistent/less impressionable. More research is needed as this observation is based on a single task.
        \end{itemize}
        \item [\llama]
        \begin{itemize}
            \item Often wants to use ``academic sources'' (Google Scholar, JSTOR, etc.) even when they are clearly not relevant.
            \item Occasionally fails simple tasks like replying with JSON conforming to a given (simple) JSON schema.
            \item Tends to think that ``You can find this on Wikipedia/Github/...'' is a good solution to the task of ``Find information X'' and generally likes deferential, terse replies.
            \item Often fails to update plans properly.
            \item Often just dismisses websites solely due to their Google search snippet not containing the information it is currently looking for.
            \item Quick to give up. Seems very terse and superficial.
            \item Not very creative, but surprises once in a while.
        \end{itemize}
        \item [\minigpt]
        \begin{itemize}
            \item Not very creative, i.e. usually only tries the simplest, most direct approach and not much else.
            \item Gets stuck in loops quite often.
        \end{itemize}
        \item [\minillama]
        \begin{itemize}
            \item Also likes ``academic sources''.
            \item Gets stuck in loops quite often.
        \end{itemize}
\end{itemize}

\bibliographystyle{plainnat}
\bibliography{bibliography}

\newpage
\appendix
\section{Derivation of scoring rubrics} \label{sec:rubric-derivation}

% \subsection{Geopolitical forecasting}
\subsection{Compile a list of Chinese AI labs that have carried out training runs using over 1e24 FLOPs}
\paragraph{Ideal agent trace outline:}
\begin{itemize}
    \item Try a direct Google search for something like \texttt{Which Chinese AI labs have carried out training runs using over 1e24 FLOPs?} just in case someone actually compiled this particular list or Google's NLP search query interpretation performs a miracle.
    \item\ [partial credit] Given that 1e24 FLOPs is (currently) \emph{a lot} of compute, we may restrict our search to well-known, big AI labs like Alibaba, Tencent, Baidu, ByteDance, NRCPC, Huawei.\footnote{This turns out to be a flawed approach, missing models like \texttt{DeepSeek-Coder-V2-236B} (DeepSeek) and \texttt{ChatGLM3} (Zhipu AI), according to estimates from \url{https://epochai.org/data/notable-ai-models}.}
          \begin{itemize}
              \item For each of these companies we then look for their biggest\footnote{Here we make the implicit assumption that the models with the highest parameter count tend to be the ones that got trained using the most FLOPs.} models (e.g. googling \texttt{Baidu biggest model} quickly yields \texttt{ERNIE 3.0 Titan}).
              \item For each of these models, we then look for estimates of how many FLOPs they took (e.g. googling \texttt{ERNIE 3.0 Titan FLOP estimate training run} yields an estimate of 3.14e11 TeraFLOPs\footnote{This estimate is to be taken with a grain of salt as it said 3.14e11 TeraFLOP\textbf{S}. We assume this to be a typo. \\\url{https://archive.ph/WSsMf} accessed on 2024-06-20.} = 3.14e23 FLOPs).
          \end{itemize}
    \item [OR]
    \item Try searching for a dataset of models with (potentially estimated) FLOPs. Googling something like \texttt{largest ai training runs "FLOP"} should yield a dataset like \url{https://ourworldindata.org/grapher/artificial-intelligence-training-computation} or its source \url{https://epochai.org/data/notable-ai-models} (or something slightly outdated like \url{https://epochai.org/blog/who-is-leading-in-ai-an-analysis-of-industry-ai-research})
    \item Either infer the answer accessing either of these sites' GUI or download the dataset and write a query for it.
\end{itemize}
\paragraph{Notes}

\

\noindent This is a very hard question because companies are not transparent about their training runs, so we will have to go by estimates instead of official numbers. This also all but rules out going by press releases or news articles (which will usually not even attempt to estimate this quantity).

Moreover, there is nothing fundamentally interesting about the 1e24 FLOPs threshold so the answer likely cannot be found directly, searching for a query including ``1e24" would not necessarily unearth an article like ``Baidu just finished training a model with 1e25 FLOPs'', and FLOPs and FLOPS are occasionally mistaken for one another.

However, 1e24 FLOPs is quite a lot at the time of writing: The biggest model at the time of writing (GPT-4) is estimated to have used less than 1e26 FLOPs and most current LLMs are estimated to have used between 1e23 and 1e25 FLOPs.\footnote{Note that ChatGPT-4o is aware of this: When asked how many FLOPs training a state-of-the-art LLM would likely take, it gives a lower bound of 1e23 (coming from known numbers for GPT-3) and claims that it could easily take up to two orders of magnitude more these days.}
So there is hope of finding a dataset with estimates for how involved training was for very large language models and, given that training runs exceeding 1e24 FLOPs should be considered quite significant, chances are that such a dataset would give an exhaustive list of such Chinese AI labs in question.

The alternative of an agent estimating these numbers itself is out of scope for this task.

The alternative of looking for individual data points (estimates) does not currently seem very feasible, but could be possible.

\subsection{Find original sources shedding light on the Sino-Russian relationship after their announcement of ``friendship without limits''.}
\paragraph{Ideal agent trace outline:}
\begin{itemize}
    \item Look up/remember when Russia and China first announced their ``friendship without limits'' and restrict future searches to after that date.
    \item Since it is not clear what kind of actions, treaties, etc. would shed the most light on the current Sino-Russian relationship, figure out what ``areas of contention'' there are between them.
    \begin{itemize}
        \item After some thinking or, alternatively, going through \url{https://en.wikipedia.org/wiki/China%E2%80%93Russia_relations} one might come up with the following list: 
        \begin{itemize}
            \item China's stance on the Russo-Ukrainian war (blocking sanctions or abstaining from voting on them; military support; arms deals),
            \item Russia's stance on Taiwan/South China Sea/Tibet/Xinjiang (merely acknowledging one-China policy; alliances; assurances of support in case of conflict; military exercises/joint patrols),
            \item economic ties like Russia's involvement in the Belt and Road Initiative or gas exports to China,
            \item China's stance on border disputes,
            \item Russia's willingness to supply military equipment or technology to China or vice versa,
            \item lists of treaties/documents signed at recent high-profile meetings.
        \end{itemize}
        \item An absence of significant announcements in these domains is also of interest.
    \end{itemize}
    \item Look for latest joint statements or signed documents with searches like ``Russia China (joint statement OR signed documents)''.
    \begin{itemize}
        \item This should show something like \url{http://en.kremlin.ru/events/president/news/74049}, where we should spot the link to \url{http://en.kremlin.ru/catalog/countries/CN/events}, a chronologically ordered list of press releases relevant to Sino-Russian relationships.
        \item The release titled ``Russian-Chinese talks''\footnote{\url{http://en.kremlin.ru/catalog/countries/CN/events/74045}} stands out as it marks the 75th anniversary of diplomatic relations between the two countries. What is most notable in this press release, however, is the absence of notable agreements, as indicated by the wish to ``to create a transborder wildlife reserve The Land of Big Cats, as well as the Unified Concept for Development of the Bolshoi Ussuriysky Island (Heixiazidao)'' being at the top of the list. 
    \end{itemize}
    \item Noting the absence of Chinese press releases in English, we might want to use Chinese search terms.
    \item Look for statements on each of the above-mentioned points.
    \begin{itemize}
        \item For example, googling \texttt{China Russia Ukraine sanctions block abstain} should show articles like \url{https://en.wikipedia.org/wiki/China_and_the_Russian_invasion_of_Ukraine} and \url{https://en.wikipedia.org/wiki/International_sanctions_during_the_Russo-Ukrainian_War}, both citing many relevant primary sources.
    \end{itemize}
    \item Extract references to primary sources from articles found in the above searches. 
\end{itemize}
\paragraph{Notes}

\

\noindent This is a fairly open-ended question. The difficulty here is to actually find original sources and to not just zero in on the first few that you happen to stumble upon after a quick Google search or trying too hard to find original sources directly. It is also not obvious whether China just does not publish transcripts, lists of signed documents, etc. like Russia does, or whether finding them would require more familiarity with Chinese search engines.

Another issue/curious observation is that the planning-based agents did significantly better on a previous iteration of the task that asked for primary sources instead of giving a definition of ``original'' sources and then asking for original sources. We suspect this is because (without more context) the mentioning of primary sources helps it ``get into the right mindset'', but since what we are ultimately after simply are not primary sources, this is just tough luck.

% \subsection{Financial forecasting}
\subsection{Use official government data to estimate how many Chinese have an annual disposable income exceeding 100,000 Yuan.}
\paragraph{Ideal agent trace outline:}
\begin{itemize}
    \item 100k is a nice number, maybe someone already figured this out. So try some naive search queries.
    \begin{itemize}
        \item This should yield this McKinsey report\cite{farrell2006value} claiming that around 1\% earn more than 100k (although based on their own econometric model, not official government data). But this report is from 2006 and income grew a lot since then, so this provides a very low lower bound of 1\%.
    \end{itemize}
    \item Failing this, google something like \texttt{China National Bureau of Statistics disposable income distribution} in an attempt to infer the answer from a full distribution.
    \begin{itemize}
        \item This should yield \url{https://www.stats.gov.cn/english/PressRelease/202404/t20240424_1948702.html} from where we can infer mean and median of the disposable income distribution although there is some potential ambiguity regarding per capita disposable income or per capita household disposable income.
    \end{itemize}
    \item Alternatively, google something like \texttt{site:stats.gov.cn disposable income distribution China 2023}.
    \begin{itemize}
        \item This should yield \url{https://www.stats.gov.cn/sj/ndsj/2023/indexeh.htm} with a snippet like \texttt{Brief Introduction · 6-1 Nationwide Per Capita Income and Consumption Expenditure · 6-2 Nationwide Per Capita Disposable Income of Households by Income Quintile …} which should make you curious about quintile data. (Unfortunately, navigating and reading this page is impossible for a text-only agent, but visual agents can easily infer quintiles from here.)
    \end{itemize}
    \item Failing this, look for official datasets or, if they are hard to find, academic papers about income inequality, for example, since, in order to calculate Gini coefficients, they would have to have access to income data.
    \begin{itemize}
        \item Take care not to mix up disposable income data with income data and per capita numbers with per capita household numbers, and income with wealth, etc. If you do, be sure to explicitly acknowledge this approximation.
        \item This should yield some discontinued datasets as well as the somewhat up-to-date China Family Panel Studies (CFPS) which, unfortunately, cannot be accessed or purchased at the time of writing because of some server problems.
        \item This might also yield papers like \cite{li2020top} from which we can infer the answer to be around 7\%; but it is not official government data, so this should not be the answer.
    \end{itemize}
    \item Failing to find the full distribution, look for quintile or decile data.
    \begin{itemize}
        \item Take care not to mix up the average over the top quintile with the 80th percentile, etc.
        \item This should yield \url{http://english.scio.gov.cn/pressroom/2024-01/26/content_116967913.htm} with the paragraph \texttt{Grouped by income quintile, the per capita disposable income of low-income group reached 9,215 yuan, the lower-middle-income group 20,442 yuan, the middle-income group 32,195 yuan, the upper-middle-income group 50,220 yuan and the high-income group 95,055 yuan.}
        \item This allows us to infer a strict upper bound on the fraction of Chinese with disposable income exceeding 100k yuan: It cannot be higher than 20\%, since the average over the top 20\% is (slightly) lower than 100k.
        \item At this point we ought to make some reasonable modeling assumptions like fitting a log-normal distribution to having its 80th percentile somewhere in the middle between 50k and 95k, its 60th percentile somewhere between 32k and 50k, etc. and inferring the answer.
    \end{itemize}
    \item If the agent happened to miss this, we should start googling in Chinese.
    % E.g. 2023年中国人均可支配收入官方统计 (2023 or, alternatively, 2024 Q1, or similar, because 2024 is not over yet), maybe with 国家统计局 (National Bureau of Statistics) and/or 年度报告 (annual report).
    \begin{itemize}
        \item This should yield \url{https://www.stats.gov.cn/sj/ndsj/2023/indexch.htm} where subsection 6.2 on quintiles should stand out.
        \item Unfortunately, the data is presented as an image, so text-only agents cannot make use of it.
    \end{itemize}
    \item Failing all of the above, we should have collected the mean $\bar X$ and median $\hat X$ annual per capita disposable income. We can calculate the parameters of a log-normal distribution from these two parameters, but only after accounting for the fact that income distributions are always mixtures of a delta distribution at 0 (children (officially) don't earn anything, etc.) and then something log-normal-like. So after estimating what percentage of Chinese are working/claiming benefits/getting a pension (say $100p$\% do), then we can model these $100p\%$ with a log-normal distribution with $\bar X\over p$ and $(50-100p)$-th percentile equal to $\hat X$.
\end{itemize}
\paragraph{Notes}

\ 

\noindent There are many ways to trip up:
\begin{itemize}
    \item (Disposable) income distributions have a delta distribution at 0 because of children,
    \item data like China Family Panel Studies (CFPS), conducted every two years since 2010, appears to be discontinued and cannot be accessed (officially) anymore (see \url{https://www.isss.pku.edu.cn/cfps/en/data/public/index.htm?CSRFT=LBAE-UBB2-AEKJ-OSJJ-6MQA-ZY8E-03XJ-OA6O}),
    \item inconsistencies between (suitably extrapolated) quarterly data and annual data,
    \item wildly inconsistent claims about Chinese income distribution,
    \item mixing up annual per capita disposable income and annual per capita disposable household income,
    \item mixing up nominal values with values adjusted for various things like PPP, etc.,
    \item mixing up various definitions of ``middle class''; the most popular one is by the National Bureau of Statistics of China, defining the ``middle-income group'' as three-person households earning between 100k to 500k per year,
    \item WorldBank data (\url{https://data.worldbank.org/indicator/SI.DST.10TH.10?contextual=income-distribution&locations=CN}) is somewhat outdated (the most recent datapoint is from 2020),
    \item outdated McKinsey report\cite{farrell2006value},
    \item misleading Statista claims about how ``the largest share of Chinese middle-class families had an annual income of between 100 thousand and 300 thousand yuan per year'' (\url{https://www.statista.com/statistics/1319678/china-income-distribution-of-middle-class-families-2022/}; but ``middle class'' is \emph{defined} to be a group of three with income between 100k and 500k),
    \item claims that the top decile of earners made 41\% of the total income in 2015 (this is based on numbers wildly incompatible with government data),
    \item being thrown off by numbers for urban vs rural households (very prominently featured in many places),
    \item statistics become outdated very fast due to fast economic growth.
\end{itemize}

\subsection{Forecasts related to ``Will BTC trade at over USD90k before 2025?''}
\paragraph{Ideal agent trace outline:}
\begin{itemize}
    \item Think about where to get forecasts for Bitcoin prices from. This should return a list like
    \begin{itemize}
      \item \textbf{Prediction markets}: Polymarket, Manifold, Kalshi, \dots
      \item \textbf{Option markets}: CME Group, \dots
      \item \textbf{Finance websites with probabilistic forecasts}: CoinMarketCap, CoinGecko, but without price targets by obvious Bitcoin bulls and similarly biased predictions.
      \item \textbf{Prediction aggregators}: Metaculus, Good Judgement Open, \dots
    \end{itemize}

    \item Get forecasts from prediction aggregators, prediction markets, and finance websites.
    \item Get European option prices with expiry date as close as possible to 2025-01-01 and multiple strike prices close to 90k.
\end{itemize}
\paragraph{Notes}

\

\noindent This is tricky because big platforms favour questions like 70k (closer to the current price) or 100k (a ``rounder'' number). It also does not help that there are a lot of extremely low-quality predictions about Bitcoin prices.

The trick to doing alright is to not be thrown off by the exact number of 90k and to find forecasts like \url{https://kalshi.com/markets/btcmaxy/how-high-will-bitcoin-get-this-year} and \url{https://www.metaculus.com/questions/3820/bitcoin-extremes-will-1-bitcoin-be-worth-100000-or-more-before-2025/}. At the very least they can serve as lower bounds.

The trick to doing \emph{well} is to think outside the box (or remember very standard financial mathematics arguments) and remember that under certain assumptions like prices following a geometric Brownian motion we can use the \href{https://en.wikipedia.org/wiki/Reflection_principle_(Wiener_process)}{reflection principle} to infer the desired probability from the probability of the BTC price exceeding USD90k \emph{at the end of the year}. This probability, in turn, can be inferred from (European) option prices via the Breeden-Litzenberger formula.\footnote{For example, the current (as of August 2, 2024) CME call option prices for strike prices 89k and 91k are 3520 and 3245, respectively. (See \url{https://www.cmegroup.com/markets/cryptocurrencies/bitcoin/bitcoin.quotes.options.html}.) So, assuming a risk-free annualised interest rate of 5\%, we have $\mathbb P(\text{BTC}_\text{2025-01-01}>\text{USD}\,90k) = -e^{{150\over365}0.05}{\partial\over\partial K}C(90k,2025-01-01)\approx -e^{{150\over365}0.05}{3245-3520\over91k-89k}\approx 14\%$.} The agent does not need to reason about any of this, but it should, ideally, realise that option prices are very useful related forecasts in a sense.

% \subsection{Epidemiological forecasting}

% \subsection{Find the cumulative number of excess deaths per million people in the UK population by 2023, according to OWID}
% \paragraph{Ideal agent trace outline:}
% \begin{itemize}
%     \item Recall\footnote{GPT-4o, for example, is aware of this fact.} that OWID stores data on GitHub.
%     \item Try a direct Google search like \texttt{OWID GitHub cumulative number of excess deaths per million people}.
%     \item Read \url{https://github.com/owid/covid-19-data/blob/master/public/data/README.md}
%     \item Click the link next to \texttt{excess\_mortality\_cumulative\_per\_million} to get to \url{https://github.com/owid/covid-19-data/tree/master/public/data/excess_mortality}.
%     \item Note that the column \texttt{cum\_excess\_per\_million\_proj\_all\_ages} records the ``cumulative number of excess deaths per million people in the population; cumulated starting 1 January 2020''---just what we are looking for.
%     \item Download \url{https://raw.githubusercontent.com/owid/covid-19-data/master/public/data/excess_mortality/excess_mortality.csv}.
%     \item Write a Python script to read off the value of \texttt{cum\_excess\_per\_million\_proj\_all\_ages} in the right row (i.e. where the country is \emph{United Kingdom} and the date is \emph{2023-01-01}) or just read it off the \texttt{csv} file's contents directly.
% \end{itemize}
% \paragraph{Notes}

% \

% \noindent A fairly straightforward task for humans and truly multi-modal agents as they can just access the corresponding Our World in Data page\footnote{\url{https://ourworldindata.org/grapher/cumulative-excess-deaths-per-million-covid}}, click on Chart, edit Countries and Regions to include the UK, and read it off a graph, hovering over the right date. Alternatively, one can download a \texttt{.csv} file from the aforementioned page or from OWID's Github data repository\footnote{\url{https://github.com/owid}}.

\subsection{Find the date of the earliest 2009-2010 H1N1 seroprevalence study}
\paragraph{Ideal agent trace outline:}
\begin{itemize}
    \item Try a direct Google search for something like \texttt{first published seroprevalence study for 2009-2010 H1N1 (swine flu)} (this is unlikely to be successful)
    \item Formulate a plan to look for review articles with a hope of at least one of them citing the first published seroprevalence study
    \item Find review articles like \emph{Seroprevalence to Influenza A(H1N1) 2009 Virus—Where Are We?}\footnote{\url{https://www.ncbi.nlm.nih.gov/pmc/articles/PMC3147351/}}
    \item Go through their bibliographies to find references like \emph{McBryde et al.: ``Early transmission characteristics of influenza A(H1N1)v in Australia: Victorian state, 16 May - 3 June 2009"} and \emph{Chen H. et al.: ``Serologic survey of pandemic (H1N1) 2009 virus, Guangxi Province, China"}
    \item Look up these studies to
          \begin{itemize}
              \item verify they are actually seroprevalence studies for 2009–2010 H1N1 and
              \item when \emph{exactly} they were published
          \end{itemize}
    \item Infer the answer from the resulting list of studies with exact publication dates
\end{itemize}
\paragraph{Notes}

\

\noindent This is a difficult task because while finding the \emph{first} published seroprevalence study may not require an exhaustive search of all such studies, it still requires a ``fairly exhaustive'' search of all such studies in the beginning of the outbreak and current search engines (including Google Scholar, Semantic Scholar, etc.) don't seem to implement this functionality. Searching Google Scholar for ``H1N1 seroprevalence'' and sorting results by date instead of relevance, for example, currently returns a meager \emph{two}(!) papers (instead of the roughly 9,780 results when sorting by relevance).

Moreover, at the time of writing Google's answer box for the query \texttt{first 2009-2010 H1N1 (swine flu) seroprevalence study publication date} (and similar queries) currently yields the very misleading ``\emph{Several serological studies have estimated attack rates, comparing samples from before and after the pandemic (Table 2). The first report was published by Miller et al. in March 2010 (27), presenting the results of United Kingdom samples taken in August and September 2009 after the first wave of the pandemic.}''.

% \subsection{Private equity research}

\subsection{Estimate the number of paying ChatGPT users across all tiers excluding Team as of June 2024}
\paragraph{Ideal agent trace outline:}
\begin{itemize}
    \item Access \url{https://openai.com/chatgpt/pricing/} and note that ChatGPT has three paid tiers: Plus, Team, and Enterprise
    \item Find the suspicious-looking page \url{https://backlinko.com/chatgpt-stats} and note the claim that ChatGPT Plus had 3.9 million US subscribers as of March 2024.
    \item Verify that the number is credible by
          \begin{itemize}
              \item Accessing the linked source \url{https://www.yipitdata.com/resources/blog/investor-webcast-20240509-ai-cloud}
              \item Signing up for the webcast, reading the email triggered by the signup, and following the link therein
              \item Watching the webcast recording and reading 3.9M off from the relevant chart when it appears (the number does not appear anywhere on the web page, nor do the speakers mention it --- it must be read from the chart appearing in the video)
          \end{itemize}
    \item Realize that ChatGPT Plus subscribers likely grew between March and June, and search for earlier figures in order to estimate growth.
    \item Find \url{https://a16z.com/the-economic-case-for-generative-ai-and-foundation-models/} and note the claim that ChatGPT Plus had 2 million US subscribers as of July 2023.
    \item Choose a reasonable growth model and use it to estimate the number of ChatGPT Plus subscribers as of June 2024 based on the two figures.
    \item Adjust from US to global subscribers by using the figures from \url{https://appfigures.com/resources/insights/20240405} and \url{https://appfigures.com/resources/this-week-in-apps/20231006?f=2} in a reasonable way (e.g. taking a weighted average).
    \item Find a reliable source for the statement from Brad Lightcap (COO) that ChatGPT Enterprise had more than 600,000 users as of April 4, 2024.
    \item Find a reliable source for the statement also from Lightcap that Enterprise has 150,000 users as of January 11, 2024.
    \item Choose a reasonable growth model and use it to estimate the number of ChatGPT Enterprise subscribers as of June 2024 based on the two figures.
    \item Final answer is sum of Plus and Enterprise subscribers.
\end{itemize}

\subsection{How many research scientists are there at OpenAI?}
\paragraph{Ideal agent trace outline:}
\begin{itemize}
    \item Try to access \url{openai.com} and notice that it doesn't list current research scientists after exploring reasonable options.
    \item Try to access LinkedIn and notice that it is not bot-friendly either.
    \item Google something very direct like \texttt{research scientists at openai list} just in case someone happened to compile a list like this already and find some incomplete (and not up-to-date) lists like
          \begin{itemize}
              \item \url{https://research.com/university/openai}
              \item \url{https://www.adscientificindex.com/?university=OpenAI} (at this point it should notice that googling for \emph{rankings} of research scientists at OpenAI is an option because people like to rank just about everything)
              \item \url{https://www.reddit.com/r/MachineLearning/comments/404r9m/ama_the_openai_research_team/}
          \end{itemize}
          but none of these lists are exhaustive and all of them risk being outdated, so we would need to double-check every single person to see whether they still work at OpenAI.
    \item Research scientists tend to write papers, so look for papers by OpenAI and check whether the authors still work for OpenAI and whether they are actual research scientists.
          \begin{itemize}
              \item Finding the GPT-4 technical report paper\footnote{\url{https://arxiv.org/abs/2303.08774}}, it should notice the large number of authors (over 200) and wonder whether all of them are research scientists.
          \end{itemize}
    \item At this point we should have a very long list of people who worked at OpenAI at some point and many of which should be research scientist. Case-by-case checking (potentially merely subsampling) should now yield a list of current research scientist at OpenAI, missing out only on newly hired research scientists who have not coauthored any papers yet.
\end{itemize}
\paragraph{Notes}

\

\noindent This question is difficult for (law abiding) agents since LinkedIn is not open to bots and OpenAI's website does not list their current employees, let alone their research scientists. And even if LinkedIn was accessible to bots, LinkedIn information is not always up-to-date, since (on 2024-06-19) Ilya Sutskever, for example, is still listed as a research scientist at OpenAI despite his departure in May 2024.

\subsection{Track down original source for a blog post claim about number of paying ChatGPT mobile users.}
\paragraph{Ideal agent trace outline:}
\begin{itemize}
    \item Visit \url{https://nerdynav.com/chatgpt-statistics/} (the page that the instructions got the original claim from) and see if it cites its sources.
    \item Realising that it merely cites TechCrunch ``The number of ChatGPT Plus subscribers is estimated between 230,000-250,000 as of October 2023 (\emph{based on app revenue} reported by TechCrunch).'' (emphasis ours), notice that we should not look these exact numbers (230,000 and 250,000), but for (monthly) app revenue instead. So google something like \texttt{ChatGPT mobile app revenue site:techcrunch.com October 2023}.
    \item Check results that mention something about app revenue in a given month, preferably October 2023; alternatively, an earlier month. Total app revenue would not allow us to infer the numbers of subscribers in October 2023.
    \item This should yield several results, all of which end up pointing to either \url{https://techcrunch.com/2024/05/20/chatgpts-mobile-app-revenue-saw-biggest-spike-yet-following-gpt-4o-launch/} or \url{https://techcrunch.com/2023/10/09/chatgpts-mobile-app-hit-record-4-58m-in-revenue-last-month-but-growth-is-slowing/}.
    \item Notice that the former (more recent) article only mentions numbers for 2024, whereas the latter mentions numbers for September 2023 and month-on-month growth rates; so focus on the latter.
    \item Notice that the latter cites appfigures.com (no specific link provided) for app revenue growth decline. It is not clear whether they also got the data for number of downloads and gross app revenue from appfigures.com, so try to find them.
    \item Googling \texttt{Appfigures ChatGPT mobile app revenue subscribers September 2023} we should find \url{https://appfigures.com/resources/insights/20231006?f=2}, which appears to be the original source.
    \item Failing that (googling reasonable queries like \texttt{site:appfigures.com ChatGPT gross app revenue 2023 September} does not return any relevant results), we should consider the possibilities that
    \begin{itemize}
        \item [(a)] TechCrunch bought an appfigures.com subscription and we cannot access this data,
        \item [(b)] it is hidden in some graph,
        \item [(c)] we just cannot find it.
    \end{itemize}
\end{itemize}
Assuming it is only (b) an agent without visual capabilities should just return \url{appfigures.com} as original source with an appropriate caveat, whereas an agent with visual capabilities ought to persevere as follows:
\begin{itemize}
    \item In case it is hidden in some graph, let us simply try \texttt{site:appfigures.com ChatGPT} and check out various pages until we find \url{https://appfigures.com/resources/insights/20240802?f=4}, which features a graph of monthly \emph{net} revenues from App Store and Google Play, i.e. app revenue after the 30\% cut that both app stores take.
    \item From the graph we can infer that September's \emph{net} revenue from both app stores was roughly \$3.18m\footnote{Take a screenshot and compute the fraction of vertical pixels in the bar for September 2023 vs pixels from 0 to \$5m. This gives a value of $3.18m\approx {70\over110}\times5m$.}.
    \item Look up how much commission App Store and Google Play take: Both apparently happen to take 30\%, so we do not need to look up how much revenue came from which and can verify that the \emph{gross} revenue is thus roughly ${3.18m\over0.7}\approx4.54m$ which is close enough to the 4.58m (``almost 4.6m'') mentioned by TechCrunch.
    \item Hence it is fair to accept this as a original source. (Although just returning \url{appfigures.com} seems equally valid, given that this post was clearly published after the blog post in question.)
\end{itemize}
\paragraph{Notes}

\

\noindent Specifically, we ask about the primary source for the claim on \url{https://nerdynav.com/chatgpt-statistics/} for ChatGPT mobile having about 230,000 to 250,000 paying users as of October 2023.
This is potentially misleading because nerdynav.com cites TechCrunch as its source, but it does not specify how its claim follows from information in the TechCrunch article\footnote{TechCrunch (\url{https://techcrunch.com/2023/10/09/chatgpts-mobile-app-hit-record-4-58m-in-revenue-last-month-but-growth-is-slowing/}) cites \$4.6 million in gross revenue \emph{in September 2023}, as well as month-on-month growth rates having dropped from around 30\% to 20\% in the months leading up to September 2023. Together with the implicit assumption that ChatGPT plus costs (around) \$20 not only in the US, but also in other markets, we can then compute $230000= {4600000\over20} < {4600000\times1.1\over20}\approx250000$ paying users. So the relevant quantities to look for happen to be gross revenue in a past month and growth rate, not number of paying users!} (making it non-trivial what exactly to look for). Moreover, it only provides a link to a potentially relevant TechCrunch article in the subsequent paragraph in the context of presenting different information.

To make matters worse, TechCrunch published other ostensibly relevant articles\footnote{E.g. \url{https://techcrunch.com/2024/08/07/chatgpts-mobile-app-just-had-its-biggest-month-yet/}} since then, but they provide more up-to-date numbers. So agents should realise that they cannot be relevant to a claim about October 2023.

\newpage

\section{Detailed remarks on agent performance} \label{sec:detailed-performance-remarks}

% \subsection{Geopolitical forecasting}

\subsection{Compile a list of Chinese AI labs that have carried out training runs using over 1e24 FLOPs}
\begin{itemize}
    \item [\newclaude]
        \begin{itemize}
            \item Planning agent:
                \begin{itemize}
                    \item Just googled \texttt{Chinese AI labs training runs over 1e24 FLOPs} and didn’t find any information about training runs >1e24 FLOPs. Then googled \texttt{Chinese AI models over 1e24 FLOPs}. Didn’t find anything either and concluded that there aren’t any training runs as of October 2023.
                \end{itemize}
            \item ReAct agent:
                \begin{itemize}
                    \item Googles texttt{Chinese AI labs training models using over 1e24 FLOPs}. Doesn’t find any answers. Formulates a thought “Since the initial search didn't provide specific information, I should identify the leading Chinese AI labs and investigate whether any have conducted training runs exceeding 1e24 FLOPs by October 2023.”
                    \item Gives up (reports that there aren’t any labs).
                \end{itemize}
            \item ReAct agent with subtasks:
                \begin{itemize}
                    \item Finds a bunch of models and tries to estimate the number of FLOPs for those. Has to make assumptions in its calculations, so the actual numbers are often not correct. The actual calculations look good though.
                    \item It doesn’t find all relevant labs/models.                     
                \end{itemize}
            \item Planning agent with subtasks:
                \begin{itemize}
                    \item Agent failed due to an \texttt{Invalid prompt: your prompt was flagged as potentially violating our usage policy. Please try again with a different prompt.} error. 
                \end{itemize}
        \end{itemize}
    \item [\newclaude]
        \begin{itemize}
            \item Planning agent:
                \begin{itemize}
                    \item Sets out to google \texttt{Chinese AI labs training runs FLOPs} to find relevant articles/reports. It plans to then track down their sources and compile the list we asked for. Failing that, it would broaden its search to global AI labs and filter for Chinese ones.
                    \begin{itemize}
                        \item Note that it’s smart enough not to try to look for the fairly arbitrary threshold directly \emph{and} it’s smart enough to realise it may have to broaden its scope to global AI labs!
                    \end{itemize}
                    \item It finds \url{https://epochai.org/blog/tracking-large-scale-ai-models} as well as \url{https://epochai.org/data/large-scale-ai-models} pretty much right away and reads them to find Chinese AI labs which with large training runs, but fails to realise that it should download the dataset (the excerpts it read from these pages do mention data being available in a csv file) and ends up returning nothing.
                \end{itemize}
            \item ReAct agent:
                \begin{itemize}
                    \item Starts with naive Google searches like \texttt{Chinese AI labs training runs over 1e24 FLOPs}, but quickly changes them to more general ones, hence finding several relevant Epoch pages and reading them both.
                    \item It notes that it `would need to analyze the full dataset from Epoch AI or find more detailed technical reports from specific Chinese AI research institutions. At this point, I don't have enough information to provide an exhaustive list of Chinese AI labs that have conducted training runs over 1e24 FLOPs.` and gives up after one more Google search restricted to arxiv.org or paperswithcode.com.
                \end{itemize}
            \item ReAct agent with subtasks:
                \begin{itemize}
                    \item A subagent finds the Epoch dataset pretty much right away and uses it. However, in the task specification for this subagent the “exhaustive” got dropped, so it only replies with “five recent models” instead of the comprehensive list (9 models at the time of writing).
                \end{itemize}
            \item Planning agent with subtasks:
                \begin{itemize}
                    \item Sets out to create a list of Chinese AI labs to research big training runs for, as well as a clarification step to research what “over 1e24” even means (extremely large and probably enough to check generative AI and LLMs).
                    \item Literally compiled a .csv file with big players. Decent list of labs including some of the newer/less well-known labs like 01.AI, Zhipu, Baichuan, but missing Deepseek.
                    \item Tasked with essentially the main task, a subagent gives up on finding FLOP estimates for training runs and instead just reports a handful of models with their parameter count.
                    \item It then tries to estimate FLOP usage for a handful of models, but misses the relevant ones.
                \end{itemize}
        \end{itemize}
    \item [\gpt]
        \begin{itemize}
            \item Planning agent: 
                \begin{itemize}
                    \item Reads a wide range of articles in an attempt to generally survey the field and finds good information including newer/less well-known companies with big models like 01.AI, DeepSeek AI, Inspur, and many more, but accidentally deletes its own memory halfway through!
                    \item So instead of actually following up on these promising candidates, it ends up only looking into very generic ones like Baidu, BAAI, Tsinghua, etc.
                    \item Until the very end it only uses naive Google searches like \texttt{Chinese AI labs training runs over 1e24 FLOPs}.
                \end{itemize}
            \item ReAct agent:
                \begin{itemize}
                    \item Accidentally finds the Epoch dataset (csv) directly (i.e. without even seeing the Epoch page) when looking into BAAI training runs right away, but decides not to use it. 
                    \item Instead it searches for a few queries like \texttt{Chinese AI labs training runs over 1e24 FLOPs BAAI Alibaba Tencent} in the hope of finding an answer directly and then gives up, reporting only BAAI (mistakenly).
                \end{itemize}
            \item ReAct agent with subtasks:
                \begin{itemize}
                    \item Starting by compiling a list of Chinese AI labs likely to have run big training runs it misses quite a few.
                    \item Curiously, a subagent finds and successfully uses the Epoch dataset to answer a subquestion about whether Alibaba ran a training run with over 1e24 FLOPs, but the agent does not change plans after seeing the mention of this dataset.
                \end{itemize}
            \item Planning agent with subtasks:
                \begin{itemize}
                    \item Comes up with a very incomplete list of big Chinese AI labs and researches estimates for their largest training runs.
                    \item Curiously, a subagent finds and successfully uses the Epoch dataset to answer a subquestion about whether Baidu ran a training run with over 1e24 FLOPs, but the agent does not change plans after seeing the mention of this dataset. 
                \end{itemize}
        \end{itemize}
        \item [\minigpt]
        \begin{itemize}
            \item Planning agent: 
            \begin{itemize}
                \item Googles naive queries like \texttt{Chinese AI labs training runs over 1e24 FLOPs} and then reads some pages with the intent to answer \texttt{Which Chinese AI labs have conducted training runs using over 1e24 FLOPs?} directly.
            \end{itemize}
            \item ReAct agent: 
            \begin{itemize}
                \item Gets stuck in an infinite loop googling \texttt{Chinese AI labs training runs over 1e24 FLOPs 2023 report details list}.
            \end{itemize}
        \end{itemize}
        \item [\llama]
            \begin{itemize}
                \item Planning agent: 
                    \begin{itemize}
                        \item Finds and reads \url{https://cdn.governance.ai/Trends_in_Chinas_LLMs.pdf} but when this does not show any Chinese AI labs that would qualify, it terminates and reports that there are none.
                    \end{itemize}
                \item ReAct agent:
                    \begin{itemize}
                        \item Comes up with a decent list of potential candidates by breaking down search to a very granular level (Chinese universities in East China, ... and searching IEEE Xplore/arXiv/Google Scholar/... using keywords like \texttt{large-scale computing}, \texttt{FLOPs}, \texttt{Chinese AI labs}), including Zhipu AI, Alibaba’s Qwen, Tsinghua, 01.AI (Yi), DeepSeek AI, and more.
                    \end{itemize}
                \item ReAct agent with subtasks:
                    \begin{itemize}
                        \item Googles for \texttt{Chinese AI labs training runs dataset} which shows the right Epoch page as the first result, but does not read it and instead continues to google queries like \texttt{Chinese AI labs with large-scale training runs} and when this does not surface any labs that would qualify, it quickly gives up and reports that there are none.
                    \end{itemize}
                \item Planning agent with subtasks:
                    \begin{itemize}
                        \item Looks for papers that mention FLOP counts, finds (only) the Falcon-180B one, checks if Chinese AI labs were involved and gives up quickly after it can’t find any involvement.
                    \end{itemize}
            \end{itemize}
    \end{itemize}

\subsection{Find original sources shedding light on the Sino-Russian relationship after their announcement of ``friendship without limits''}
\begin{itemize}
    \item [\oone]
        \begin{itemize}
            \item Planning agent:
                \begin{itemize}
                    \item Instantly reports a list of hallucinated links with (correctly) hallucinated claims; this includes their joint statement (the only correctly hallucinated link), some joint military drills, increased bilateral trade, increased use of national currencies reducing reliance on the dollar, and a long-term gas supply contract.
                \end{itemize}
            \item ReAct agent:
                \begin{itemize}
                    \item Found the Wikipedia article, but failed to read it, choosing to read the official announcement on the Kremlin’s website instead.
                    \item Continues to look for joint military exercises and then reports the (very incomplete) result.
                \end{itemize}
            \item ReAct agent with subtasks:
                \begin{itemize}
                    \item Looks for original sources for significant actions, treaties, etc., military cooperation, economic agreements, political disagreements/tensions, and official statements about changes since the announcement of their “friendship without limits”.
                    \item Finds all sorts of joint military exercises, various bilateral agreements (without reporting the source; only mentioning that sources for them can be found on official websites), economic agreements, articles on Xi warning Putin not to use nuclear attacks in Ukraine (and denials of this ever happening), and energy cooperation.
                \end{itemize}
            \item Planning agent with subtasks:
                \begin{itemize}
                    \item Plans to look into military drills, arms sales, trade agreements, energy deals, joint statements, speeches by Xi or Putin, UN voting records, border incidents, official complaints about each other (in the context of trade). Distilling this plan into concrete steps unfortunately gets rid of some of the details.
                    \item Some subtasks end up being too general like “Collect treaties and agreements signed between China and Russia”
                    \item Ends up replying with “Here is the link to Chinese/Russian press releases/UN documents/…, you can search for keywords like XXX to find relevant press documents” several times.
                    \item Finds several joint military drills, joint statements, gas supply agreements, space cooperation, maritime law enforcement cooperation, agreement that NATO shouldn’t expand further, opposed sanctions in the UN, differing stances on climate change in the context of G20.
                    \item Proposes searching in Russian and Chinese, but errors out before it can do so.
                \end{itemize}
        \end{itemize}
    \item [\newclaude]
        \begin{itemize}
            \item Planning agent:
            \begin{itemize}
                \item Agent came up with a good plan to look for general deals/agreements/treaties, as well as evidence for economic, military, and trade cooperation, but ended up mistakingly modifying its plan to get rid of these todo items.
                \item Agent ended up trying to find direct sources directly.
                \item Ends up hallucinating links in the final steps since it only stored takeaways, not links, in its memory.
            \end{itemize}
            \item Planning subtask agent:
            \begin{itemize}
                \item Does not explicitly look up the date when ``friendship without limits'' was announced, but works around it by asking ``since the announcement of ...'' instead.
                \item Comes up with a good plan and finds lots of interesting things, but struggles with subagents' responses not pointing to any sources.
            \end{itemize}
            \item ReAct agent:
            \begin{itemize}
                \item Failed to respect the output format (links to sources with excerpts/explanations why they are relevant), but gave a good summary of sources it found.
                \item Trying to look for government websites, it initially tried to restrict its searches to \texttt{site:.gov OR site:.org}; a curious mistake since even \minigpt knows to look for \url{kremlin.ru} and \url{.gov.cn} sites.
                \item It also googled for the (exact) date of the announcement of friendship without limits as opposed to just remembering February 2022 like other LLMs (including the small \minigpt).
                \item It overlooked the relevant Wikipedia page (\url{en.wikipedia.org/wiki/China%E2%80%93Russia_relations}) where it could easily have found lots of links to original sources. Like other LLMs it exhibited a bit of tunnel vision for original sources.
                \item It inferred a lot from search result snippets as opposed to actually reading a lot of the websites in question.
            \end{itemize}
            \item ReAct agent with subtasks:
            \begin{itemize}
                \item Starts by coming up with proper plan: Find out when exactly the announcement was made, identify events/actions after the announcement, find original source for each, verify originality by checking links in each source, compile them into a list.
                \item Funnily enough it already knows that this announcement was made in February 2022, so the first set of tasks to be run in parallel include figuring out when the announcement was made \emph{and} looking for events, treaties, or deals \emph{since February 2022}.
                \item Is much better at viewing and putting things in context. Some examples:
                \begin{itemize}
                    \item It notices that despite expressed intentions to cooperate in advanced technologies, there is no evidence of joint AI projects/
                    \item It speculates that the reason it could not find official sources for transfer of military technology and semiconductor trades is due to the strategic nature of such collaborations instead of trying increasingly harder to solve this likely unsolveable task.
                    \item It puts (post 2022) joint military exercises into perspective by pointing out that they already started in 2003, but have grown in frequency and complexity.
                \end{itemize}
                \item Explicitly searches also for statements/documents indicating \emph{limitations/tensions} in their partnership.
            \end{itemize}
          \end{itemize}
          \item [\gpt]
          \begin{itemize}
              \item Planning agent: 
              \begin{itemize}
                \item Agent came up with an interesting idea of not even trying to find original sources directly, but simply google things like \texttt{China Russia friendship without limits key events treaties deals actions 2023} and find original/primary sources after extracting insights. (Unfortunately, it ended up exceeding the maximum number of tool calls before it got to this part.)
                \item Since the search for non-primary sources was not very sophisticated, we ended up with a bunch of wishy-washy ``insights'' like ``China has kept a distance as Russia and North Korea grow closer'', as well as outdated facts like the resolved border dispute in the 2000s (way before the ``friendship without limits'' announcement), etc.
                \item Believes that returning the very first set of Google search results already fulfils the task and ends early.
              \end{itemize}
              \item ReAct agent:
              \begin{itemize}
                  \item Initially tried the naive Google search \texttt{Sino-Russian relationship after friendship without limits official government announcements}, but realised this does not yield a lot of original sources, and then tried the much better \texttt{Sino-Russian relationship official government announcements site:gov.cn OR site:kremlin.ru}.
                  \item Read a few documents it found with this search and almost immediately reported the result.
              \end{itemize}
              \item ReAct agent with subtasks:
              \begin{itemize}
                  \item Implicit in its first set of subtasks is a plan to look for information about military and economic agreements.
                  \item Still somewhat aimless: Some of the first subtasks it comes up with are like ``Find \emph{an} official statement or document from the Russian/Chinese Ministry of Foreign Affairs regarding the Sino-Russian relationship post the 'friendship without limits announcement'.''.
                  \item When coming up with subtasks, it does not realise that it should ask for links to its sources as well to complete its main task, but instead adds the task ``Locate the original source of any official treaty or agreement signed between China and Russia after the 'friendship without limits' announcement.'' This kind of implicit reference to other subtasks being run in parallel is explicitly discouraged in the prompt for coming up with subtasks.
                  \item Get sidetracked by random bits and pieces surfaced during working on its first set of subtasks like the ``Joint Statement on the Plan to Promote Key Elements of Chinese-Russian Economic Cooperation until 2030'', when it should instead have continued to prioritise breadth-first search.
              \end{itemize}
              \item Planning agent with subtasks:
              \begin{itemize}
                  \item Attempts to directly find original sources for developments in key areas like the economics, diplomacy, global organisations, and military cooperation. 
                  \item Splits up tasks like ``Find sources related to economic deals, ...'' into ``Find sources related to economic deals from 2000 to 2010'' and ``Find sources related to economic deals from 2011 to 2020'' despite explicitly being instructed not to do so.
                  \item Found the most (out of all agents) on dual-use Chinese exports to Russia and other military deals.
                  \item Also fails to keep track of sources.
              \end{itemize}
          \end{itemize}
    \item [\llama]
          \begin{itemize}
              \item Planning agent:
              \begin{itemize}
                  \item Comes up with a decent plan: First look for general news/updates on Sino-Russian relationship, identify key events such as military exercises, economic deals, diplomatic meetings, and then track down original sources for each of those.
                  \item Fails to mark the first step on its plan as DONE and hence gets stuck in a loop repeatedly googling the same query.
              \end{itemize}
              \item ReAct agent:
              \begin{itemize}
                  \item Does two fairly general searches, reads one article each, and returns these two links without an explanation of why they are relevant and without having tried to find where these claims originated from (one, in particular, is not a government website).
              \end{itemize}
              \item ReAct agent with subtasks:
              \begin{itemize}
                  \item Simply remembered that the agreement was signed in 2022 without looking it up; a bit sloppy because it only remembered 2022, not the month, let alone the day.
                  \item Made a plan (implicit in its choice of subtasks) to look for joint military exercises, economic agreements/trade deals, forums, and instances where the two countries supported each other's defense ecosystems/economic growth and \emph{then} decided to look for sources, as opposed to restricting its search to original sources from the beginning.
                  \item Tries to contact Chinese embassy by phone, etc., despite explicitly being instructed not to try to contact people/organisations.
                  \item Picks up on several promising-looking agreements that later turn out to have been signed before the agreement of ``friendship without limits'' (in 2014 and 2018, for example), but fails to back off upon learning that.
                  \item Gets stuck trying to track down sources, e.g. repeatedly asks to find sources for ``Find the full text of the China-Russia Nuclear Industry Cooperation Agreement signed in 2023 and provide the URL or text.''; this is despite being explicitly instructed not to repeat approaches that have not worked out before.
              \end{itemize}
              \item Planning agent with subtasks:
              \begin{itemize}
                \item Comes up with a good plan of researching and listing significant events, actions, treaties, deals, and announcements, then collecting news articles related to them, tracking down original sources, verifying sources(?), explaining their relevance, and distilling it into the format the task asks for.
                \item Seems a bit gullible and noticeably un-opinionated by interpreting every joint military exercise to strengthen ties and noticing that ``the exact details and implications of these agreements are not specified in the search results provided''.
                \item Very self-critical: Tasking itself with listing significant military cooperation agreements, etc. and having replied with some examples, it does not think it completed the task successfully. As a result it errors out before reporting its findings.
              \end{itemize}
          \end{itemize}
    \item [\minigpt]
          \begin{itemize}
              \item Planning agent: 
              \begin{itemize}
                \item Fails to formulate plan in the intended way, but comes up with a decent plan nonetheless. 
                \item Correctly remembers government websites and relevant dates like when Russia and China announced their friendship without limits.
              \end{itemize}
              \item ReAct agent: 
              \begin{itemize}
                \item Gets stuck in a look googling \texttt{Sino-Russian military cooperation agreements 2023 site:.gov.cn OR site:.gov.ru}. Note that \texttt{.gov.ru} will not find many interesting sources (as opposed to \texttt{kremlin.ru}).
              \end{itemize}
          \end{itemize}
    \item [\minillama]
          \begin{itemize}
            \item Planning agent:
            \begin{itemize}
                \item Hallucinates, often returns JSON schemas instead of JSON objects conforming to a given schema, ``remembers'' incorrect URLs for Russian and Chinese government websites.
            \end{itemize}
            \item Planning agent:
            \begin{itemize}
                \item Repeatedly reads \url{https://www.isdp.eu/75-years-of-china-russia-relations-indeed-a-no-limits-partnership}
            \end{itemize}
          \end{itemize}
 
\end{itemize}

% \subsection{Financial forecasting}

\subsection{Find forecasts helpful for predicting whether BTC will trade at over USD90k before 2025}
\begin{itemize}
    \item [\oone]
        \begin{itemize}
            \item Planning agent:
                \begin{itemize}
                    \item Immediately reports the result without even Googling once. Mentions Metaculus, Finder.com, Bloomberg Intelligence, JPMorgan Chase, and the Stock-to-Flow (S2F) Model as good sources.
                \end{itemize}
            \item ReAct agent:
                \begin{itemize}
                    \item Sets out to check Metaculus, prediction markets, financial analysts' reports, and credible financial news outlets. 
                    \item Somehow fails to find the 100k Metaculus forecast.
                    \item Thinks of options implied probability! At first it only googles \texttt{Bitcoin options implied probability \$90k before 2025} directly, but then it writes a Python script fetching options data from deribit.
                    \item Instead of using the Breeden-Litzenberger formula, it uses a Black-Scholes approach with some guesstimated constants for the risk-free interest rate and volatility, implying a roughly 17\% probability.
                \end{itemize}
            \item ReAct agent with subtasks:
                \begin{itemize}
                    \item Sets out to look for related forecasts on Metaculus, prediction markets like Augur and Polymarket, financial analysts' reports or institutional forecasts, surveys or expert opinions, and relevant academic publications.
                    \item It fails at finding relevant academic publications, replying with \texttt{As of my knowledge cutoff in October 2023, there are no specific academic publications or statistical models that predict Bitcoin's price reaching \$90,000 before 2025 with detailed probabilistic forecasts and methodologies. [...]}
                    \item Finds the 100k Metaculus forecast, but incorrectly interprets it to imply a 20\% chance, when the community prediction is at 10\% (at the time of writing).
                    \item Found the 90k Manifold Markets market, but also infers the incorrect implied probability from an outdated Google snippet.
                    \item Estimates interest rate and volatility and looks up the current price to calculate the probability via Black-Scholes.
                    \item Also picks up some lower quality sources and finishes its report.
                \end{itemize}
            \item Planning agent with subtasks:
                \begin{itemize}
                    \item Thinks of Metaculus, GJO, financial analysts, investment banks, cryptocurrency research firms, as well as analysing trends.
                    \item Doesn’t find probabilistic forecasts from financial analysts, etc.
                    \item Finds Manifold Markets 90k market (again with incorrect probability inferred from outdated Google snippets), as well as some less related probabilistic predictions from crypto sources.
                    \item Fails to find Metaculus' 100k question at first, but finds it during its second attempt. It only summarises good arguments in the comment section though, without reporting back with the current community prediction (or any number for that matter).
                    \item Finds a point estimate from Finder.com's Cryptocurrency Predictions Report and some outdated report making forecasts for 2023 (given that o1’s knowledge cutoff is around October 2023 and the current date is not inserted in every single LLM call we make, there is a good chance it never realised that this is horribly outdated).
                    \item Performs some time series and Monte Carlo analyses.
                    \item Finds the 100k Metaculus question \emph{and correctly infers the current community prediction}.
                    \item Finds a lot of point forecasts from Bloomberg, ARK Research, etc.
                    \item Gets lost in overly specific subtasks like \texttt{Assess the impact of geopolitical events and tensions up to October 2023 on the price of Bitcoin}, \texttt{Evaluate technological advancements and scalability solutions for Bitcoin up to October 2023 that could impact its price}, \texttt{Identify and summarize significant regulatory changes and governmental actions regarding cryptocurrencies up to September 12, 2024, that could influence the likelihood of Bitcoin reaching \$90,000 before January 1, 2025}, etc.
                \end{itemize}
        \end{itemize}
    \item [\newclaude]
        \begin{itemize}
            \item Planning agent:
                \begin{itemize}
                    \item Starts with the odd plan of starting with a generic Google search and identifying reputable forecasting platforms/cryptocurrency analysis websites from the search results.
                    \item If it cannot find probabilistic forecasts it plans to compute the fraction of price predictions for ``BTC $\geq$90k before 2025'' and ``use this percentage as an approximation of the probability'', hinting at a poor understanding of probabilities as well as price predictions.
                    \item Spends a lot of time looking at odd cryptocurrency websites and getting confusing price predictions ``for 2025'' with forecasts about whether the price will exceed 90k before 2025.
                \end{itemize}
            \item Planning subtask agent:
                \begin{itemize}
                    \item Completely misses the point when it sets out to ``research current BTC price and recent trends'', ``gather historical BTC data'', ``investigate macroeconomic factors'', and ``examine technological developments in the crypto space''.
                    \item Ends up trying to write its own simulations that occasionally crash and hardly ever succeed.
                    \item Also produces ``rather optimistic'' content like adoption rate comparisons between ``internet, smartphone, and Bitcoin'' (tacitly assuming sigmoidal growth until 100\% adoption for all three at somewhat different speeds)
                    
                    \includegraphics[width=0.8\textwidth]{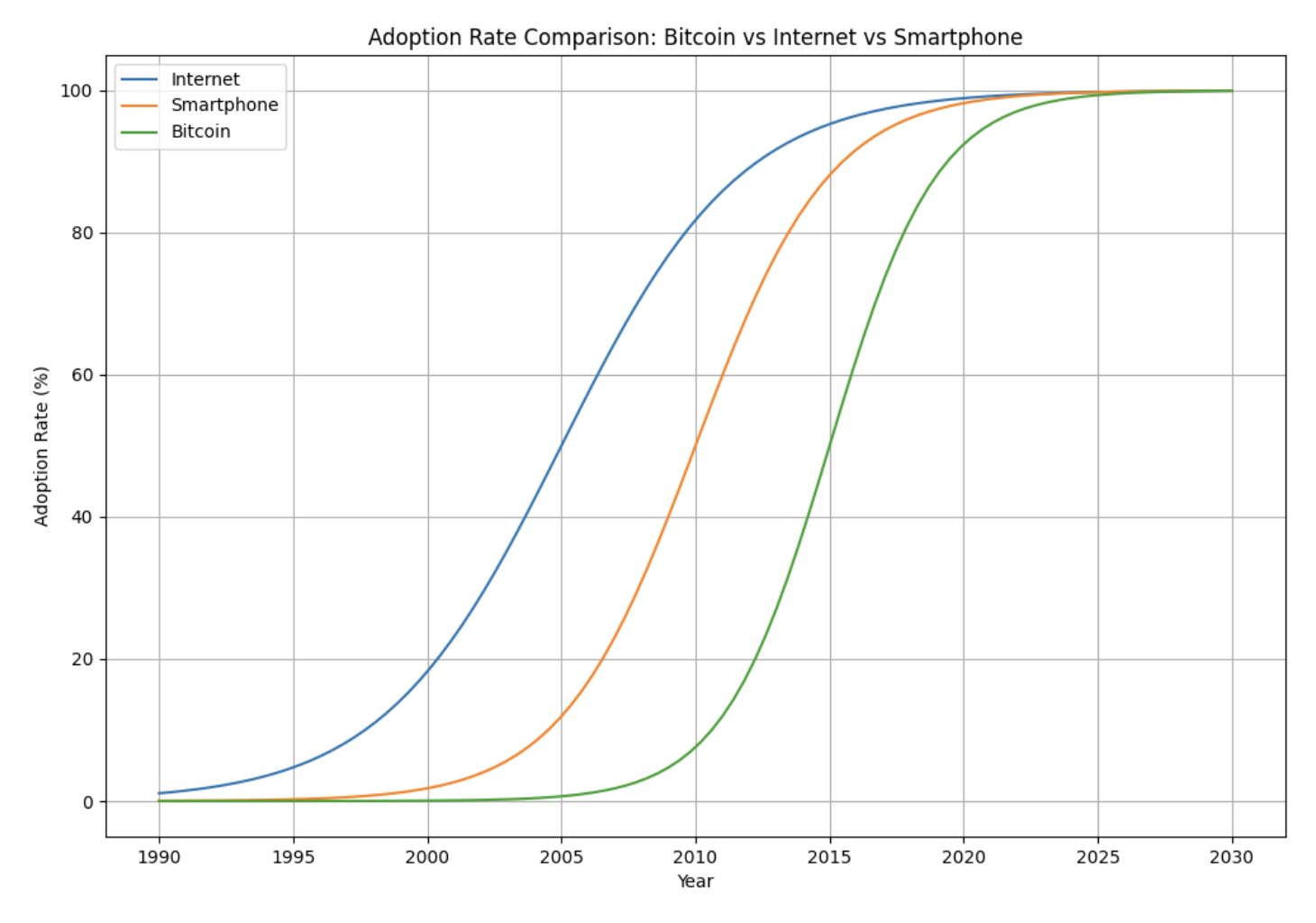}
                \end{itemize}
            \item ReAct agent:
                \begin{itemize}
                    \item Tries to look for sources (forecasting aggregators, specifically), but Google mostly shows low-quality BTC pages, so it zeros in on Forbes and a low-quality source with the caveat of searching for more specialised prediction markets or crypto forecasting platforms if that does not work out.
                    \item Seems to understand the intricacies of peak price as opposed to the price at a given point in time as well as peak 2024 as opposed to `in 2025'.
                    \item Finds an exact match on Manifold Markets (\url{https://manifold.markets/pa/will-the-price-of-bitcoin-hit-90000}), but infers an incorrect probability (30-35\% despite the current price indicating a 28\% chance).
                    \item Looks for a similar question on Kalshi and finds its question on whether 100k will be reached.
                    \item Looks for Metaculus.
                \end{itemize}
            \item ReAct agent with subtasks:
                \begin{itemize}
                    \item Starts by looking for the current BTC price, identifying financial/crypto websites with forecasts, as well as prediction markets and forecasting platforms.
                    \item Gets a bit confused with dates: “As of the most recent data available from CoinDesk, the current price of Bitcoin (BTC) is \$59,545.21. However, it's important to note that this is not the price for 2024-08-21, as that date is in the future. The price provided is the most up-to-date information available at the time of the query, and cryptocurrency prices are highly volatile and subject to change.” (This was run on Aug 21, 2024.)
                    \item Does some superficially relevant research on “recent market trends and significant events affecting Bitcoin’s price in the past year”, concluding that conditions look optimistic.
                    \item Would have found the right Polymarket market if they did not block bots, but only ended up finding the closely related Metaculus question.
                    \item It should really have attempted another Google search upon noticing that querying Polymarket directly did not return anything.
                    \item Wrote an incorrect Monte Carlo simulation using made-up (not completely wrong, but still quite incorrect) current value and historical volatility, yielding a probability of 61\%.
                    \item Continues to attempt Monte Carlo simulations, comparing growth rates to what would be needed to exceed 90k and trips up badly. (E.g. by reporting negative volatility.) While it gets the Monte Carlo simulation right on its third try, it continues to be bad at maths/statistics as evidenced by a confidence interval of 36\% to 105\%(!) for the probability.
                    \item Ends up getting lost in tasks of increasing complexity like writing entire reports itself.
                \end{itemize}
        \end{itemize}
    \item [\gpt]
        \begin{itemize}
            \item Planning agent: 
                \begin{itemize}
                    \item Mediocre plan of just using a single generic Google search and taking it from there.
                    \item Ends up finding and (uncritically) recording exclusively low-quality results.                    
                \end{itemize}
            \item ReAct agent:
                \begin{itemize}
                    \item Starts by looking for reliable sources of forecasts for BTC, which, for some reason, includes academic journals.
                    \item Ends up checking whether there are forecasts for that very specific forecasting question on arXiv, Reuters, and some more random low-quality sources, none of which are both reliable and relevant.
                \end{itemize}
            \item ReAct agent with subtasks:
                \begin{itemize}
                    \item Seems gullible as it thinks “cryptocurrency market research firms and expert opinions” will yield reliable forecasts. 
                    \item Starts looking for price forecasts (and probabilistic predictions), so it seems to be somewhat confused about the intricacies of the question asking about the maximal value of BTC before 2025.
                    \item Understands the difference between “deterministic predictions” (presumably it means point estimates like price predictions?) and probabilistic predictions.
                    \item It happens to stumble upon the exact match on Manifold Markets (\url{https://manifold.markets/pa/will-the-price-of-bitcoin-hit-90000}), but overlooks a similar question on Metaculus.
                    \item Noticeably does not bother with clearly low-quality sources.
                \end{itemize}
            \item Planning agent with subtasks:
                \begin{itemize}
                    \item Starts by looking for forecasts on financial websites, expert analyses, and prediction markets/aggregators and finds the closely related Metaculus question right away.
                    \item Gets confused with dates (returns forecasts about the end of 2025).
                    \item Simple searches anchor it to random investors that mention 90k specifically, which gives a very skewed set of forecasts.
                    \item Does not pay attention to low-quality websites.
                    \item It gets lost in summarising information from the first few sources it found over and over again.
                    \begin{itemize}
                        \item Interestingly, this makes it the only LLM with this architecture to look at websites not only to extract probabilities, but also to take relevant insight from the comment section, etc.
                    \end{itemize}
                \end{itemize}
        \end{itemize}
    \item [\minigpt]
        \begin{itemize}
            \item Planning agent: 
                \begin{itemize}
                    \item Googles and finds Forbes, CoinCodex, Metaculus, Forbes, and CCN, all of which it deems to be credible sources.
                    \item Records its findings as TODO items.
                    \item Sometimes repeats the same actions, but moves on quickly (as opposed to getting stuck repeating the same action >10 times).
                    \item Reads the Metaculus result (\url{https://www.metaculus.com/questions/3820/bitcoin-extremes-will-1-bitcoin-be-worth-100000-or-more-before-2025}), but fails to record information it found, hence repeating the action of reading it over and over.
                    \item Infers a 30\% chance from BTC reaching \$100,000 before 2025, when the community prediction is 14\%.
                \end{itemize}
            \item ReAct agent: 
                \begin{itemize}
                    \item Keeps googling ``Bitcoin price prediction August 2024 expert analysis 90k before 2025'' and nothing else.
                \end{itemize}
        \end{itemize}
    \item [\llama]
        \begin{itemize}
            \item Planning agent: 
                \begin{itemize}
                    \item Unlike with other LLMs, this one actually thinks of several approaches like looking for experts/prediction models, as well as writing a Monte Carlo simulation based on historical data itself.
                    \item Unfortunately it doesn't get to any of these alternative approaches because it wastes all its time on trivial Google searches not getting anywhere.
                    \item Keeps looking for \texttt{predicted price of Bitcoin in 2025}, which completely misses the point.
                \end{itemize}
            \item ReAct agent:
                \begin{itemize}
                    \item Starts with the overly general Google search \texttt{reliable sources of probabilistic forecasts}, but notices the problem and restricts its search to sources for cryptocurrency/financial markets. Gives up upon not finding anything after one revised Google search.
                    \item Keeps using overly naive Google searches until it errors out due to repeatedly failing to adhere to a provided (simple) JSON schema.
                \end{itemize}
            \item ReAct agent with subtasks:
                \begin{itemize}
                    \item Sets out to look only at financial websites and forecasts by market analysts \& crypto experts. Somehow manages to try academic journals in the process again, however.
                    \item Repeatedly fails to find any probabilistic predictions and only reports qualitative measures like “very likely that…” based on point estimates.
                    \item Harder tasks simply are not finished often (if ever): E.g. for the task \texttt{Use machine learning models to predict future BTC price movements based on historical data, [...]} it ends up replying with the kind of machine learning models it would use.
                    \item Starts getting lost in random tasks like analysing Twitter sentiment and building more machine learning models before erroring out.
                \end{itemize}
            \item Planning agent with subtasks:
                \begin{itemize}
                    \item Wants to conduct a literature review on research papers, articles, etc. to understand historical trends, etc. for BTC, as well as analysing BTC through social media, and technical analysis, and many other approaches unrelated to finding existing probabilistic forecasts.
                    \item For the social media analysis its subagents just reply with \texttt{Create a Twitter Developer account to access the Twitter API for sentiment analysis}.
                    \item Ends up finding low-quality point forecasts from various crypto sources, but not a single probabilistic forecast.
                \end{itemize}
        \end{itemize}

\end{itemize}

\subsection{Use official government data to estimate how many Chinese have an annual disposable income exceeding 100,000 Yuan}
\begin{itemize}
    \item [\oone]
        \begin{itemize}
            \item Planning agent:
                \begin{itemize}
                    \item Plans to check NBS for the distribution and then googling for statistical reports/bulletins that provide data per income bracket.
                    \item Fails to update plan correctly: After a single Google search, it marks the first todo item as DONE and the second as FAILED.
                    \item Performs the usual log-normal calculation where it fails to account for fraction of the population not earning anything.
                    \item Wants to double-check with more government statistics, but finds outdated, unofficial posts where it mistakes income and disposable income, hence being very confused.                    
                \end{itemize}
            \item ReAct agent:
                \begin{itemize}
                    \item Starts with a plan to look for disposable income data and already anticipates potentially having to settle for percentile data, mean/median, or Gini coefficient and then having to fit a parametric model.
                    \item Finds mean and median right away and fits a log-normal distribution (without realising that a fraction of the population don’t earn anything, so that the disposable income distribution is not log-normal before removing this component).
                    \item Continues to look for income brackets to refine its approach and does the same log-normal calculation after breaking the population down into rural and urban households.
                    \item Googles for decile data, but stops because it doesn’t find anything that looks sufficiently official after one Google search.
                \end{itemize}
            \item ReAct agent with subtasks:
                \begin{itemize}
                    \item Sets out to check NBS for the disposable income distribution, to find the answer directly, check data on low, medium and high income classifications, and sources fitting parametric models to the income (not disposable income!) distribution.
                    \item Finds interesting (albeit somewhat outdated, e.g. from 2014) studies about how to model Chinese income distribution, claiming that the distribution is roughly log-normal in the bulk and Pareto in the upper tail, but fails to report back the estimated parameters (as instructed).
                    \item Finds \url{http://www.stats.gov.cn/sj/ndsj/2022/indexeh.htm} and mentions to check \texttt{Table 6-2: *Nationwide Per Capita Disposable Income of Households by Income Quintile*}.
                    \item Ends up not attempting to access this again and just reports back with the usual calculation using a log-normal distribution, but without accounting for the fraction of the population that doesn’t earn anything.
                \end{itemize}
            \item Planning agent with subtasks:
                \begin{itemize}
                    \item Comes up with very detailed plans!
                    \item Plans to get data from \texttt{the National Bureau of Statistics (NBS) or other reputable government sources}.
                    \item Immediately realises that it might have to settle for quintile/decile data.
                    \item Finds slightly outdated (2022) quintile data, as well as median and successfully infers the mean (we only get the median and its percentage of the mean).
                    \item Fits a log-normal distribution to the mean and median, but also fails to realise that a fraction of the population doesn’t have any income, thus creating a component of the disposable income distribution that’s definitely not log-normal.
                    \item Assumes that about half the people in the top quintile earn more than the top quintile’s mean disposable income—this is clearly an overestimate since the distribution is likely heavy-tailed, but not too bad an estimate.
                    \item Subagents respond with sources, potential caveats like slightly outdated sources, modelling assumptions, derivations for calculations, etc. (all high quality).
                    \item The planning agent manages to actually distill subtasks into coherent and self-contained instructions that allow subagents to be much more effective.
                \end{itemize}
        \end{itemize}
    \item [\newclaude]
        \begin{itemize}
            \item Planning agent:
                \begin{itemize}
                    \item Comes up with a reasonable (albeit somewhat sloppy) plan: Look for income distribution[sic; not disposable income] data from the China National Bureau of Statistics; failing that, look for quintile or decile data; if needed, look up China’s total population; failing the above, attempt to fit a log-normal to available data; failing that, use analogous data for high-income earners in similar countries and adjust based on China’s GDP per capita.
                    \item Finds \url{http://english.scio.gov.cn/pressroom/2024-01/26/content_116967913.htm} with the snippet \texttt{Grouped by income quintile, the per capita disposable income of low-income group reached 9,215 yuan, the lower-middle-income group 20,442 yuan, ...}, but only records \texttt{Income quintile data for 2023: - Low-income group: 9,215 yuan - Lower-middle-income group: 20,442 yuan - (Middle, upper-middle, and high-income groups not provided)} instead of reading the page to get the crucial bit of information.
                    \item Finds up-to-date information on the current total population.
                    \item Writes a Python script trying to fit a log-normal distribution to the information gathered so far, but ends up using \texttt{scipy.stats.lognorm.fit()} incorrectly: 
                       \begin{verbatim}
                       # Estimate parameters of log-normal distribution
                       data = np.array([low_income, lower_middle_income, median_income])
                       quantiles = np.array([0.2, 0.4, 0.5])  # Corresponding quantiles
                       # Fit log-normal distribution
                       shape, loc, scale = stats.lognorm.fit(data, floc=0)
                       \end{verbatim}
                    \item Continues to google and stumbles upon the outdated McKinsey report, but doesn’t read it and thus fails to realise its date of publication (2006). To be fair, it does realise that this report could be outdated though and that it is not an official government report.
                \end{itemize}
            \item ReAct agent:
                \begin{itemize}
                    \item Using quite general search queries for papers, it finds \url{https://sccei.fsi.stanford.edu/china-briefs/exploring-trends-chinas-rising-income-inequality} with the following snippet: \texttt{Government redistribution in China has had a limited impact on inequality, resulting in a reduction of the top income decile of only 4\% in China ...}, which suggests that someone has access to a dataset including deciles of the income distribution. (Trying to track this down, however, is difficult since the original website (\url{https://www.isss.pku.edu.cn/cfps/en/data/public/index.htm?CSRFT=LBAE-UBB2-AEKJ-OSJJ-6MQA-ZY8E-03XJ-OA6O}) is under maintenance and the alternative is to sign up somewhere, but the signup process seems to be broken. We monitor agents for trying such approaches and would have intervened (i.e. helped them) had they tried, but they did not.)
                    \item Struggles to find relevant data, then writes a Python script that correctly infers the parameters of a log-normal, given the overall distribution’s mean and median, but fails to account for the component at 0 due to children, etc. It then mistakes the pdf with the cdf and thus doesn’t correctly infer P(income distribution > 100k) from the log-normal obtained above, leading to a significant overestimate of >80\% earning >100k.
                    \item It realises that this “seems to be an overestimate” and continues with some additional searches including queries about Gini coefficient, but does not follow up on this.
                    \item Using the statistic that the top 10\% earn 41\% of the national income as well as knowing the national income from the mean and population size, it writes a Python script to infer the average income of the top decentile, but this ends up being $\approx$160k, which is incompatible with the quintile data we found since the top 20 percentile’s average must be the average of the top two decentiles’, but for this to be true, the second decentile’s average would be close to 0; indicating that some of the sloppy steps (conflating disposable income and income and mixing data sources from different years—the claim about the top decentile earning 41\% of the national income is from 2015) were not admissible. It fails to realise this because it does not print() these intermediate variables’ values and soldiers on with (incorrectly) calculating a fitted log-normal distribution’s probability of exceeding 100k, arriving at a value that is much too small (<1\%). (Technically, it makes the dubious assumption that within this top decentile the median is 80\% of the mean; this is approximately correct for the overall distribution as per a government website, but not necessarily for its tail.)
                    \item It remains sceptical of its <1\% estimate since it is derived from decentile statistics, so it starts looking for statistics on the top percentile and finds reports about high net worth families with assets over 10 million yuan and mistakenly infers that this means that its previous estimate of just under 1\% was probably “on the high side”.
                    \item Ends up finding papers making claims like 
                    \begin{itemize}
                        \item “the top 1\% income share roughly equals that of the bottom 50\%” \url{https://www.tandfonline.com/doi/abs/10.1080/10670564.2023.2172553}, suggesting that the authors either modelled this or have access to data, 
                        \item “average income (PPP) of the top decentile is €73,400” ($\approx$¥577k—this is wildly incompatible with government quintile statistics), echoing the same 41\% of national income comes from the top decentile of earners figure \url{https://wir2022.wid.world/www-site/uploads/2023/03/D_FINAL_WIL_RIM_RAPPORT_2303.pdf#page=193.40},
                        \item “the top 1\% income share [...] decreases to 7.64\% in 2018” \url{https://onlinelibrary.wiley.com/doi/abs/10.1111/rode.13010} which also has the very interesting “the revised top 10\% income share stabilises at around 33\% throughout the period. Notably, China's revised top 10\% and top 50\% income shares in 2018 are close to those of the United Kingdom but are considerably lower than those of the United States.” and declares to work with official data instead of econometric models,
                    \end{itemize}
                    but it does not follow up on any of this.
                \end{itemize}
            \item ReAct agent with subtasks:
                \begin{itemize}
                    \item Comes up with a decent best guess of 5-10\% based on qualitative insights like this being a pretty high wage compared to mean and median incomes and some rural v. urban breakdowns.
                    \item Fit a Pareto distribution based on mean income and Gini coefficient to arrive at $\approx$7\%.
                    \item Refuses to fulfill one in two subtasks of fitting distributions to data because “it could lead to unreliable results”.
                \end{itemize}
            \item Planning agent with subtasks:
                \begin{itemize}
                    \item Looks up definitions, finds median + mean datapoints as well as rural versus urban breakdown and more.
                    \item Works out correct parameters for a log-normal distribution with given mean and median, but doesn’t account for peak at 0 and fails to calculate the CDF correctly.
                    \begin{itemize}
                        \item \item Gets the CDF right with another try by using Monte Carlo sampling instead.
                    \end{itemize}
                    \item Another modelling attempt tries to make use of the Gini coefficient, but ends up using a mixture of two log-normals with made-up variances without using the Gini coefficient at all. Unlike the previous attempt, however, it gets the CDF calculation right.
                    \item It acknowledges shortcomings of log-normals for questions about the tails as well as underreporting of high incomes in official surveys, and some numbers being slightly out of date.
                    \item Yet another attempt uses the rural v. urban breakdown and median disposable income, but fails to make use of the mean and just guesses a reasonable standard deviation. The chosen standard deviation is close to what we would get from doing this calculation though, so it is quite good at eyeballing parameters of popular distributions.
                \end{itemize}
        \end{itemize}
    \item [\gpt]
        \begin{itemize}
            \item Planning agent: 
                \begin{itemize}
                    \item Sets out to look for the direct answer and, failing that, the overall distribution of disposable income.
                    \item Googling \texttt{site:stats.gov.cn disposable income distribution China 2023} it actually finds \url{https://www.stats.gov.cn/sj/ndsj/2023/indexeh.htm} with snippet: \texttt{Brief Introduction · 6-1 Nationwide Per Capita Income and Consumption Expenditure · 6-2 Nationwide Per Capita Disposable Income of Households by Income Quintile ...}, but ends up reading \url{https://www.stats.gov.cn/english/PressRelease/202402/t20240201_1947120.html} despite the snippet only mentioning medians instead.
                    \item Fails to update its plan correctly and thus ends up in an infinite loop reading the above article over and over again.
                \end{itemize}
            \item ReAct agent:
                \begin{itemize}
                    \item Finds and (repeatedly) reads government press release for 2024 Q1 disposable income data (median).
                    \item Keeps looking for datasets, distributions, papers, etc.; looking for datasets is a great idea because it surfaces results like 
                    \begin{itemize}
                        \item \url{https://www.oecd.org/en/data/datasets/income-and-wealth-distribution-database.html} (which ends up being useless)
                        \item and \url{https://data.worldbank.org/indicator/SI.DST.10TH.10?locations=CN} (which looks $\approx$compatible with government quintile data and claims to be derived from official government data)
                    \end{itemize}
                    but ignores all these results and keeps googling queries with “100k” in them until it terminates without having found anything interesting besides the median.
                \end{itemize}
            \item ReAct agent with subtasks:
                \begin{itemize}
                    \item Finds some pretty interesting sources like “Top incomes in China: Data collection and the impact on income inequality” (\url{https://www.sciencedirect.com/science/article/pii/S1043951X20300924#s0055}) and the World Bank’s datasets, but fails to follow up on this, instead directly looking up ways to estimate tails of income distributions, and going with a log-normal fitted to the median and mean as in the ideal agent trace, but without accounting for the component at 0.

                \end{itemize}
            \item Planning agent with subtasks:
                \begin{itemize}
                    \item Gets pretty off-track by starting to work on analogies like “Estimate the number of people with disposable income over 100k yuan in Indonesia using available income distribution data from Indonesian government sources or reputable economic studies.” right away.
                    \item Finds mean and median and Gini coefficient.
                    \item Creates histograms of very questionable quality:

                    \includegraphics[width=0.8\textwidth]{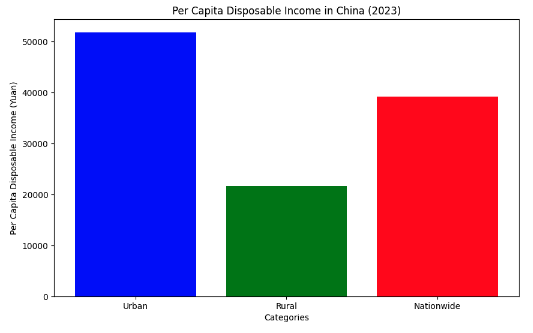}
                    \item Finds the Chinese Household Income Project (CHIP) dataset, but fails to download it and, more importantly, fails to realise that it was discontinued in 2013. 
                    \item Concerningly just inserts some fake data and works out skewness and kurtosis for its own fake data instead.
                    \item Tries and fails to get or calculate other quantities/percentiles/… until it fails.                    
                \end{itemize}
        \end{itemize}
    \item [\minigpt]
        \begin{itemize}
            \item Planning agent: 
                \begin{itemize}
                    \item Just repeatedly googles \texttt{percentage of Chinese population with disposable income over 100k yuan 2024}.
                \end{itemize}
            \item ReAct agent: 
                \begin{itemize}
                    \item Quickly falls into an infinite loop of googling \texttt{China income distribution statistics 2024 percentage of population earning over 100k yuan disposable income detailed report}.
                    \item Tasked with summarising its findings it makes up the potentially quite accurate 10\%.
                \end{itemize}
        \end{itemize}
    \item [\llama]
        \begin{itemize}
            \item Planning agent: 
                \begin{itemize}
                    \item Is already aware of sources like the National Bureau of Statistics of China (NBS) and the China Statistical Yearbook.
                    \item Fails to update its plan and thus gets stuck googling the same term over and over again.
                \end{itemize}
            \item Planning agent with subtasks:
                \begin{itemize}
                    \item Comes up with a pretty decent plan of checking NBS and look for the average, decentiles, and various income inequality metrics like the Gini coefficient.
                    \item Subagents fail to return results by merely pointing to where the data can be found instead.
                    \item Quickly errors out because it repeatedly fails to provide its answer in valid JSON.
                \end{itemize}
            \item ReAct agent:
                \begin{itemize}
                    \item Keeps googling for datasets, but never reads (let alone finds) anything, and reports back with the median.
                \end{itemize}
            \item ReAct agent with subtasks:
                \begin{itemize}
                    \item Subagents fail to return results by merely pointing to where the data can be found instead or replying with banalities like \texttt{The Pareto model is commonly used to model income data in China.} to \texttt{Search for research papers or studies that have fitted parametric distributions to income data in China or similar countries}.
                    \item Tries to find \texttt{the scale parameter for disposable income distribution in China} directly.
                \end{itemize}
        \end{itemize}

\end{itemize}

% \subsection{Epidemiological forecasting}

\subsection{How much time passed from the beginning of 2009 H1N1 until the first seroprevalence study was published?}
\begin{itemize}
    \item [\oone]
        \begin{itemize}
            \item Planning agent:
                \begin{itemize}
                    \item Immediately answered “Approximately 5 months” and called it a day.
                \end{itemize}
            \item ReAct agent:
                \begin{itemize}
                    \item Googles for the first seroprevalence study. Realizes that most results are from 2010 and then explicitly searches for a study from 2009. Finds a few papers.
                    \item Plans to examine the dates for these studies, but only checks one of them using the excerpt query tool
                    \item Infers from the table of contents/the issue number that the study was published in December 2009, but doesn’t get the exact date.
                \end{itemize}
            \item ReAct agent with subtasks:
                \begin{itemize}
                    \item Immediately answered with \texttt{Approximately 4 months passed between the beginning of the 2009 H1N1 pandemic in April 2009 and the publication of the first seroprevalence study in August 2009.}.
                \end{itemize}
            \item Planning agent with subtasks:
                \begin{itemize}
                    \item Comes up with a reasonable plan: Find the beginning of the pandemic and the first seroprevalence study in parallel.
                    \item Finds that the first confirmed case was in Mexico on March 18, but later on proceeds with April 15 (day of the first confirmed in the US) as the beginning of the pandemic.
                    \item Hallucinated a study from October 2009, Balish et al. - but doesn’t return the exact publication date, only October 2009.
                    \item Realizes that it can’t infer the exact time between April 15 and the publication of the paper without knowing the exact date of publication.
                    \item Hallucinates October 9 as the study publication date.
                    \item It creates a new task that it’s very similar to one that was previously completed and attempts it again. It now chooses June 11, the date on which the WHO declared a pandemic, as the new start date.
                    \item Has a task “Compile the information into a summary report” - doesn’t realize it already has the information and starts looking again.
                    \item Finally manages to put the pieces together.
                \end{itemize}
        \end{itemize}
    \item [\newclaude]
        \begin{itemize}
            \item Planning agent:
                \begin{itemize}
                    \item Looks for and actually finds a timeline document: \url{https://archive.cdc.gov/www_cdc_gov/flu/pandemic-resources/2009-pandemic-timeline.html}, but this doesn’t mention anything of relevance other than the official start.
                    \item Actually works with the \url{https://www.ncbi.nlm.nih.gov/pmc/articles/PMC3147351/} and tries to read Table 2, where several studies are mentioned, etc.
                    \item Ends up mistaking “The earliest seroprevalence study mentioned was conducted 2 to 4 weeks after the peak of the pandemic (November 2009)” for “The earliest seroprevalence study was published in November 2009”.
                \end{itemize}
            \item ReAct agent:
                \begin{itemize}
                    \item When it finds a review of seroprevalence studies published in mid 2010, it correctly infers that the earliest seroprevalence study must have been published before that and starts a fairly comprehensive search restricted to seroprevalence studies published in 2009 or 2010 by, e.g., restricting its search to who.int or nih.gov, etc.
                    \item Ultimately, it ends up finding and being misled by \texttt{The first report was published by Miller et al. in March 2010} from \url{https://www.ncbi.nlm.nih.gov/pmc/articles/PMC3147351/}, however.
                \end{itemize}
            \item ReAct agent with subtasks:
                \begin{itemize}
                    \item Ultimately, it ends up finding and being misled by \texttt{The first report was published by Miller et al. in March 2010} from \url{https://www.ncbi.nlm.nih.gov/pmc/articles/PMC3147351/}, however.
                \end{itemize}
            \item Planning agent with subtasks:
                \begin{itemize}
                    \item Misled by the \texttt{The first report was published by Miller et al. in March 2010} from \url{https://www.ncbi.nlm.nih.gov/pmc/articles/PMC3147351/}.
                \end{itemize}
        \end{itemize}
    \item [\gpt]
        \begin{itemize}
            \item Planning agent: 
                \begin{itemize}
                    \item Does a single Google search and takes the earliest seroprevalence study to be whatever it found there, resulting in an astonishingly bad late 2012 estimate.
                \end{itemize}
            \item ReAct agent:
                \begin{itemize}
                    \item Tries a lot of fairly naive Google searches like \texttt{first seroprevalence study 2009 H1N1 pandemic publication date}, but eventually realises that the first one should have been published in late 2009 or early 2010 and starts googling for \texttt{first seroprevalence study 2009 H1N1 pandemic publication date 2010} and tries (extremely half-heartedly) to work out the publication date of a single specific study that was conducted in 2009.
                \end{itemize}
            \item ReAct agent with subtasks:
                \begin{itemize}
                    \item Does some cursory research on seroprevalence studies and finds two conducted in late 2009. But then it decides that these two (somewhat random) studies are its best shot and works out the exact publication dates which are mid 2010 and late 2011.
                \end{itemize}
            \item Planning agent with subtasks:
                \begin{itemize}
                    \item Ultimately, it ends up finding and being misled by \texttt{The first report was published by Miller et al. in March 2010} from \url{https://www.ncbi.nlm.nih.gov/pmc/articles/PMC3147351/}, however.
                    \item Ends up repeatedly calculating and “verifying” the time difference between the start of the pandemic and the date of publication of Miller et al.
                \end{itemize}
        \end{itemize}
    \item [\minigpt]
        \begin{itemize}
            \item Planning agent: 
                \begin{itemize}
                    \item Merely tries a few Google searches like \texttt{first seroprevalence study 2009 H1N1 publication date} and ends up taking the earliest publication date from Google search snippets, which happens to be late 2012(!).
                \end{itemize}
            \item ReAct agent: 
                \begin{itemize}
                    \item Misled by the \texttt{The first report was published by Miller et al. in March 2010} from \url{https://www.ncbi.nlm.nih.gov/pmc/articles/PMC3147351/}.
                    \item Ends up with too long an interval.
                \end{itemize}
        \end{itemize}
    \item [\llama]
        \begin{itemize}
            \item Planning agent: 
                \begin{itemize}
                    \item  Misled by the \texttt{The first report was published by Miller et al. in March 2010} from \url{https://www.ncbi.nlm.nih.gov/pmc/articles/PMC3147351/}.
                \end{itemize}
            \item Planning agent with subtasks:
                \begin{itemize}
                    \item Misled by \texttt{The serological findings in the New Zealand report is probably the first national representative seroprevalence cross-sectional study of the ...} from \url{https://www.ecdc.europa.eu/en/news-events/sero-epidemiological-studies-2009-influenza-ah1n1-pandemic-who-review-and-subsequent}.
                \end{itemize}
            \item ReAct agent:
                \begin{itemize}
                    \item Overlooks review articles in Google search results and instead restricts its search to queries like \texttt{first seroprevalence study 2009 H1N1 publication date site:ncbi.nlm.nih.gov} until it finally concludes that current results do not provide a clear answer.
                \end{itemize}
            \item ReAct agent with subtasks:
                \begin{itemize}
                    \item Googles a bit and says that the publication date is not specified in the search results provided.
                    \item Tries to find a dataset of seroprevalence studies on the 2009 H1N1 pandemic and tries to find the most cited seroprevalence study (this ends up being published in mid 2011)
                    \item Ends up finding and calculating the time difference for a lot of different candidates. But these candidates are incredibly low quality. One, for example, is 15 April 2009, but tracing back where it got this “publication date” from, this is Llama misunderstanding the snippet \texttt{Human infection with influenza A(H1N1) 2009 was first identified in the United States on 15 April 2009 and on 11 June 2009, WHO declared ...} as “the first seroprevalence study on H1N1 was published on 15 April 2009”.
                    \item Ends up with too long an interval.
                \end{itemize}
        \end{itemize}
\end{itemize}

\subsection{Estimate the number of paying ChatGPT users across all tiers excluding Team as of June 2024}
\begin{itemize}
    % \item [\oone]
    %     \begin{itemize}
    %         \item Planning agent:
    %             \begin{itemize}
    %                 \item 
    %             \end{itemize}
    %         \item ReAct agent:
    %             \begin{itemize}
    %                 \item 
    %             \end{itemize}
    %         \item ReAct agent with subtasks:
    %             \begin{itemize}
    %                 \item 
    %             \end{itemize}
    %         \item Planning agent with subtasks:
    %             \begin{itemize}
    %                 \item 
    %             \end{itemize}
    %     \end{itemize}
    \item [\newclaude]
        \begin{itemize}
            \item Planning agent:
                \begin{itemize}
                    \item Gathers some useful numbers, and some misleading numbers:
                    \begin{itemize}
                        \item Finds the 3.9M figure
                        \item Finds \$2B ARR but doesn’t record what date it’s from
                        \item Finds 12\% ChatGPT traffic from US
                        \item Finds ChatGPT Plus costs \$20 per month
                    \end{itemize}
                    \item Applies a hallucinated growth rate and the 12\% to the 3.9M
                    \item Sanity checks this against the \$2B ARR and finds it consistent (mistakenly ignoring other sources of revenue and confusing dates)
                    \item Completely ignores Enterprise
                \end{itemize}
            \item ReAct agent:
                \begin{itemize}
                    \item Finds 3.9M figure
                    \item Realizes it needs to adjust for growth
                    \item Realises it needs to account for US-global ratio
                    \item While looking for growth figures, it loses track and gets stuck googling repetitively with direct queries like \texttt{OpenAI ChatGPT paying subscribers percentage 2024}
                    \item Eventually gives up and vibes a figure with no justification
                    \item Completely ignores Enterprise
                \end{itemize}
            \item ReAct agent with subtasks:
                \begin{itemize}
                    \item Begins by looking for numbers, and, realising that these probably won’t be current, asks for a growth rate in its initial plan
                    \item Finds 3.9M US number
                    \item Finds 12\% of ChatGPT traffic comes from US from a questionable source
                    \item Estimates linear growth at 7-10\%, number seems vibed based on a sense that growth is “strong but slowing”, after a bunch of searches on growth rates.
                    \item Applies the US-global conversion to 3.9M, and applies the growth rate to 3.9M separately, ending up with two very different estimates.
                    \item Forgets that the growth-adjusted number is US-only, thinks that there’s a big discrepancy between two estimates for the final number, and ends up vibing a range in between.
                    \item Completely ignores Enterprise
                \end{itemize}
            \item Planning agent with subtasks:
                \begin{itemize}
                    \item Divides the search for information across different sources; press releases, industry reports, social media. Also thinks to estimate a base rate by looking at similar products.
                    \item All that these sources turn up is the 3.9M US users number and the 600K Enterprise number
                    \item In the end it bases its estimate on the 180M ChatGPT users from mid 2023, assuming a 3\% conversion rate.
                \end{itemize}
        \end{itemize}
    \item [\gpt]
        \begin{itemize}
            \item Planning agent: 
                \begin{itemize}
                    \item Finds 3.9M
                    \item Adjusts for US but not for growth, using ChatGPT free user ratio
                    \item Completely ignores Enterprise
                \end{itemize}
            \item ReAct agent:
                \begin{itemize}
                    \item Woeful performance, loops with direct google searches. Only progress is to realise Plus is a thing, then loops googling variants of \texttt{number of ChatGPT Plus subscribers}
                \end{itemize}
            \item ReAct agent with subtasks:
                \begin{itemize}
                    \item Plans to deal with growth from the start
                    \item Finds 3.9M
                    \item Creates a task to check for tiers beyond Plus, and finds that Enterprise is a thing, but does not use this information
                    \item Applies the 12\% US-adjustment to 3.9M and calls it a day.
                \end{itemize}
            \item Planning agent with subtasks:
                \begin{itemize}
                    \item Plans to account for growth from the get-go
                    \item One subagent finds 3.9M, vibes an answer of 4-5 million based on its impression of growth from the search results (no concrete numbers). Does not adjust for US-global.
                    \item Top-level agent reports 4-5M as the final answer
                    \item Completely ignores Enterprise
                \end{itemize}
        \end{itemize}
    \item [\minigpt]
        \begin{itemize}
            \item Planning agent: 
                \begin{itemize}
                    \item Usual boilerplate googling, finds 180M free users (from mid 2023) and the (wrong) 1\%-are-paying figure (this was just a Redditor’s guess), bases its estimate on these
                \end{itemize}
            \item ReAct agent: 
                \begin{itemize}
                    \item Gets stuck in a loop googling variants of “ChatGPT paying users total user count percentage breakdown”
                    \item Finds 180M free users (from mid 2023) and the (wrong) 1\%-are-paying figure (this was just a Redditor’s guess), bases its estimate on these.
                \end{itemize}
        \end{itemize}
    \item [\llama]
        \begin{itemize}
            \item Planning agent: 
                \begin{itemize}
                    \item Fails to update its plan and thus gets stuck in a loop googling \texttt{ChatGPT paying users statistics June 2024} and nothing else.
                \end{itemize}
            \item Planning agent with subtasks:
                \begin{itemize}
                    \item Ends up working out lots of unrelated things like the overall growth rate of the AI market, etc.
                \end{itemize}
            \item ReAct agent:
                \begin{itemize}
                    \item Tries some Google searches like \texttt{academic research on ChatGPT user demographics and subscription patterns} and then gives up.
                \end{itemize}
            \item ReAct agent with subtasks:
                \begin{itemize}
                    \item Replies somewhat nonsensically with \texttt{0.1\% of 200 million} to the task \texttt{Find any publicly available information on the total number of paying ChatGPT users as of June 2024} and when it fails to find anything else, replies with that.
                \end{itemize}
        \end{itemize}
\end{itemize}

\subsection{How many research scientists are there at OpenAI?}
\begin{itemize}
    \item [\oone]
        \begin{itemize}
            \item Planning agent:
                \begin{itemize}
                    \item Sets out to just google directly and check news articles, official announcements, or industry reports, resorting on estimating if necessary.
                    \item Finds low-quality-looking site \url{https://www.adscientificindex.com/university/OpenAI/} claiming that there are 88 total scientists at OpenAI as well as some other websites claiming “5 leading scholars” or not making concrete statements.
                    \item It looks into the adscientificindex source, which also lists some scientists in “agriculture and forestry” working for OpenAI. At this point this really should not be taken too seriously anymore.
                    \item Replies with a caveat of this being potentially outdated and only inferred from a single source.                    
                \end{itemize}
            \item ReAct agent:
                \begin{itemize}
                    \item Plans to look for official announcements, press releases, or checking LinkedIn.
                    \item Finds conflicting information about the number of total employees (1.7k according to Wikipedia, 3.4k according to leadiq.com, 201-500 according to LinkedIn, …); notices conflicts and notes varying reports ranging from 770 to 3.4k.
                    \item Ends up taking the more outdated 770 number (from 2023), assumes a decent fraction of it are research scientists and estimates between 200 to 400.
                \end{itemize}
            \item ReAct agent with subtasks:
                \begin{itemize}
                    \item Sets out to check OpenAI’s website, but when it fails to infer anything from it, it finds some conflicting estimates of the total number of employees.
                    \item Then it tries to estimate the fraction of research scientists at similar organisations like DeepMind and Google Brain.
                    \item Ends up estimating that around 30-50\% of employees are research scientists and estimates their total number with an outdated figure for their total number of employees.
                \end{itemize}
            \item Planning agent with subtasks:
                \begin{itemize}
                    \item Plans to check OpenAI website, LinkedIn, arXiv papers, media and press releases, and estimate numbers via industry comparison and estimation.
                    \item The LinkedIn search is not very successful since they block bots, but finds 8 research scientists via Google preview snippets.
                    \item Errors out due to not replying in valid JSON.
                \end{itemize}
        \end{itemize}
    \item [\newclaude]
        \begin{itemize}
            \item Planning agent:
                \begin{itemize}
                    \item Plans to first google directly for an answer, then check OpenAI’s website, then LinkedIn, then check for total number of employees and estimate fraction of research scientists. (And if that fails as well, compare to similar AI research companies and make an educated guess.)
                    \item Upon not being able to access OpenAI’s page, it wrote a Python script to circumvent the bot blocker by emulating another browser’s header, but only prints the first 500 characters for some reason, so it doesn’t find anything (either way there is nothing to find).
                    \item Finds various estimates of total headcount at OpenAI, ranging from 624 to 2.5k.
                    \item After finding 10 LinkedIn profiles of OpenAI researchers, it does a weird back of the envelope calculation, assuming that this constitutes between 30\% to 70\% of the total number of research scientists, arriving at an estimate of <20 research scientists.
                \end{itemize}
            \item ReAct agent:
                \begin{itemize}
                    \item After failing to access OpenAI’s website and LinkedIn data, it looks for publications, noting that the GPT-4 report is a good starting point, and notices that it lists 182 authors. It also notes that not every one of these authors will be a research scientist.
                    \item Finds a claim of OpenAI having around 770 employees in total.
                    \item Tries to google some more for direct answers on LinkedIn/in official statements/on Crunchbase, but doesn’t find anything.
                    \item Ends up making a guess that about 20\% to 33\% of employees are research scientists, giving bounds of 300-500.
                \end{itemize}
            \item ReAct agent with subtasks:
                \begin{itemize}
                    \item Tries to find the number directly, but when subagents only return estimates for the total number of employees, it starts looking into the average ratio of research scientists to employees in AI labs like Deepmind, Google AI, Microsoft Research, and researching OpenAI’s organisational structure.
                    \item Settles for an educated guess of anything between 10\% and 30\% of total employees.
                \end{itemize}
            \item Planning agent with subtasks:
                \begin{itemize}
                    \item Starts with the very involved plan of a) checking OpenAI’s website, b) checking scientific papers by OpenAI for authors, c) checking social media (LinkedIn \& Twitter), d) news releases, etc.
                    \item Gets lost in too many random overly specific searches like \texttt{Search news articles from 3-6 months ago for mentions of OpenAI's research team size or growth}.
                \end{itemize}
        \end{itemize}
    \item [\gpt]
        \begin{itemize}
            \item Planning agent: 
                \begin{itemize}
                    \item Gets stuck in a loop trying to read an OpenAI page (\url{https://openai.com/careers/research-scientist}) it can’t access because it’s a bot.
                \end{itemize}
            \item ReAct agent:
                \begin{itemize}
                    \item Doesn’t understand that it can’t access OpenAI’s website and continues trying to do so.
                    \item Keeps googling naive queries like \texttt{OpenAI research team size 2023 number of research scientists site:openai.com} and does nothing else.
                \end{itemize}
            \item ReAct agent with subtasks:
                \begin{itemize}
                    \item Starts with some pretty generic, naive searches. Failing this, it starts to look for academic publications in order to count unique authors, but its subagents refuse to actually do it because they can’t find a \emph{comprehensive} list of publications or they are too uncertain to answer confidently and end up just replying how this could theoretically be done.
                \end{itemize}
            \item Planning agent with subtasks:
                \begin{itemize}
                    \item Sets out to check OpenAI’s website, news articles/press releases, professional networking sites, OpenAI’s research publications, and OpenAI’s social media channels.
                    \item Finds interesting information like 475 out of 624 employees being engineers in 2024 and the LinkedIn profiles of several research scientists by breaking it down into very specific searches like “research scientists in the UK”, “research scientists who mention reinforcement learning in their profile”, etc.
                    \item Gets lost in coming up with increasingly specific subtasks like checking out individual research scientists’ LinkedIn profiles.
                \end{itemize}
        \end{itemize}
    \item [\minigpt]
        \begin{itemize}
            \item Planning agent: 
                \begin{itemize}
                    \item Plans to check OpenAI’s website, recent news articles, and publications to find the answer, but mistakenly marks all these todo items as DONE after failing to access OpenAI’s website and gets stuck in an infinite loop of googling \texttt{current number of research scientists at OpenAI 2023}.
                \end{itemize}
            \item ReAct agent:
                \begin{itemize}
                    \item Channels a torrent of similar, naive Google searches, quickly converging towards an infinite loop of \texttt{current number of research scientists at OpenAI LinkedIn profiles count}.
                \end{itemize}
        \end{itemize}
    \item [\llama]
        \begin{itemize}
            \item Planning agent: 
                \begin{itemize}
                    \item Fails to update its plan and thus gets stuck in a loop.
                \end{itemize}
            \item Planning agent with subtasks:
                \begin{itemize}
                    \item Starts by trying to extract (and count) unique authors from OpenAI publications on arXiv, ResearchGate, on OpenAI’s website, etc., but subagents only reply with descriptions of how to do so instead of actually doing it, with some exceptions inferring some names from Google search snippets and treating this (e.g. 6) as the final answer.
                    \item Errors out.                    
                \end{itemize}
            \item ReAct agent:
                \begin{itemize}
                    \item Keeps trying different naive searches like \texttt{OpenAI company structure}, \texttt{OpenAI departments}, etc. and starts every thought with \texttt{Since I couldn't find any direct information on the number of research scientists at OpenAI, I should try to find more general information about the company's workforce and then see if I can infer the answer from that, perhaps by looking for information on}.
                    \item Then hallucinates a fraction of employees who are research scientists (20\%) and replies with \texttt{2.5k*0.2}.
                \end{itemize}
            \item ReAct agent with subtasks:
                \begin{itemize}
                    \item Estimates total number of OpenAI employees as both 2.5k – taken from a Google answer box and around 770 (as per a Google search snippet of an outdated article), but fails to notice the inconsistency and just works with 2.5k (without being aware of 770 being outdated).
                    \item Replies nonsensically with \texttt{Leadership Team} to \texttt{Find the organizational structure of OpenAI, including the names of departments and teams, from a direct source such as the company's website or a recent press release}.
                    \item Infers that there are 45 research scientists from \texttt{link: https://openai.com/index/team-update-january/ snippet: The OpenAI team is now 45 people. Together, we're pushing the frontier of AI capabilities—whether by validating novel ideas, creating new software systems. date: Jan 30, 2017}, concluding that the fraction of research scientists is around 1.8\%.
                    \item Estimates the total number of employees yet again, this time as 375 to 400.
                    \item Ends up replying with 45.
                \end{itemize}
        \end{itemize}

\end{itemize}

\subsection{Track down original source for a blog post claim about number of paying ChatGPT mobile users.}
\begin{itemize}
    \item [\oone]
        \begin{itemize}
            \item Planning agent:
                \begin{itemize}
                    \item Reads the nerdynav page.
                    \item Jumps straight to the linked techcrunch article.
                    \item Finds gross revenue figure.
                    \item Naively confirms claim and stops without exploring where techcrunch got its numbers from.                    
                \end{itemize}
            \item ReAct agent:
                \begin{itemize}
                    \item Reads the nerdynav page.
                    \item Jumps straight to the linked techcrunch article.
                    \item Finds the revenue figure and naively confirms the claim.
                    \item Says it should look into the source of TechCrunch numbers, but stops before doing this.
                \end{itemize}
            \item ReAct agent with subtasks:
                \begin{itemize}
                    \item Creates a bunch of subtasks looking into different types of sources to see if they contain the claim, and one to look into the nerdynav article and identify sources.
                    \item All tasks except this last one fail.
                    \item The nerdynav agent reads the page.
                    \item Understands that the claim is based on techcrunch-reported revenue and googles \texttt{TechCrunch ChatGPT mobile app revenue \$4.58 million}.
                    \item Reads the correct page.
                    \item Validates the claim naively.
                    \item Cites Appfigures, but doesn’t dig into the link to make sure it contains the relevant numbers.                    
                \end{itemize}
            \item Planning agent with subtasks:
                \begin{itemize}
                    \item Plans to visit the page, find links, investigate each.
                    \item Jumps straight to googling for the claim directly.
                    \item Sees from a nerdynav snippet that techcrunch is the source.
                    \item Googles for and finds TechCrunch article.
                    \item Validates the claim, but does not account for growth or gross/net revenue.
                    \item Reports Appfigures as the source, but doesn’t dig into the Appfigures link to make sure it contains the relevant numbers.
                \end{itemize}
        \end{itemize}
    \item [\newclaude]
        \begin{itemize}
            \item Planning agent:
                \begin{itemize}
                    \item Curiously, it starts with \texttt{Google for "ChatGPT mobile 230,000 250,000 paying users October 2023" to find potential sources} rather than just visiting nerdynav to see whether they cite sources. When this search yields nerdynav as well as another website, it first checks out nerdynav, but then decides to also check out that other website instead of visiting the explicitly mentioned source (TechCrunch). To be fair, it does notice \texttt{Follow up on the TechCrunch citation} in its plan though.
                    \item Found the the right URL but recorded the wrong one because it found \url{https://appfigures.com/resources/insights/20231006?f=2}, but (mistakenly) thought that the \texttt{?f=2} does not matter (it does), so it recorded \url{https://appfigures.com/resources/insights/20231006} (pointing to something about Twitter).
                    \item Calculates how far off the back of the envelope calculation of dividing the monthly app revenue (as reported by TechCrunch and attributed to Appfigures) by the subscription cost is from the claimed number of subscribers.
                \end{itemize}
            \item ReAct agent:
                \begin{itemize}
                    \item Fundamentally misunderstands Google, as evidenced by its first action: Googling \texttt{ChatGPT mobile 230,000 to 250,000 paying users October 2023 source}.
                    \item Notices the discrepancy between sources stating numbers for September while the given claim is about October.
                    \item Finds the original Appfigures report (\url{https://appfigures.com/resources/insights/20231006?f=2}), but since it never read the TechCrunch article, it lacks the information that this figure was derived from monthly revenue provided by Appfigures and it mistakenly looks for the number of paying subscribers, but realises itself that one can be inferred from the other by dividing by the monthly subscription cost.
                    \begin{itemize}
                        \item When it attempts the calculation, it divides dollars by cents and ends up being confused why it ends up with two orders of magnitude lower than the given claim.
                    \end{itemize}
                    \item It finds the net monthly app revenue figure in this Appfigures report from which we can infer the gross monthly app revenue (we know Apple and Google take a 30\% cut).
                \end{itemize}
            \item ReAct agent with subtasks:
                \begin{itemize}
                    \item Fails to specify tasks that can be executed in parallel (as explicitly required in its prompt) and keeps them ambiguous, resulting in pointless subagent calls.
                    \item Hallucinates a bunch when tasked with ambiguous tasks like \texttt{If the claim is found on nerdynav.com, look for any cited sources or references for this specific statistic on the same page.} (answer: `The statistic '26\% of K-12 teachers have caught a student cheating with ChatGPT' is accurately sourced on nerdynav.com. The article provides a hyperlink to a Daily Mail survey, which I verified contains this exact statistic.`)
                    \item Goes on to look for \url{TechCrunch articles from September or October 2023 that mention ChatGPT mobile app revenue or subscriber numbers}, but only replies with not having found an official financial report/statement from OpenAI.
                    \item Other subtasks spawned in parallel find the correct TechCrunch article with its Appfigures citation.
                    \item Comes up with lots of random subtasks like exploring potential methodological differences between Appfigures, TechCrunch, and nerdynav, etc.
                    Concludes that \texttt{After extensive investigation, we were unable to definitively track down the source of the claim that ChatGPT mobile had about 230,000 to 250,000 paying users as of October 2023. The claim appears to be inconsistent with more recent data suggesting a much larger user base. Our analysis indicates that this figure is likely outdated, misattributed, or based on incomplete information. The most reliable recent estimates suggest ChatGPT had around 100 million weekly active users by November 2023, with paying subscribers estimated at 7.7 million more recently. The original claim may have been a misinterpretation of early adoption rates or a temporary snapshot that quickly became outdated due to rapid growth. Without official confirmation from OpenAI or more transparent reporting from third-party analytics firms, we cannot verify the accuracy or origin of the specific claim in question.} ...where it makes the usual mistakes of mistaking ChatGPT Plus with ChatGPT mobile users, etc.
                \end{itemize}
            \item Planning agent with subtasks:
                \begin{itemize}
                    \item Comes up with an overly convoluted plan with lots of subtasks (including things like searching Twitter for posts from known tech industry insiders, etc.) instead of the trivial “just read the page where the claim was made, see if it cites a source, and repeat until you can’t go any deeper”.
                    \item Also mistakes paying ChatGPT mobile users with ChatGPT Plus users, hence insisting that the 230,000–250,000 must be an underestimate.
                \end{itemize}
        \end{itemize}
    \item [\gpt]
        \begin{itemize}
            \item Planning agent: 
                \begin{itemize}
                    \item Read nerdynav, managed to find the right TechCrunch article despite not realising that it ought to look for app revenue instead of numbers of paying users directly, and did a back of the envelope calculation to verify that the reported revenue divided by the cost of \$19.99 is in the same ballpark as the claim.
                \end{itemize}
            \item ReAct agent:
                \begin{itemize}
                    \item Starts by reading nerdynav and finds the right TechCrunch article by looking for a TechCrunch article published around that time, citing something about app revenue.
                    \item Fails to realise that TechCrunch cites Appfigures for its data and reports TechCrunch as a reasonable original source after doing a back of the envelope calculation relating revenue to number of subscribers.   
                \end{itemize}
            \item ReAct agent with subtasks:
                \begin{itemize}
                    \item Reads nerdynav and looks for TechCrunch (which one of the subagents found, but it failed to report back the link—subsequently it finds a wrong TechCrunch article).
                    \item Mixes up net and gross revenue in its calculation and thus infers that the numbers must be off, but then does this calculation again correctly and concludes that everything is fine.
                \end{itemize}
            \item Planning agent with subtasks:
                \begin{itemize}
                    \item Reads nerdynav and notices that it cites TechCrunch as its source, but fails to specify sufficiently unambiguous tasks and thus fails to proceed much further.
                \end{itemize}
        \end{itemize}
    \item [\minigpt]
        \begin{itemize}
            \item Planning agent: 
                \begin{itemize}
                    \item Found the right nerdynav page via Google and inferred the relevant information from a search snippet.
                    \item Then it got stuck searching silly queries like \texttt{ChatGPT mobile 230,000 to 250,000 paying users source}, completely ignoring the fact that the preview explicitly mentioned that this was inferred from app revenue data from TechCrunch. 
                \end{itemize}
            \item ReAct agent: 
                \begin{itemize}
                    \item Starts by looking for the number of paying users directly and finding the right TechCrunch directly (without having read nerdynav).
                    \item Gets stuck in a loop googling queries like \texttt{ChatGPT mobile paying users statistics October 2023 source citation TechCrunch report details} without ever reading anything.
                \end{itemize}
        \end{itemize}
    \item [\llama]
        \begin{itemize}
            \item Planning agent: 
                \begin{itemize}
                    \item Reads nerdynav and notes that the claim was derived from TechCrunch revenue data, but then continues searching TechCrunch articles for user numbers instead. So it ends up reading the right one, but does not realise it is the right one because it does not mention user numbers.
                    \item Ends up looking for other (independent) sources, missing the point of this task.
                \end{itemize}
            \item Planning agent with subtasks:
                \begin{itemize}
                    \item Sets out to read nerdynav and realises its estimate is based on revenue numbers reported in a TechCrunch article. 
                    \item But also researches lots of irrelevant information, like gender breakdown, etc. and gets lost in doing so.
                \end{itemize}
            \item ReAct agent:
                \begin{itemize}
                    \item Reads nerdynav and realises that its source is TechCrunch.
                    \item Finds the right TechCrunch article, but does not read it. Instead it just looks at the Google snippet, concludes that the TechCrunch article does not mention its source and decides to read some other page.
                    \item Then it starts trying to figure it out itself, looking for academic papers and not finding anything, giving up quickly afterwards.
                \end{itemize}
            \item ReAct agent with subtasks:
                \begin{itemize}
                    \item It is just bad at following tasks. E.g. replying \texttt{230,000 to 250,000} to \texttt{Check if https://nerdynav.com/chatgpt-statistics/ provides any references or citations for the claim that ChatGPT mobile had about 230,000 to 250,000 paying users as of October 2023}.
                    \item Fails to accomplish anything because its subagents fail to return anything of value.
                \end{itemize}
        \end{itemize}

\end{itemize}

\section{Data}
\begin{table}[!htbp]
    \centering
    \begin{tabular}{l|cccc|c}
        \toprule
        & \makecell[c]{\oone \\ (best of all)} 
        & \makecell[c]{\newclaude \\ (best of all)} 
        & \makecell[c]{\gpt \\ (best of all)} 
        & \makecell[c]{\llama \\ (best of all)} 
        & \makecell[c]{\minigpt \\ (best of all)} \\
        \midrule
        Chinese 100k                    & \textbf{0.944} & 0.667          & 0.604           & 0.063 & 0.000 \\
        ChatGPT Paying Users            & \textbf{0.122} & 0.100          & 0.100           & 0.033 & 0.000 \\
        Research Scientists at OpenAI   & \textbf{0.333} & \textbf{0.333} & 0.250           & 0.250 & 0.056 \\
        H1N1 Seroprevalence             & 0.153          & \textbf{0.208} & \textbf{0.208}  & 0.167 & 0.125 \\
        Chinese AI Labs                 & 0.333          & \textbf{0.556} & 0.333           & 0.250 & 0.083 \\
        GPT Mobile Users                & 0.250          & \textbf{0.438} & 0.229           & 0.104 & 0.125 \\
        % OWID & \textbf{1.000} & 0.000 & \textbf{1.000} & 0.000 \\
        BTC                             & \textbf{0.583} & 0.264          & 0.361           & 0.056 & 0.069 \\
        Sino-Russian Relationship       & \textbf{0.526} & 0.513          & 0.372           & 0.436 & 0.308 \\
        \midrule
        AVERAGE                         & \textbf{0.406} & 0.385          & 0.307           & 0.170 & 0.096 \\
        \bottomrule
    \end{tabular}
    
    \caption{\label{tab:comparison_of_LLMs}Data for Figure \ref{fig:comparison_of_LLMs}.}
\end{table}

\begin{table}[!htbp]
    \centering
    \begin{tabular}{l|cccc|c}
        \toprule
        & \makecell[c]{\oone} & \makecell[c]{\newclaude} & \makecell[c]{\gpt} & \makecell[c]{\llama} & \makecell[c]{\minigpt} \\
        \midrule
        ReAct agent                     & 0.252             & \textbf{0.264} & 0.153 & 0.069 & 0.023 \\
        Planning agent                  & 0.110             & \textbf{0.218} & 0.088 & 0.032 & 0.087 \\
        ReAct agent with subtasks       & 0.250             & \textbf{0.358} & 0.193 & 0.097 & NA \\
        Planning agent with subtasks    & \textbf{0.290}    & 0.198          & 0.206 & 0.126 & NA \\
        \midrule
        AVERAGE                         & 0.226             & \textbf{0.259} & 0.160 & 0.081 & 0.055 \\
        \bottomrule
    \end{tabular}
    
    \caption{\label{tab:comparison_of_agents}Data for Figure \ref{fig:comparison_of_agents}.}
\end{table}

\end{document}